\documentclass[10pt,journal,compsoc]{IEEEtran}

\ifCLASSOPTIONcompsoc
\usepackage[utf8]{inputenc} 
\usepackage[T1]{fontenc}    
\usepackage{amsmath,amsfonts}
\usepackage{nicefrac} 
\usepackage{array}
\usepackage[caption=false,font=normalsize,labelfont=sf,textfont=sf]{subfig}
\DeclareCaptionFont{customfont}{\fontsize{7}{9}\selectfont\fontfamily{ptm}\selectfont}
\usepackage{url}
\usepackage{graphicx}
\usepackage{algorithmic}
\usepackage{algorithm}
\usepackage{amsthm}
\usepackage{wrapfig} 
\usepackage{multirow}
\usepackage{threeparttable}
\usepackage{booktabs}
\usepackage{adjustbox}
\usepackage{pdfpages}
\usepackage[misc]{ifsym}
\usepackage{xcolor,colortbl}
\usepackage{makecell}
\usepackage{color}
\usepackage{tikz}
\usepackage[edges]{forest}
\usepackage{placeins}
\usepackage{pifont}
\usepackage{xspace} 
\usepackage{chronology}
\usepackage{cite} 
\usepackage[pagebackref,breaklinks=true,colorlinks,citecolor=blue,urlcolor=magenta,linkcolor=red,bookmarks=false,backref=false]{hyperref}
\usepackage{orcidlink}
\usepackage{enumitem}
\usepackage{ragged2e}

\definecolor{myblue}{RGB}{0,176,240}
\definecolor{myorange}{RGB}{238,130,47}
\definecolor{mygreen}{RGB}{0,176,80}
\definecolor{myred}{RGB}{229,76,94}

\usepackage{enumitem}

\newcommand{\xmark}{\ding{55}}
\newcommand{\dmark}{\ding{51}}
\makeatletter
\DeclareRobustCommand\onedot{\futurelet\@let@token\@onedot}
\def\@onedot{\ifx\@let@token.\else.\null\fi\xspace}

\def\eg{\emph{e.g}\onedot} 
\def\ie{\emph{i.e}\onedot} 
 
\def\etc{\emph{etc}\onedot} \def\vs{\emph{vs}\onedot}
 
\def\etal{\emph{et al}\onedot}
\makeatother

\ifCLASSINFOpdf
\else
\fi
            
\begin{document}

\title{Unified Static and Dynamic Network: Efficient Temporal Filtering for Video Grounding
}

\author{Jingjing Hu$^{\orcidlink{0009-0003-6944-0848}} $, Dan Guo$^{\orcidlink{0000-0003-2594-254X}}$,~\IEEEmembership{Senior Member,~IEEE}, Kun Li$^{\orcidlink{0000-0001-5083-2145}}$, Zhan Si$^{\orcidlink{0000-0002-4517-3509}}$, Xun Yang$^{\orcidlink{0000-0003-0201-1638}}$, \\Xiaojun Chang$^{\orcidlink{0000-0002-7778-8807}}$,~\IEEEmembership{Senior Member,~IEEE} and Meng Wang$^{\orcidlink{0000-0002-3094-7735}}$,~\IEEEmembership{Fellow,~IEEE}

\IEEEcompsocitemizethanks{\IEEEcompsocthanksitem J. Hu, D. Guo, K. Li, and M. Wang are with Key Laboratory of Knowledge Engineering with Big Data (HFUT), Ministry of Education, School of Computer Science and Information Engineering (School of Artificial Intelligence), Hefei University of Technology (HFUT), and Intelligent Interconnected Systems Laboratory of Anhui Province (HFUT), Hefei, 230601, China (e-mail: xianhjj623@gmail.com; guodan@hfut.edu.cn; kunli.hfut@gmail.com; eric.mengwang@gmail.com). 

\IEEEcompsocthanksitem Z. Si is with the Department of Chemistry and Centre for Atomic Engineering of Advanced Materials, Anhui University, Hefei, Anhui 230601, P.R. China (e-mail: naa0528@163.com).

\IEEEcompsocthanksitem X. Yang and X. Chang are with the Department of Electronic Engineering and Information Science, School of Information Science and Technology, University of Science and Technology of China, Hefei 230026, China (e-mail: xyang21@ustc.edu.cn; xjchang@ustc.edu.cn).

\IEEEcompsocthanksitem D. Guo and M. Wang are also with the Institute of Artificial Intelligence,
Hefei Comprehensive National Science Center, Hefei, 230026, China.

\IEEEcompsocthanksitem Corresponding authors: D. Guo, X. Yang, X. Chang, M. Wang.} 

\thanks{This work was supported in part by the National Natural Science Foundation of
China (62272144, 72188101, 62020106007, and U20A20183), and the Major Project of Anhui Province (202203a05020011, 2408085J040, 202423k09020001). Fundamental Research Funds for the Central Universities (JZ2024HGTG0309, JZ2024AHST0337 and JZ2023YQTD0072).}}

\markboth{Transactions on Pattern Analysis and Machine Intelligence}
{Shell \MakeLowercase{\textit{et al.}}: Bare Demo of IEEEtran.cls for Computer Society Journals}

\IEEEtitleabstractindextext{
\begin{abstract}
\justifying
Inspired by the activity-silent and persistent activity mechanisms in human visual perception biology, we design a Unified Static and Dynamic Network (UniSDNet), to learn the semantic association between the video and text/audio queries in a cross-modal environment for efficient video grounding. For static modeling, we devise a novel residual structure (ResMLP) to boost the global comprehensive interaction between the video segments and queries, achieving more effective semantic enhancement/supplement. For dynamic modeling, we effectively exploit three characteristics of the persistent activity mechanism in our network design for a better video context comprehension. Specifically, we construct a diffusely connected video clip graph on the basis of 2D sparse temporal masking to reflect the ``short-term effect'' relationship. We innovatively consider the temporal distance and relevance as the joint ``auxiliary evidence clues'' and design a multi-kernel Temporal Gaussian Filter to expand the context clue into high-dimensional space, simulating the ``complex visual perception'', and then conduct element level filtering convolution operations on neighbour clip nodes in message passing stage for finally generating and ranking the candidate proposals. Our UniSDNet is applicable to both \emph{Natural Language Video Grounding (NLVG)} and \emph{Spoken Language Video Grounding (SLVG)} tasks. Our UniSDNet achieves SOTA performance on three widely used datasets for NLVG, as well as three datasets for SLVG, \eg, reporting new records at 38.88\% $R@1,IoU@0.7$ on ActivityNet Captions and 40.26\% $R@1,IoU@0.5$ on TACoS. To facilitate this field, we collect two new datasets (Charades-STA Speech and TACoS Speech) for SLVG task. Meanwhile, the inference speed of our UniSDNet is 1.56$\times$ faster than the strong multi-query benchmark. 
Code is available at: \url{https://github.com/xian-sh/UniSDNet}.
\end{abstract}

}

\maketitle

\IEEEdisplaynontitleabstractindextext

\IEEEpeerreviewmaketitle

\begin{figure}[t]
\centering
\includegraphics[width=1\linewidth]{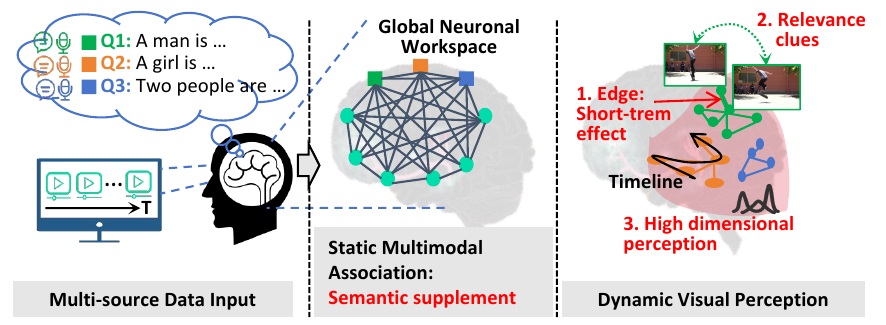}
\caption{
A schematic illustration of the biology behind how people understand the events of a video during solving video grounding tasks. 
Firstly, according to the theory of GNW (\emph{Global Neuronal Workspace})~\cite{deco2021revisiting}, the brain engages in static multimodal information association to achieve semantic complements between multimodalities. Then the focus will be brought to the dynamic perception of the video content along the timeline, and during which three characteristics will be expressed: 1) Short-term Effect: the most recent perceptions have a high impact on the present; 2) Relevance Clues: semantically scenes will provide clues to help understand the current scene; 3) Perception Complexity: visual perception is high-dimensional and non-linear~\cite{barbosa2020interplay}. 
}
\label{fig:task_bio}
\end{figure}

\begin{figure*}[t]
\centering
\includegraphics[width=0.85\textwidth]{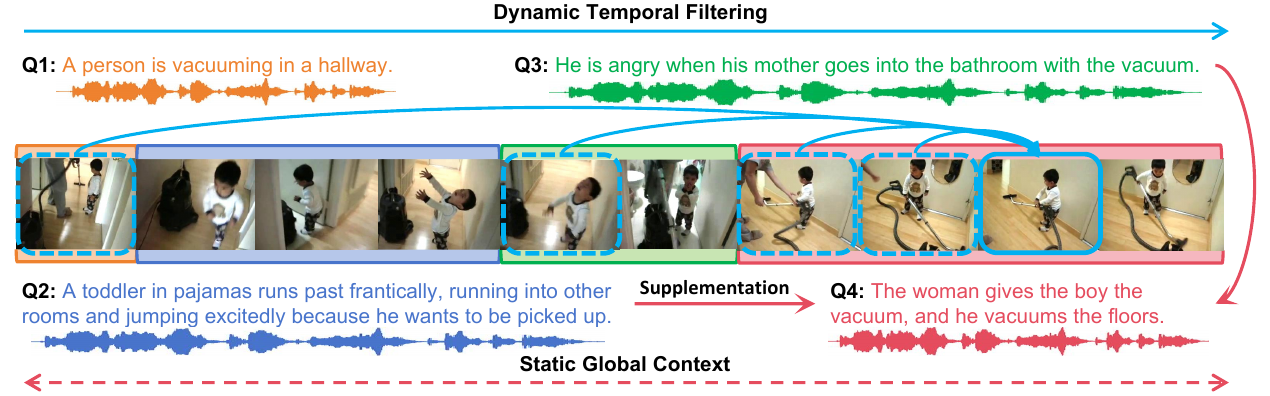}
\caption{ 
An illustrating example for the video grounding task (query: text or audio). This video is described by four queries (events), all of which have separate semantic contexts and temporal dependencies. Other queries can provide a global context (antecedents and consequences) for the current query (\eg, query $Q4$). Besides, historical similar 
scenarios (such as in the {blue} dashed box) help to discover relevant event clues (time and semantic clues) for understanding the current scenario ({blue} solid box). 
}
\label{fig:fe}
\end{figure*}

\section{Introduction}
\label{sec:intro}
\IEEEPARstart{T}emporal Video Grounding (TVG), also called language-queried moment retrieval (MR), as a fundamental and challenging task in video understanding, has gained importance with the surge of online videos, attracting significant attention from both academia and industry in recent years. 
Generally, the TVG task refers to a natural language sentence as a query, with the goal of locating the accurate video segment that semantically corresponds to the query~\cite{gao2017tall,anne2017localizing}, and the task is named \emph{Natural Language Video Grounding (NLVG)}. With the development of Automatic Speech Recognition (ASR) and Text-to-speech (TTS), speech is becoming an essential medium for Human-Computer Interaction (HCI). 
\emph{Spoken Language Video Grounding (SLVG)}~\cite{xia2022video} has also gained a lot of attention.  
We find that whether using text or speech as a query, the key to solving TVG lies in video understanding and cross-modal interaction. 
Our work is devoted to multimodal semantics-driven video understanding, namely, how to aggregate multimodal information for better video understanding?

In this work, we revisit solving TVG tasks through the lens of human visual perception biology~\cite{deco2021revisiting, barbosa2020interplay}, as illustrated in Fig.~\ref{fig:task_bio}. 
We observe that humans quickly comprehend queried events in a video, a process linked to the Global Neuronal Workspace (GNW) theory and dynamic visual perception theory in the brain's prefrontal cortex (PFC)~\cite{deco2021revisiting, barbosa2020interplay}. 
These theories describe the interplay between \emph{activity-silent and persistent activity mechanisms} in the PFC \cite{barbosa2020interplay}. 
The GNW theory suggests that when the brain processes multi-source data, it creates shallow correlations, allowing for semantic complementation between multimodal information. 
This step might not need overly complex deep networks for multimodal interactions between video and language. After that, the brain might pay attention on correlating as much useful information as possible. It will then focus on 
the video content and conduct dynamic visual perception that is transmitted along the \textbf{Timeline Main Clue} and exhibits \textbf{three characteristics}: \textbf{1) Short-term Effect:} nearby perceptions strongly affect current perceptions; 
\textbf{2) Auxiliary Evidence (Relevance) Cues:} semantically relevant scenes in the video provide auxiliary time and semantic cues;
\textbf{3) Perception Complexity:} the perception process is time-series associative and complex, demonstrating high-dimensional nonlinearity~\cite{barbosa2020interplay}. 

Inspired by the above biological theories, we view the process of video grounding as the two-stage cross-modal semantic aggregation, beginning with the global feature interactions of video and language in \emph{text or audio modality}, followed by a deeper video semantic purification based on the dynamic visual perception of the video, and thus design a unified static and dynamic framework for both NLVG and SLVG tsaks. 
\textbf{For the static stage}, static multimodal information will be comprehensively handled based on the language and video features and semantic connections between them are learned. 
\textbf{For the dynamic stage}, we further consider the aforementioned three characteristics of visual perception transmission, and integrate the key ideas of them into our model design. Specifically, as the example shown in Fig.~\ref{fig:fe}, we first comprehensively communicate multiple queries and video clips to obtain contextual information for the current query (\eg, $Q4$) and associate different queries to understand video scenes (\eg, query $Q2$ {supplements} query $Q4$ with more contextual information, in terms of semantics). This video-query understanding process is deemed as a \emph{static global interaction}. 
Then we design a visual perception network to imitate \emph{dynamic context information transmission} in the video with a dynamic filter generation network. 
We build a sparely connected relationship ({blue} arrow in Fig.~\ref{fig:fe}) between video clips to reflect ``Short-term Effect'' (\eg, the video frames in the two dashed boxes closest to the solid {blue} box have the greatest impact on the current solid box frame, in terms of temporal direction and action continuity), 
and collect ``Evidence (Relevance) Clues'' (\eg, the {orange} and {green} video clips in the dashed boxes contain the cause and course of the whole video event, providing the time and semantic clues for current query sub-event) from these neighbor clips ({blue} dashed box in Fig.~\ref{fig:fe}) by conducting a high-dimensional temporal Gaussian filtering convolution 
(in Section~\ref{sec:DTFNet}, imitating visual Perception Complexity). 

Technically, existing methods primarily focus on solving a certain methodological aspect of Temporal Video Grounding tasks, such as learning self-modality language and video representation \cite{xia2022video,rodriguez2023memory}, multimodal fusion \cite{li2021proposal,liu2023exploring}, cross-modal interaction \cite{liu2020jointly,sun2023video}, candidate generation of proposals \cite{zhang2020learning,zhang2021multi}, proposal-based cross-modal matching \cite{gao2021fast,zheng2023phrase}, target moment boundary regression \cite{zhang2021natural,liu2022skimming}, \etc. 
Most current methods prefer to unilaterally consider the static feature interactions by employing the attention computation \cite{zhang2020span, zeng2020dense, mun2020local,li2021proposal, zhang2021multi_msat, zhang2021natural,zheng2022weakly, xia2022video} or 
graph convolution \cite{liu2020jointly, gao2021relation, liu2022skimming, sun2023video} and relation computation \cite{zhang2020learning, xiao2021boundary, gao2021fast, ning2021interaction, zhang2021multi, wang2022negative, zhang2022dual, zheng2023phrase, rodriguez2023memory, zheng2023progressive} to associate the query and related video clips, 
rather than comprehensively expressing both static and dynamic visual perception simultaneously. 
Our work actually proposes a new paradigm for a two-stage unified static-dynamic semantic complementary new architecture. 

In this paper, we propose a novel \textbf{Unified Static and Dynamic Networks (UniSDNet)} for both NLVG and SLVG. The overview of UniSDNet is shown in Fig.~\ref{fig:method}. 
Specifically, \textbf{for static modeling}, we propose a Static Semantic Supplement Network (S$^3$Net), which contains a purely multilayer perceptron within the residual structure (ResMLP) and serves as a static multimodal feature aggregator to capture the association between queries and associate queries with video clips. Unlike the traditional transformer attention~\cite{vaswani2017attention} network, this is a non-attention architecture that constitutes an efficient feedforward and facilitates data training for easy optimization of model performance-complexity trade-offs  (in Section~\ref{sec:3SNet}). 
\textbf{For the dynamic modeling}, we design a Dynamic Temporal Filtering Network (DTFNet) based on a Gaussian filtering GCN architecture to capture more useful contextual information in the video sequence (in Section~\ref{sec:DTFNet}). We firstly construct a diffusely connected video clip graph to reflect the ``\emph{short-term effect}'' relationship between video clip nodes. 
Then we redesign the aggregation of messages from neighboring nodes of the graph network by innovatively introducing the joint clue of the relative temporal distance $r$ between the nodes and the relevance weight of the node $a$ for measuring \emph{relevance between nodes}. We employ the multi-kernel Temporal Gaussian Filter to extend the joint clue to high-dimensional space, and by performing high-dimensional Gaussian filtering convolution operations on neighbor nodes, we imitate \emph{visual perception complexity} and model fine-grained context correlations of video clips.

Notably, our proposed UniSDNet method shows encouraging performance and high inference efficiency in both NLVG and SLVG tasks, as shown in Section~\ref{sec:model_eff}. Particularly, our model achieves higher efficiency (as shown in Fig.~\ref{fig:eff} and Table~\ref{tab:efficiency}). For example, our proposed UniSDNet-M achieves 10. 31\% performance gain on the $R@1, IoU @0.5$ metric while being 1.56× faster than multi-query training SOTA methods PTRM~\cite{zheng2023phrase} and MMN~\cite{wang2022negative}, and, notably, the static and dynamic modules of UniSDNet-M are parameterized only by 0.53M and 0.68M (Table~\ref{tab:config}), respectively.

Our main contributions are summarized as follows:
\begin{itemize}
    \item We make a new attempt in solving video grounding tasks from the perspective of visual perception biology and propose a Unified Static and Dynamic Networks (UniSDNet), where the static module is a fully interactive ResMLP network that provides a global cross-modal environment for multiple queries and the video, and a Dynamic Temporal Filter Network (DTFNet) learns the fine context of the video with query attached. 
    \item In dynamic network DTFNet, we innovatively integrate dynamic visual perception transmission biology mechanisms into the node message aggregation process of the graph network, including a newly proposed joint clue of relative temporal distance $r$ and the node relevance weight $a$, and a multi-kernel Temporal Gaussian Filtering approach. 
    \item In order to facilitate the research about the spoken language video grounding, we collect the new Charades-STA Speech and TACoS Speech datasets with diverse speakers.
    \item We conduct experiments on three public datasets for NLVG and one public dataset and two new datasets for SLVG, and verify the effectiveness of the proposed method. The SOTA performance on NLVG and SLVG tasks demonstrates the generalization of our model.
\end{itemize}

\section{Related works}
\label{sec:relate}

{
Temporal Video Grounding (TVG) includes 
Natural Language Video Grounding (NLVG) and Spoken Language Video Grounding (SLVG). NLVG uses text to locate video moments, while SLVG relies on spoken language. NLVG is widely studied due to advancements in natural language processing, with most existing works focus on it~\cite{zhang2020learning, liu2020jointly, zhang2021multi, zhang2021multi_msat, gao2021relation, ning2021interaction, gao2021fast, yuan2022semantic, wang2022negative, sun2022you, liu2022skimming, zhang2022dual, zheng2022weakly, zheng2023phrase, sun2023video, zheng2023progressive}. SLVG, on the other hand, has gained attention recently due to its flexible speech-based querying. However, NLVG methods cannot be directly applied to SLVG without performance loss~\cite{xia2022video, wang2023weakly}, and few works address SLVG, leaving room for improvement. In this work, we consider both NLVG and SLVG tasks. 
} 
\subsection{Natural Language Video Grounding (NLVG)}

Generally, existing popular methods for solving \emph{NLVG} can be categorized into two main approaches: proposal-free~\cite{zhang2020span, mun2020local, zeng2020dense, li2021proposal, zhang2021natural, xiao2021boundary, xia2022video, rodriguez2023memory, liu2023exploring,li2023vigt} and proposal-based~\cite{zhang2020learning, liu2020jointly, zhang2021multi, zhang2021multi_msat, gao2021relation, ning2021interaction, gao2021fast, yuan2022semantic, wang2022negative, sun2022you, liu2022skimming, zhang2022dual, zheng2022weakly, zheng2023phrase, sun2023video, zheng2023progressive, liu2023m} methods, with detailed comparative methods listed in Section~\ref{sec: nlvg}. 
\emph{1) Proposal-free} methods directly regress the target temporal span based on multimodal features. {These proposal-free methods are mainly often divided into two main categories: Attention-based models~\cite{li2021proposal, zhang2020span, yuan2019find, mun2020local, zeng2020dense} and Transformer-based models~\cite{escorcia2019temporal, lei2020tvr, radford2021learning, lei2021detecting, liu2022umt}.}
\emph{2) Proposal-based methods} use a two-stage strategy of ``generate and rank''. First, they generate video moment proposals, and then rank them to obtain the best match. Herein, 2D-TAN~\cite{zhang2020learning} is the first solution depositing possible candidate proposals via a 2D temporal map for temporal grounding and MMN~\cite{wang2022negative} further 
optimizes it for NLVG by introducing metric learning to align language and video modalities. Because of the elegance of 2D-TAN, we incorporate the concept of 2D temporal map modeling into our model, buffering the possible candidate clues. Our approach is a proposal-based architecture method. 
{Otherwise, some proposal-based methods also focus on using {Attention-based}~\cite{liu2023exploring, gao2021fast, xiao2021boundary, ning2021interaction, yuan2022semantic, sun2022you} and {Transformer-based}~\cite{rodriguez2023memory, zhang2021multi_msat} architectures to address text-video interaction and modal semantic extraction in NLVG tasks. Additionally, some approaches utilize Graph-based architectures~\cite{liu2020jointly, sun2023video, liu2022skimming, gao2021relation} 
for modeling static interactions between video clips.} 
{Although existing NLVG methods have made significant strides in video grounding, but they rely on {single, static architectures}~\cite{liu2023exploring, gao2021fast, xiao2021boundary, ning2021interaction, yuan2022semantic, sun2022you, rodriguez2023memory, zhang2021multi_msat, liu2020jointly, sun2023video, liu2022skimming, gao2021relation}, limiting their ability to capture dynamic interactions as the video progresses.
}

No matter what, regardless of proposal-free or proposal-based manner, previous methods primarily emphasize feature learning with cross-modal attention~\cite{zhang2020span, zeng2020dense, mun2020local, li2021proposal, zhang2021multi_msat, zhang2021multi, zhang2021natural, liu2022skimming}, multi-level feature fusion~\cite{xiao2021boundary, zheng2023phrase}, relational computation~\cite{zhang2020learning, wang2022negative, gao2021fast, ning2021interaction, zhang2022dual}, \etc; all the works are conducted \emph{in a relatively static global perceptual mechanism mode}. 
Additionally, more and more methods are dedicated to capturing \emph{the dynamics of the video}. On one side, temporal feature modeling are studied, such as using RNN to learn the temporal video relationship~\cite{yuan2019find, rodriguez2020proposal} and conditional video feature manipulation~\cite{zheng2023progressive}. On the other side, graph methods are explored for relational learning. For instance, 
CSMGAN~\cite{liu2020jointly} integrates RNN for video temporal capture followed by full-connected graph for cross-modal interaction. 
RaNet~\cite{gao2021relation} and CRaNet~\cite{sun2023video} initially utilize the GC-NeXt~\cite{xu2020g} to aggregate the temporal and semantic context of the video, and then a specially designed semantic graph network is used for cross-modal relational modeling. 
{The current graph models~\cite{liu2020jointly, sun2023video, liu2022skimming, gao2021relation} with Graph Attention Networks (GAN) and Graph Convolutional Networks (GCN) are prominent for modeling static interactions between video clips.
}
They overemphasize the correlation between video clip nodes but ignore the intrinsic high-dimensional time-series nature of the video 
In this work, we 
examine both static feature interactions and dynamic video representation in a unified video grounding framework, considering them in light of the motivations behind human visual perception. 
In the effort to achieve this, we design a lightweight ResMLP network for static semantic complements and exploit the relational learning in a video clip graph. Especially, we fresh sparse masking strategy in a 2D temporal map to build a diffusive connected video clip graph with dynamic Temporal Gaussian filtering for video grounding. 
Extensive experiments in Section~\ref{sec:exp} prove that this artifice is available for both NLVG and SLVG tasks. Such an integrated approach also offers broader applicability across both NLVG and SLVG tasks.

\begin{figure*}[t]
\centering
\includegraphics[width=0.9\textwidth]{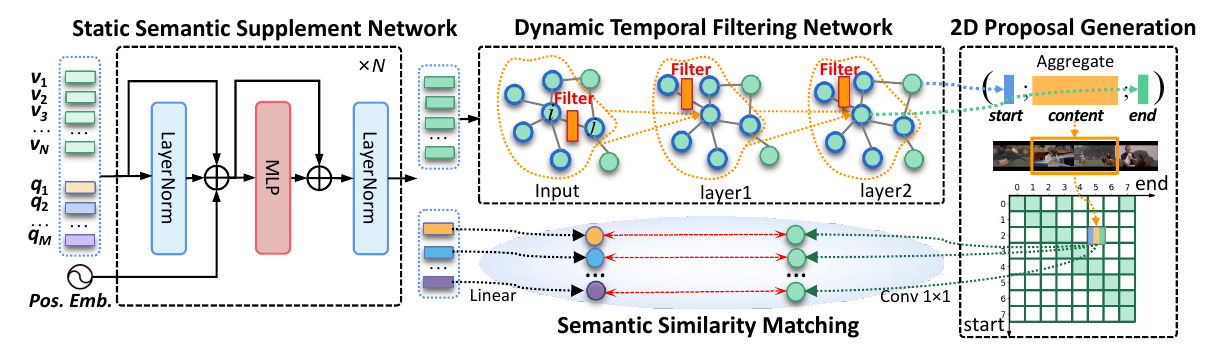}
\caption{ 
\textbf{The architecture of the Unified Static and Dynamic Network (UniSDNet).} 
It mainly consists of static and dynamic networks: Static Semantic Supplement Network (S$^3$Net) and Dynamic Temporal Filtering Network (DTFNet).
\textbf{S$^3$Net} concatenates video clips and multiple queries into
a sequence and encodes them through a lightweight single-stream ResMLP
network.
\textbf{DTFNet} is a 2-layer graph network with a dynamic Gaussian filtering convolution mechanism, which is designed to control message passing between nodes by considering temporal distance and semantic relevance as the Gaussian filtering clues when updating node features. 
The role of 2D temporal map is to retain possible candidate proposals and represent them by aggregating the features of each proposal moment. Finally, we perform semantic matching between the queries and proposals and rank the best ones as the predictions.
}
\label{fig:method}
\end{figure*}

\subsection{Spoken Language Video Grounding (SLVG)}
{ 
To the best of our knowledge, the only available SLVG works at present are VGCL~\cite{xia2022video} and SIL~\cite{wang2023weakly}, both of them have been assessed using the \emph{ActivityNet Speech} dataset that has collected in VGCL's work.   
The VGCL proposes a proposal-free method that utilizes CPC~\cite{oord2018representation} as the audio decoder and transformer encoder as the video encoder to guide audio decoding with the curriculum learning. 
The SIL proposes the acoustic-semantic pre-training to improve spoken language understanding and the acoustic-visual contrastive learning to maximize acoustic-visual mutual information. 
VGCL firstly explore whether the virgin speech rather than text language can highlight relevant moments in unconstrained videos and propose the SLVG task. Compared to NLVG, the challenge of SLVG lies in the discretization of speech semantics and the audio-video interaction. The new task demonstrates that text annotations are not necessary to pilot the machine to understand video. 
Recently, with the development of audio pre-training, a breakthrough has been made in the discretization feature representation of speech~\cite{wav2vec2,data2vec,wang2021unispeech}.In this work, we focus on the audio-video interaction challenge of SLVG through the proposed UniSDNet. 
More importantly, to facilitate the research of SLVG, we collect two new audio description datasets named Charades-STA Speech and TACoS Speech that originate from the NLVG datasets of Charades-STA~\cite{gao2017tall} and TACoS~\cite{regneri2013grounding}. For more details, please refer to Section~\ref{sec:dataset}. 
}

\section{Proposed Framework}
\label{pro}

\subsection{Task Definition \& Framework Overview} \label{sec:task_definition}
The goal of the NLVG (\emph{natural language video grounding}) and SLVG (\emph{spoken language video grounding}) is to predict the temporal boundary $(t^s, t^e)$ 
of the specific moment in the video in response to a given query in text or audio modality. 
Denote the input video as $\mathcal{V} =\{v_i\}^T_{i=1}\in \mathbb{R}^{T\times d^v}$, where $d^v$ and $T$ are the feature dimension and total number of video clips, respectively. 
Each video has an annotation set of $\{\mathcal{Q}, \mathcal{M}\}$, 
in which $\mathcal{Q}$ is a $M$-query set in the \emph{text} or \emph{audio} modality and $\mathcal{M}$ represents the corresponding video moments of the queried events, denoted as $\mathcal{Q}=\{q_i\}^M_{i=1}\in \mathbb{R}^{M\times d^q}$,  and $\mathcal{M} =\{(t_i^s,t_i^e)\}^M_{i=1}$, where $(t_i^s,t_i^e)$ represents the starting and ending timestamps of the $m$-th query, $d^q$ is the dimension of query feature, and $M$ is the query number.

In this paper, we present a unified framework, named \textbf{Uni}fied \textbf{S}tatic and \textbf{D}ynamic \textbf{Net}work (\textbf{UniSDNet}), for both NLVG and SLVG tasks, focusing on video content understanding in the multimodal environment. Fig.~\ref{fig:method} illustrates the overview of our proposed 
architecture. 
Our UniSDNet comprises the \textbf{\emph{Static Semantic Supplement Network (S$^3$Net)}} and \textbf{\emph{Dynamic Temporal Filtering Network (DTFNet)}}. It adopts a two-stage information aggregation strategy, beginning with a global interaction mode to perceive all multimodal information, followed by a graph filter to purify key visual information. Finally, we extract enhanced semantic features of the video clip for high-quality 2D video moment proposals generation. 
In the following subsections, we introduce the core modules, S$^3$Net (Section~\ref{sec:3SNet}), DTFNet (Section~\ref{sec:DTFNet}), and 2D proposal generation (Section~\ref{sec:2d_proposal}) of our proposed unified framework.

\subsection{Static Semantic Supplement Network}
\label{sec:3SNet}
The static network S$^3$Net is inspired by the concept of the global neuronal workspace (GNW)~\cite{deco2021revisiting} in the human brain, which aggregates the multimodal information in the first stage of visual event recognition.
In terms of the functionality of the static network for video understanding, it provides more video descriptions information and significantly fills the gap between vision-language modalities, aiding in understanding video content. 

Technically, the S$^3$Net can be seen as a fully interactive and associative process involving static queries and video features. 
From the aforementioned perspective, we have designed the static semantic supplement network S$^3$Net (as shown in Fig.~\ref{fig:method}) by integrating the MLP into the residual structure (ResMLP). 
The incorporation of a multilayer perceptron within ResMLP enables the fulfillment of 
{
static feature's linear interaction requirement for achieving multimodal information aggregation.}
This setup constitutes an efficient feedforward network that facilitates data training and allows for easy optimization of model performance/complexity trade-offs. Additionally, employing a linear layer offers the advantage of having long-range filters at each layer ~\cite{touvron2022resmlp}.

Before feature interaction, we utilize pre-trained models (C3D~\cite{tran2015learning}, GloVe~\cite{pennington2014glove}, Data2vec~\cite{baevski2022data2vec}, \etc) to extract the original video and query features, which are then linearly converted into a unified feature space. 
This yields video and query features $F_{\mathcal{V}}\in\mathbb{R}^{T\times d}$ and $F_\mathcal{Q}\in \mathbb{R}^{M\times d}$, respectively, with $F_{\mathcal{V}\mathcal{Q}}=[F_\mathcal{V}||F_\mathcal{Q}]\in \mathbb{R}^{(T+M)\times d}$. 
{Inspired by the existing multi-modal Transformers work~\cite{wang2023one, kim2021vilt, yu2022coca}, we independently add position embeddings for video and queries, to distinguish modality-specific information. More ablation studies on adding position embeddings are discussed in Appendix B.2.} 
Specifically, we incorporate the position embedding~\cite{vaswani2017attention} $P_\mathcal{V}\in \mathbb{R}^{T\times d}$ 
for video feature and $P_\mathcal{Q}\in \mathbb{R}^{M\times d}$ for query feature, and concatenate them into $P_{\mathcal{V}\mathcal{Q}}=[P_\mathcal{V}||P_\mathcal{Q}]\in \mathbb{R}^{(T+M)\times d}$. 
We use MLPBlock, which is a combination of a LayerNorm layer, a Linear layer, a ReLU activation layer, and a Linear layer, to obtain the static interactive video clip features $\hat{F}_\mathcal{V}$ and query features $\hat{F}_\mathcal{Q}$:
\begin{equation}\label{eq:1}
\small
\begin{aligned}
& \tilde{F}_{\mathcal{V}\mathcal{Q}}=F_{\mathcal{V}\mathcal{Q}}+{\rm LayerNorm}(F_{\mathcal{V}\mathcal{Q}})+P_{\mathcal{V}\mathcal{Q}},\\
& \hat{F}_{\mathcal{V}\mathcal{Q}} = {\rm LayerNorm}(\tilde{F}_{\mathcal{V}\mathcal{Q}}+{\rm MLPBlock}(\tilde{F}_{\mathcal{V}\mathcal{Q}})),\\
& \hat{F}_\mathcal{V} = \hat{F}_{\mathcal{V}\mathcal{Q}}[1 :T;:]\in\mathbb{R}^{T\times d}, \\
& \hat{F}_\mathcal{Q} = \hat{F}_{\mathcal{V}\mathcal{Q}}[T+1:T+M;:]\in\mathbb{R}^{M\times d}.
\end{aligned}
\end{equation}

Note that UniSDNet can accommodate any number of queries as inputs during training. Proving a single query input is the traditional training mode for NLVG and SLVG tasks. When multiple queries are fed as inputs, there are interactions among the queries, within the video (across multiple video clips), and between the queries and video. 
This approach enables the learning of self-modal and cross-modal semantic associations between video and queries without semantic constraints, allowing the model to leverage the complementary effects among multiple queries related to the same video content. The semantics, either in a single query or multiple queries, can offer more comprehensive semantic supplementation for a effective and efficient understanding of the entire video content.

\subsection{Dynamic Temporal Filtering Network}
\label{sec:DTFNet}
The second stage (DTFNet) of UniSDNet dynamically filters out important video content, inspired by the dynamic visual perception mechanism observed in human activity~\cite{barbosa2020interplay}, as introduced in Section~\ref{sec:intro}. 
We imitate the three characteristics of this visual perception mechanism by learning a video graph network. We restate the key points of these three characteristics here: \textbf{1) Short-term Effect:} nearby perceptions strongly affect current perceptions; 
\textbf{2) Auxiliary Evidence (Relevance) Cues:} semantically relevant scenes in the video provide auxiliary time and semantic cues;
\textbf{3) Perception Complexity:} the perception process is time-series associative and complex, demonstrating high-dimensional nonlinearity~\cite{barbosa2020interplay}. These characteristics play a crucial role in assisting individuals in locating queried events within the video, which have been explained in Fig.~\ref{fig:task_bio} and Fig.~\ref{fig:fe}. 
Graph neural networks have shown efficacy in facilitating intricate information transmission between nodes~\cite{velivckovic2017graph}. 
To emulate the human visual perception process, we introduce a new message passing approach between video clip nodes and propose a Dynamic Temporal Filtering Graph Network (DTFNet as depicted in Figs.~\ref{fig:method} and~\ref{fig:filter}).

To imitate the Short-term Effect, we construct a diffusive connected graph based on the 2D temporal video clip map (please see ``Graph Construction'' below). For discovering Auxiliary Evidence Cues, we integrate the message passed from each node's neighbors by measuring the relative temporal distance and the semantic relevance in the graph (as explained in the filter clue introduced in ``How to construct $\mathcal{F}_{filter}$?'' below). 
Finally, we employ a multi-kernel Gaussian filter-generator to expand the auxiliary evidence clues to a high-dimensional space, simulating the complex visual perception capabilities of humans (explained in the filter function in ``How to construct $\mathcal{F}_{filter}$?'' below). 

\subsubsection{Graph Construction}
Let us denote a video graph $\mathcal{G} =(\mathcal{G_V},\mathcal{G_E})$ to represent the relation in the video $\mathcal{V}$. 
In the graph $\mathcal{G}$, node $v_i$ is the $i$-th video clip and edge $e_{ij}$$\sim$$(v_i,v_j)\in \mathcal{G_E}$ represents whether $v_j$ is $v_i$'s connective neighbor. 
We obtain $\hat{F}_{\mathcal V}$ from the S$^3$Net 
(in Eq.~\ref{eq:1}) and take it as the initialization of clip nodes in the graph, namely the initial node embedding of the graph is set to $\mathcal{G_V}^{(0)}\!=\!\hat{F}_{\mathcal V} \in \mathbb{R}^{T\times d}$.  
{For the graph edge set $\mathcal{G_E}$, we utilize a diffusive connecting strategy~\cite{zhang2020learning} based on the temporal distance of two nodes, to determine the edge status $e_{ij}$. The temporal distance between node \( v_j \) and node \( v_i \) is defined as $r_{ij}=\|j-i\|$, 
setting the hyperparameter $k$, for the current node $v_i$, we define the \textbf{short distance} as $0\leq r_{ij} < k$ and the \textbf{long distance} as $r_{ij}\geq k$.
Based on these two distances, there are two types of edge connections: (1) Dense connectivity for nodes with a short distance: when $0\leq r_{ij} < k$, we densely connect two nodes, \ie, $\mathcal{G_E}_{_{short}} = \{e_{ij}\,|\,0\leq r_{ij} < k\}$. (2) Sparse connectivity for nodes with a long distance: when $r_{ij} \geq k$, we connect them at exponentially spaced intervals, \ie,   
the following conditions should be met when $e_{ij}$ exists: 
\begin{equation}\label{eq:ckk}
\mathcal{G_E}_{_{long}} = \{e_{ij}\}, \quad s.t.\left\{
\begin{aligned}
& i \, \bmod \, 2^{n+1} = 0 \\
& r_{ij} \, \bmod \, (2^n k) = 0 \\
& 2^n k \leq r_{ij} < 2^{n+1} k 
\end{aligned}
\quad,
\right.
\end{equation}
where $n=(0,1,\cdots, \lceil{log_2\frac{T}{k}\rceil-1})$. $\lceil{\cdot\rceil}$ is the ceil function. 
we obtain a sparsely connected edge set $\mathcal{G_E} = \mathcal{G_E}_{_{short}} \cup \mathcal{G_E}_{_{long}}$. 
Please note that we model forward along the timeline, resulting in that the edge set $\mathcal{G_E}$ is reflected as a 
upper triangular adjacency matrix. For more explanation and discussion on the edge construction, please see Appendix B.1. }

\subsubsection{Temporal Filtering Graph Learning}
\begin{figure}[t]
\centering
\includegraphics[width=0.9\linewidth]{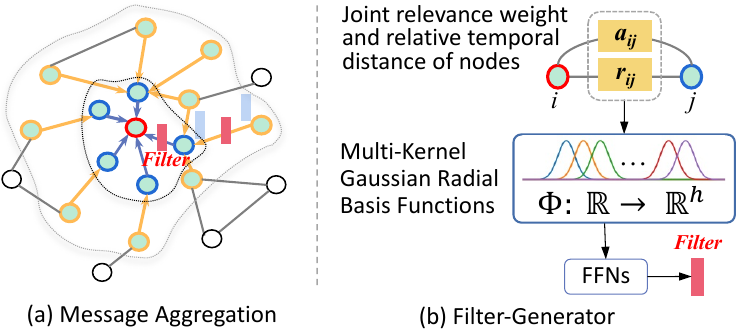}
\caption{The process of (a) node message aggregation in the Dynamic Temporal Filtering graph and (b) 
dynamic filter-generator {\color{red}$Filter$}, which is built based on the joint clue of relevance weight $a_{ij}$ and relative temporal distance $r_{ij}$ between two nodes. This joint clue is expanded into high dimensions representation through a multi-kernel Gaussian radial basis function. 
}
\label{fig:filter}
\end{figure}

We build $L$-layer graph filtering convolutions in our implementation. During training, the node embedding $\mathcal{G_V}^{(l)}=\{v^{(l)}_i\}^T_{i=1}$ is optimized at each graph layer, $1\leq l\leq L$. 
In this part, we introduce a Gaussian Radial \textbf{Filter-Generator} $\mathcal{F}_{filter}$ shown in Fig.~\ref{fig:filter} to imitate the dynamic flashback process of video for visual perception. There are two core technical difficulties to be resolved below.

\textbf{How to construct $\mathcal{F}_{filter}$?} 
Since visual perception is transmitted along the timeline, we consider the relative time interval between nodes as the primary clue. 
Additionally, similar scenes work appropriately on the comprehension of current scene, so we take into account the semantic relevance between graph nodes as auxiliary clue.
Specifically, we compute the two clues of the relative temporal distance $r_{ij}$ of node $v_j$ and node $v_i$ ($r_{ij}=||j-i||$) and the relevance weight $a_{ij}$ of this two-node pair measured by the $cos(\cdot)$ similarity function. We combine them as the joint clue  
$d_{ij} = (1-a_{ij})\cdot r_{ij}$. 
To mimic the dynamic nature, continuity (high dimensionality), and non-linearity (complexity) of visual perception transmission, we use the filter-generating network to dynamically generate 
high-dimensional filter operators that control message passing between nodes, rather than directly applying the simple discrete scalar $d_{ij}$ to compute message aggregation weights, which is insufficient to express these properties. 
The filter-generator (as illustrated in Fig.~\ref{fig:filter}) is given in the form of $\mathcal{F}_{filter}(d_{ij}):\, \mathbb{R} \,\to\,\mathbb{R}^{h}$. 
Gaussian function has already been exploited in deep neural networks, such as Gaussian kernel grouping~\cite{long2019gaussian}, learnable Gaussian fucntion~\cite{zheng2022weakly}, Gaussian radial basis function~\cite{schutt2017schnet} 
that have been proven to be effective in simulating high-dimensional nonlinear information in various scenes. 
Inspired by these works, we adopt multi-kernel Gaussian radial basis to extend the influence of the clue $d_{ij}$ into high-dimensional space, thereby reflecting the continuous complexity of the perception process. 
Specifically, we design a temporal Gaussian basis function to build the $\mathcal{F}_{filter}$ and 
expand the joint clue $d_{ij}$ to a high dimension vector ${f_{ij}}\in \mathbb{R}^{h}$ in message passing process.
We express the form of a single kernel temporal Gaussian as $\phi(d_{ij},z)={\rm exp}({-\gamma(d_{ij}-z)^2})$, 
where $\gamma$ is a Gaussian coefficient that reflects the amplitude of Gaussian kernel function and controls the gradient descent speed of the function value, $z$ is a bias we added to avoid a plateau at the beginning of training due to the highly correlated Gaussian filters. 
Furthermore, we expand it to multiple-kernel Gaussian function $\Phi(d_{ij}, Z)={\rm exp}({-\gamma(d_{ij}-z_k)^2}),\,k\in [1,h]$ to fully represent the complex nonlinear of video perception. 
Based on the single kernel term, we construct $h$ kernel functions, more studies on the settings of ($\gamma,\,z,\,h$) are in the Section~\ref{sec:ana_dtfnet}. 
The way we generate the filter $f_{ij}$ of node $v_j$ to node $v_i$ through the multi-kernel Gaussian filer is: 
\begin{equation}
\label{eq:f}
\begin{aligned}
&{f_{ij}}=\mathcal{F}_{filter}(d_{ij})=(\phi_1(d_{ij}),\phi_2(d_{ij}),\cdots,\phi_{h}(d_{ij})).
\end{aligned}
\end{equation}

\textbf{How to update the nodes in graph $\mathcal{G_V}$?} 
In the stage of message passing on $l$-th layer, we update each node representation by aggregating its neighbor node message to obtain $\mathcal{G_V}^{(l)}$. For node $v_i$, its neighbor set is $\{v_j\,|\,v_j\in \mathcal{N}(v_i)\}$ corresponding to the  adjacency map $\mathcal{G_E}$. With the multi-kernel Gaussian filter $f_{ij}$, 
the update of node feature $v_i$ on $l$-th graph layer is described as:
\begin{equation}\label{eq:gnn}
\small
\begin{aligned}
v^{(l)}_i=\sigma \left(
\sum_{j \in \mathcal{N}(v_i)} {\rm FFN_1} (f_{ij})\odot {\rm FFN}_0(v_j^{(l-1)}) \right),
\end{aligned}
\end{equation}
where $\odot$ represents element-wise multiplication and $\sigma$ is a ReLU activation function. So far, a video graph with spatiotemporal context correlation of video clips is learned.

\subsection{2D Proposal Generation}\label{sec:2d_proposal}
\textbf{Proposal Generation.} 
After obtaining the updated video clip features from the above DTFNet module, we implement the moment sampling~\cite{zhang2020learning} on the features to generate a 2D temporary proposal map ${M}^{2\rm D}\in\mathbb{R}^{T\times T\times d}$ that indicates all candidate moments (2D Proposal Generation in Fig.~\ref{fig:method}). The element $m_{ij}$ in the map ${M}^{2\rm D}$ indicates the candidate proposal $[v_i,\cdots,v_j]$. 
For each moment $m_{ij}$, we consider all the clips in the moment interval and the boundary feature is further added to the moment representation (Eq.~\ref{eq:7}). Afterwards, a stack of 2D convolutions is used to encode the moment feature. For the detailed ablation studies about the moment sampling strategy, please refer to Section~\ref{sec:ana_proposal}. 
\begin{equation}\label{eq:7}
\small
\begin{aligned}
&m_{ij}={\rm MaxPool}(v^L_i,v^L_{i+1},\cdots,v^L_j)+v^L_i+v^L_j \in\mathbb{R}^d, \\
&{M}^{2\rm D}=CNN(m_{ij}) \in\mathbb{R}^{T\cdot T\cdot d}.
\end{aligned}
\end{equation}

\noindent\textbf{Modality Alignment Measurement.} 
We calculate the relevance of each \{\emph{query}, \emph{moment proposal}\} pair according to the semantic similarity, generating new 2D moment score maps for the $M$-queries. 
Specifically, a $1\times 1$ convolution and an FFN are respectively used to project the moment feature $M^{2D}$ and the query feature $\hat{F}_{\mathcal{Q}}$ into the same dimensional vectors ${S}^{\mathcal{M}}\in\mathbb{R}^{T\times T \times d}$ and ${S}^{\mathcal{Q}}\in\mathbb{R}^{M\times d}$. 
{Following MMN~\cite{wang2022negative}, we use cosine similarity to measure the semantic similarity between queries and moment proposals, it is defined as $\tilde{S}={\rm CoSine}({S}^{\mathcal{M}}, {S}^{\mathcal{Q}})$. 
Thereby, $M$ similarity score maps for input $M$ queries are computed:  
\begin{equation}\label{eq:sim}
\small
\begin{aligned}
&{S}^{\mathcal{M}}={\rm Norm}({\rm Conv2d}_{1\times 1}({{M}^{2\rm D}}))\in\mathbb{R}^{T\cdot T\cdot d},\\
&{S}^{\mathcal{Q}}={\rm Norm}({\rm FNN}(\hat{F}_\mathcal{Q}))\in\mathbb{R}^{M\cdot d}, \\
&\tilde{S}
={\rm CoSine}({S}^{\mathcal{M}},{S}^{\mathcal{Q}})\, 
= \,\{\tilde{s}^1,\tilde{s}^2,\cdots,\tilde{s}^M\}\in \mathbb{R}^{(T\times T) \cdot M}, 
\end{aligned}
\end{equation}
where for each query $q_i$, the proposal corresponding to the maximum value in $\tilde{s}^i$ is selected as the best match for the given query $q_i$. There are some other semantic similarity functions for measuring modal alignment. Please refer to Appendix B.3 for relevant ablation study.  
}

\begin{table*}[t]
\centering
\caption{Data statistics of three widely used datasets for NLVG task, ActivityNet Captions, Charades-STA and TACoS datasets.
}
\resizebox{1\linewidth}{!}{
\renewcommand{\arraystretch}{1}
\begin{tabular}{c|c|ccc|ccc|ccc|ccc}
\toprule
{\multirow{2}{*}{{\bf Datasets}}} &{\multirow{2}{*}{{\bf Domain}}} &\multicolumn{3}{c|}{\bf\# Videos} &\multicolumn{3}{c|}{\bf \# Sentences} & \multicolumn{3}{c|}{\bf Average Length}&\multicolumn{3}{c}{\bf Average Queries per Video}\\
& & Train& Val & Test & Train& Val & Test & Video &  Words & Moment & Train & Val & Test \\
\midrule 
ActivityNet Captions~\cite{krishna2017dense} &Open &10,009&4,917&4,885&37,421&17,505&17,031 &117.60s &14.41 &37.14s&3.74&3.56&3.49\\
Charades-STA~\cite{gao2017tall} & Indoors&5,336&-&1,334&12,408&-&3,720& 30.60s &7.22 &8.09s&2.33&-&2.79\\
TACoS~\cite{regneri2013grounding} & Cooking&75 &27&25&9,790&4,436&4001&286.59s &9.42 &27.88s&130.53&164.30&160.04\\
\bottomrule
\end{tabular}
}

\label{tab:dataset_nlvg}
\end{table*}
\begin{table*}[t!]
\centering
\caption{Data statistics of datasets for SLVG task. \protect\emph{Charades-STA Speech$^*$} and \protect\emph{TACoS Speech$^*$} are new datasets collected by us, using machine simulation~\cite{ao2022speecht5} from CMU ARCTIC database, offering more diverse pronunciations than AcitivityNet Speech.}
\resizebox{0.95\linewidth}{!}{
\renewcommand{\arraystretch}{1}
\begin{tabular}{c|c|ccc|ccc|ccc|c}
\toprule
{\multirow{2}{*}{{\bf Datasets}}} &{\multirow{2}{*}{{\bf Domain}}} &\multicolumn{3}{c|}{\bf\# Videos} &\multicolumn{3}{c|}{\bf \# Audios}&\multicolumn{3}{c|}{\bf Average Length}&{\multirow{2}{*}{{\bf Audio Source
}}} \\
& & Train& Val & Test & Train& Val & Test & Video &  Audio & Moment &  \\
\midrule 
ActivityNet Speech~\cite{xia2022video} &Open &10,009&4,917&4,885&37,421&17,505&17,031   &117.60s &6.22s &37.14s& 58 Volunteers\\
\rowcolor{gray!15}
{\bf \emph{Charades-STA Speech$^*$}} & Indoors&5,336&-&1,334& 12,408&-&3,720  &30.60s &{\bf 2.33s} &8.09s& {\bf 3,869 Readers} \\
\rowcolor{gray!15}
{\bf \emph{TACoS Speech$^*$}} & Cooking&75 &27&25& 9790& 4436& 4001 &286.59s &{\bf 2.89s} &27.88s& {\bf 126 Readers}\\
\bottomrule
\end{tabular}
}

\label{tab:dataset_slvg}
\end{table*}

\subsection{Training and Inference}
\label{sec:train}
Our UniSDNet is proposal-based, thereby we optimize the score map $\tilde{S}$ with IoU regression loss and contrastive learning loss. Following 2D-TAN~\cite{zhang2020learning}, we compute the groundtruth IoU Map ${\bf\rm IoU}^{\rm GT} = \{{IoU}^i\}^M_{i=1}\in\mathbb{R}^{(T\times T) \cdot M}$ corresponding to queries. That is, we compute the value of intersection over union between each candidate moment and the target moment $(t^s_{gt},t^e_{gt})$, and scale this value to (0,1), with total $N$ moment scores. The IoU prediction loss is 
\begin{equation}\label{eq:9}
\small
\begin{aligned}
\mathcal{L}_{iou}=\frac{1}{N}{\sum^N_{j=1}{\left(iou_i\cdot{\rm log}(y_{i})+(1-iou_{i})\cdot{\rm log}(1-y_{i})\right)}}, 
\end{aligned}
\end{equation}
where $iou_i$ is the groundtruth from IoU$^{\rm GT}$, and $y_i$ is the predicted IoU value from $\tilde{S}$ in Eq.~\ref{eq:sim}. 

Besides, we adopt contrastive learning~\cite{wang2022negative} as an auxiliary constraint, to fully utilize the positive and negative samples between queries and moments to provide more supervised signals. The noise contrastive estimation~\cite{oord2018representation} is used to estimate two conditional distributions $p(q|m)$ and $p(m|q)$. The former represents the probability that a query $q$ matches the video moment $m$ when giving $m$, and the latter represents the probability that a video moment $m$ matches the query $q$ when giving $q$. 
\begin{equation}\label{eq:10}
\small
\begin{aligned}
\mathcal{L}_{contra}=-(\sum_{q\in{Q}^B}{{\rm log}p(m_q|q)} + \sum_{m\in{M}^B}{{\rm log}p(q_m|m)}),
\end{aligned}
\end{equation}
where ${Q}^B$ and ${M}^B$ are the sets of queries and moments in a training batch. 
$m_q \in \{m^+_q, m^-_q \}$, $m^+_q$ is the moment matched to query $q$ (solo positive sample) and $m^-_q$ denotes the moment unmatched to $q$ in the 
training batch (multiple negative samples). The definition of $q_m \in \{q^+_m, q^-_m \}$ for moment $m$ is similar to that of $m_q \in \{m^+_q, m^-_q \}$. 
The objective of contrastive learning is to guide the representation learning of video and queries and effectively capture mutual matching information between modalities. As a result, the total loss is $\mathcal{L}=\mathcal{L}_{iou}+\mathcal{L}_{contra}$. Non-Maximum Suppression (NMS) threshold is 0.5 during inference.

\section{Datasets}
\label{sec:dataset}
To validate the effectiveness of our proposed unified static and dynamic framework for both NLVG and SLVG tasks, we conduct experiments on the popular video grounding benchmarks. 
There are three classic benchmarks for NLVG task, \ie, \emph{ActivityNet Captions}~\cite{krishna2017dense}, \emph{Charades-STA}~\cite{gao2017tall}, and \emph{TACoS}~\cite{regneri2013grounding} datasets. 
For SLVG task, only the \emph{ActivityNet Speech dataset}~\cite{xia2022video} is publicly available, an extension of \emph{ActivityNet Captions} dataset used for NLVG task. 
To accelerate SLVG development, we collect \textbf{two new Speech datasets}: \textbf{\emph{Charades-STA Speech} and \emph{TACoS Speech}} based on the original Charades-STA and TACoS datasets.

\subsection{Existing Datasets for NLVG Task and SLVG Task}
\label{sec:dataset_nlvg}
The dataset benchmarks used for the NLVG task consist of the untrimmed video and its annotations (text sentence descriptions and video moment pairs). \textbf{\emph{(1) ActivityNet Captions}}~\cite{krishna2017dense} dataset includes 19,209 videos sourced from YouTube's open domain collection, initially proposed by~\cite{krishna2017dense} for dense video captioning task and later utilized for video grounding task. The dataset is divided according to the partitioning scheme in~\cite{zhang2020learning,li2021proposal}; it comprises 37,417, 17,505, and 17,031 sentence-moment pairs for training, validation, and testing, respectively.
\textbf{\emph{(2) Charades-STA}}~\cite{gao2017tall} dataset consists of 9,848 relatively short indoor videos from Charades dataset~\cite{sigurdsson2016hollywood} originally designed for action recognition and localization. It is extended by~\cite{gao2017tall} to include language descriptions for the {NLVG} task, including 12,408 and 3,720 sentence-moment pairs for training and testing, respectively. 
\textbf{\emph{(3) TACoS}}~\cite{regneri2013grounding} dataset focuses on 127 activities within a kitchen, constructed based on the MPII-Compositive dataset~\cite{rohrbach2012script}. Following the split outlined in \cite{zhang2020learning}, the dataset includes 10,146, 4,589, and 4,083 sentence-moment pairs for training, validation and testing, respectively. Compared to \emph{ActivityNet Captions} and \emph{Charades-STA}, the \emph{TACoS} features longer and more annotated queries for each video, with an average of 286.59s and 130.53 per video in the training set.

Currently, there is only one dataset, \emph{ActivityNet Speech} proposed by Xia~\etal\cite{xia2022video}, publicly available for the SLVG task. 
The dataset is collected based on the ActivityNet Captions dataset~\cite{krishna2017dense}, consisting of 37,417, 17,505, and 17,031 audio-moment pairs for training, validation, and testing (as the same split as in \cite{krishna2017dense}), where audio is obtained by 58 volunteers (28 male and 30 female) reading the text fluently in a clean surrounding environment.

\subsection{New Collected Datasets for SLVG Task}
\label{sec:dataset_slvg}
 
In this work, we collected two new datasets to facilitate SLVG research. 
Unlike the \emph{ActivityNet Speech}~\cite{xia2022video} with manual text-to-speech reading, we use \emph{machine simulation} to synthesize audio subtitle datasets and release two \textbf{new \emph{Charades-STA Speech} and \emph{TACoS Speech} datasets}\footnote{\emph{Charades-STA Speech} dataset is available at \url{https://zenodo.org/records/8019213} and \emph{TACoS Speech} dataset is available at \url{https://zenodo.org/records/8022063}}. 
The considerations for adopting the machine simulation are:  
\begin{itemize}
    \item \textbf{High-quality synthesised voice.} Thanks to advancements in text-to-speech (TTS) technology~\cite{ao2022speecht5,pratap2023scaling}, TTS is capable of closely simulating the human voice, effectively capturing and expressing intricate voice characteristics, including speaking style and tone, and generating a high-quality synthesised voice. 
    \item \textbf{Diverse readers.} We randomly select a ``reader'' from the CMU ARCTIC database\footnote{CMU ARCTIC database is available at \url{http://www.festvox.org/cmu_arctic/}} to ``read'' text sentences in Charades-STA and TACoS datasets. The database contains 7,931 vocal embeddings with different English pronunciation characteristics. 
    \item \textbf{Cost savings and high-quality annotation.} With the strong ability of TTS technology to prevent errors like word mispronunciations, incoherent sentence delivery, and audio-text mismatches caused by manual annotation, the necessity for manual text reading, recording, and file annotation processes is mitigated. Machine emulation reduces the cost of manual annotation and avoids manual reading errors. 
\end{itemize}

\begin{figure}[t!]
\centering

\subfloat[ActivityNet Captions]{\includegraphics[width=0.32\linewidth]{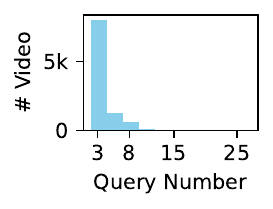}}
\hfill
\subfloat[TACoS]{\includegraphics[width=0.29\linewidth]{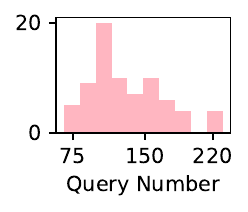}}
\hfill
\subfloat[Charades-STA]{\includegraphics[width=0.29\linewidth]{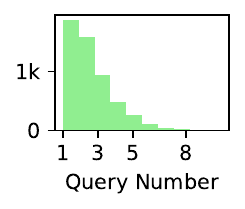}}

\caption{Statistics on the query number size of each video in training set for NLVG\&SLVG datasets (1k=1,000). The datasets can be divided into three categories: large query size (TACoS \& \emph{TACoS Speech}, most sizes are 110), middle query size (ActivityNet Captions \& ActivityNet Speech, most sizes are 3), and small query size (Charades-STA \& \emph{Charades-STA Speech}, most sizes are 1, and the query description is often ambiguous and semantically insufficient as the video is too short with mostly 30s duration for manually annotating events). }
\label{fig:dataset_nlvg}
\end{figure}

Based on the above considerations, we adopt the TTS technology ``microsoft/speecht5\_tts''\footnote{Source code of Microsoft TTS
technology is available at \url{https://huggingface.co/microsoft/speecht5_tts}} to collect the audio description of the text query with a random virtual ``reader'' in the CMU ARCTIC database to guarantee the diversity of voice, style, and tone.  
Compared to the \emph{ActivityNet Speech} dataset, the \emph{Charades-STA Speech} and \emph{TACoS Speech} datasets we collected have more diverse pronunciations. 
The average of each speech recording is 2.33 seconds and 2.89 seconds in the \emph{Charades-STA Speech} and \emph{TACoS Speech} datasets, respectively. It is important to note that the partitioning of both the \emph{Charades-STA Speech} and \emph{TACoS Speech} datasets is consistent with their source datasets \emph{Charades-STA}~\cite{gao2017tall} and \emph{TACoS}~\cite{regneri2013grounding}. We have summarized the statistics of these two new datasets for SLVG task in Table~\ref{tab:dataset_slvg}.

\begin{figure}[t!]
  \centering

  \subfloat[ActivityNet Captions]{\includegraphics[width=0.345\linewidth]{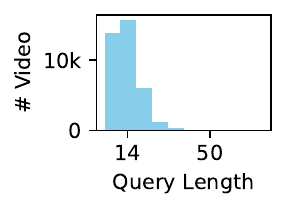}}
  \hfill
  \subfloat[TACoS]{\includegraphics[width=0.315\linewidth]{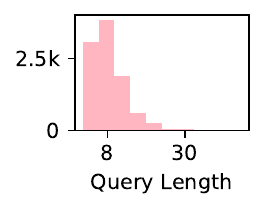}}
  \hfill
  \subfloat[Charades-STA]{\includegraphics[width=0.3\linewidth]{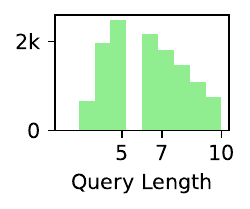}}

  \caption{Statistics on the query length (\emph{counted by word number}) in training set for NLVG\&SLVG datasets (1k=1000). The query length of the ActivityNet Captions dataset are generally long (mostly 14 words and mostly 6s per query), having more detailed descriptions compared to the other two datasets. 
  }
  \label{fig:dataset_slvg}
\end{figure}

\begin{table*}[t]
\centering
\caption{
The hyperparameter settings of UniSDNet framework for different NLVG\&SLVG datasets with the specific pre-extracted video features. It is worth noting that the number of parameters in the static (S$^3$Net) and dynamic (DTFNet) modules of UniSDNet is extremely small on all datasets.
}
\resizebox{1\linewidth}{!}{
\renewcommand{\arraystretch}{1}
\begin{tabular}{c|c|c|cc|ccc|ccc}
\toprule
{\bf Datasets} &{\bf \#Clips}&{\bf Static S$^3$Net} &\multicolumn{2}{c|}{\bf Dynamic DTFNet}&\multicolumn{3}{c|}{\bf 2D Proposal Generation}
&\multicolumn{3}{c}{\bf\#Parameters}\\
&&Hidden size &\#Layers&Hidden size&\#Layers&Kernel size&Hidden size &\cellcolor{gray!15}\textbf{S$^3$Net}~\ref{sec:3SNet} &\cellcolor{gray!15}\textbf{DTFNet}~\ref{sec:DTFNet} &Proposal Generation~\ref{sec:2d_proposal}\\
\midrule
ActivityNet Captions (C3D) &64&1024 &2 &256 &4&9 &512  &\cellcolor{gray!15}{\bf 0.53M} &\cellcolor{gray!15}{\bf 0.68M} &76.79M \\ 
\midrule
Charades-STA (VGG) &16&1024 &2&512  &3&5&512  &\cellcolor{gray!15}{\bf 1.05M} &\cellcolor{gray!15}{\bf 2.68M} &20.19M \\
Charades-STA (C3D) &16&1024 &2&512  &3&5&512  &\cellcolor{gray!15}{\bf 1.05M} &\cellcolor{gray!15}{\bf 2.68M} &20.19M \\
Charades-STA (I3D) &64&1024 &2&256  &2&17&512  &\cellcolor{gray!15}{\bf 0.53M} &\cellcolor{gray!15}{\bf 0.68M} &113.91M \\
\midrule
TACoS (C3D) &128&1024 &2&256  &3&5&512  &\cellcolor{gray!15}{\bf 0.53M} &\cellcolor{gray!15}{\bf 0.68M} &16.65M \\ 

\bottomrule
\end{tabular}
}

\label{tab:config}
\end{table*}

\subsection{Datasets Analysis} \label{sec:dataset_analysis}
First of all, please note that the SLVG datasets are derived from the NLVG datasets, sharing the same video and query sentence. The main difference between them is the modality of query used: SLVG use audio-moment pairs, while NLVG use text-moment pairs. 
The datasets exhibit distinct characteristics in the following aspects: 
\textbf{\emph{(1) Video Duration}}. The average video duration is counted in Table~\ref{tab:dataset_slvg} with the datasets {ActivityNet}, {Charades-STA}, and {TACoS} of 117.60s, 30.60s, and 286.59s, respectively. 
The minimum video duration in the {Charades-STA} implies a stricter judgment of event boundaries than the other two datasets. 
\emph{\textbf{(2) Query Length} (counted by word number in a text sentence or audio duration).} Generally, the longer the audio duration, the more words in the text annotation, and the richer the information provided by the query to describe the video. Notably, the ActivityNet Speech dataset has longer queries (mostly 14 words and mostly 6s per query as shown in Fig.~\ref{fig:dataset_slvg}), providing more detailed descriptions. 
\textbf{\emph{(3) Query Number}}. Fig.~\ref{fig:dataset_nlvg} shows the distributions of the video's query numbers, the datasets can be divided into three categories: large (TACoS), medium (ActivityNet Captions), and small (Charades-STA). 
Particularly, the Charades-STA is minimal with at most 1 query per video, suggesting a potential limitation in description detail provided for the video.

\section{Experiments}
\label{sec:exp}
\subsection{Experimental setup}\label{sec:exp_set}
\textbf{Evaluation Metrics.}
Following the convention in the video grounding and video moment retrieval tasks~\cite{gao2017tall,mun2020local,li2021proposal}, we compute the ``$R@h, IoU@u$'' and ``$mIoU$'' for performance evaluation of both NLVG and SLVG tasks.  
The metric ``$R@h, IoU@u$'' denotes the percentage of samples that have at least one correct answer in the top-$h$ choices, where the criterion for correctness is that the moment ${\rm IoU}$ between the predicted result and the groundtruth is greater than a threshold $u$. 
{Mathematically, ``$R@h, IoU@u$'' is defined as:
\begin{equation}\label{eq:rh-iou}
R@h, IoU@u = \frac{1}{N_q} \sum_{i=1}^{N_q} r(h, u, q_i),
\end{equation}
where $N_q$ denotes the number of queries in the test set and $q_i$ represents the $i$-th query. 
In the top $h$ predicted moments of query $q_i$, if the moment IoU between prediction and groundtruth is larger than $u$, $r(h, u, q_i)$ equals 1; otherwise, $r(h, u, q_i)$=0. }
{Specifically, we set $h\in\{1,5\}$ and $u\in\{0.3,0.5,0.7\}$.}
Also, we use $mIoU$, the average ${\rm IoU}$ between the prediction and groundtruth across the test set, as an indicator to compare overall performance: 
\begin{equation}\label{eq:miou}
{mIoU} = \frac{1}{N_q} \sum_{i=1}^{N_q} {IoU}_i\,,
\end{equation}
where $N_q$ is the total number of queries, and ${IoU}_i$ is the IoU value of the predicted moment for the $i$-th query.

\textbf{Hyperparameter Settings.}
Table~\ref{tab:config} shows hyperparameter settings of UniSDNet. For data preparation, we evenly sample 64 and 128 video clips for ActivityNet Captions dataset with C3D features, and 
16, 16, and 64 video clips for the Charades-STA dataset with VGG, C3D, and I3D features, respectively. 
In the static module, we conduct two ResMLP blocks ($N$=2), and feature hidden size is set to 1024. 
In the dynamic module, DTFNet has two graph layers, 
{Based on the average clips of target moments in training set, hyperparameter $k$ in Eq.~\ref{eq:ckk} -- dividing value between short and long distances in video graph -- is set to 16. More discussion and ablation studies of $k$ are in Appendix B.1.} 
We empirically set hyperparameter $\gamma$ to 10.0, Gaussian kernels number $h$ to 50, 
and generate $h$ biases with equal steps from 0 with step 0.1.
For dynamic filter $\mathcal{F}_{filter}$, settings of convolution layers, kernel size, and hidden size for 2D proposal generation are listed in Table~\ref{tab:config}. 
Parameters size of S$^3$Net (Section~\ref{sec:3SNet}), DTFNet (Section~\ref{sec:DTFNet}) and proposal generation (Section~\ref{sec:2d_proposal}) are also provided in Table~\ref{tab:config}.

\begin{table*}[t]
\centering
\caption{
Comparison with the state-of-the-arts on the \emph{ActivityNet Captions} and \emph{TACoS} datasets for \emph{NLVG} task. ${\ddagger}$ denotes multi-query training mode, others are single-query training mode. UniSDNet-S is single-query training result, and UniSDNet-M is multi-query training result. 
We evaluate our model with two different text feature: GloVe~\cite{pennington2014glove} and BERT~\cite{sanh2019distilbert}. 
}
\resizebox{1\linewidth}{!}{
\renewcommand{\arraystretch}{1}
\begin{tabular}{l|l|l|l|l|ccc|ccc|c|ccc|ccc|c}
\toprule
{\multirow{3}{*}{{\bf }}} &{\multirow{3}{*}{{\bf Methods}}} &{\multirow{3}{*}{{\bf Venue}}} &{\multirow{3}{*}{{\bf Text}}}&{\multirow{3}{*}{{\bf Video}}}&\multicolumn{7}{c|}{\bf ActivityNet Captions}
&\multicolumn{7}{c}{\bf TACoS}\\
\cline{6-19}
\rule{0pt}{6pt}
&&&&&\multicolumn{3}{c|}{\bf R@1, IoU@} &\multicolumn{3}{c|}{\bf R@5, IoU@} &{\multirow{2}{*}{{\bf mIoU}}}&\multicolumn{3}{c|}{\bf R@1, IoU@} &\multicolumn{3}{c|}{\bf R@5, IoU@} &{\multirow{2}{*}{{\bf mIoU}}}\\

{\multirow{2}{*}{}} &{\multirow{2}{*}{}}&{\multirow{2}{*}{}}&{\multirow{2}{*}{}} &{\multirow{2}{*}{}} &\multicolumn{1}{c}{\bf 0.3} &\multicolumn{1}{c}{\bf 0.5} &\multicolumn{1}{c|}{\bf 0.7} &\multicolumn{1}{c}{\bf 0.3}  &\multicolumn{1}{c}{\bf 0.5} &\multicolumn{1}{c|}{\bf 0.7} &{\multirow{2}{*}{}}
&\multicolumn{1}{c}{\bf 0.3} &\multicolumn{1}{c}{\bf 0.5} &\multicolumn{1}{c|}{\bf 0.7} &\multicolumn{1}{c}{\bf 0.3} &\multicolumn{1}{c}{\bf 0.5} &\multicolumn{1}{c|}{\bf 0.7} &{\multirow{2}{*}{}}\\

\midrule

{\multirow{7}{*}{\rotatebox{90}{proposal-free}}}&VSLNet~\cite{zhang2020span} &\emph{ACL'20} &GloVe & C3D & 63.16&43.22 &26.16 & -&-&- &43.19 &29.61 &24.27& 20.03 &-&-&- &24.11 \\

&LGI~\cite{mun2020local} &\emph{CVPR'20} &-&C3D & 58.52&41.51 &23.07 & -&-&-&41.13  &-&-& - &-&-&-&-\\

&CPNet~\cite{li2021proposal} &\emph{AAAI'21} &GloVe&C3D & -&40.56 &21.63 -& -&- &-&40.65  &42.61 &28.29 & -& - &-&- & 28.69\\

&VSLNet-L~\cite{zhang2021natural} &\emph{TPAMI'21} &GloVe&C3D & -&43.86 &27.51  & -&-&-&44.06  &47.11 &36.34 &\textbf{26.42}  &-&-&- &36.61\\

&VGCL~\cite{xia2022video}  &\emph{ACM MM'22} &GloVe&C3D & 60.57&42.96 &25.68  & -&-&-&43.34 &-&-& -& - &-&- &-\\

&METML~\cite{rodriguez2023memory} &\emph{EACL'23}&BERT &I3D & 60.61&43.74 &27.04 & -&-&- &44.05 &-&- & -&-&-&- &-\\

&MA3SRN~\cite{liu2023exploring}&\emph{TMM'23}& GloVe&C3D+Object & -&51.97 &31.39 & -&{84.05} &{68.11} &- &  47.88 &{37.65} &-& 66.02 & 54.27 &- &-\\

\hline
{\multirow{18}{*}{\rotatebox{90}{proposal-based}}}&2D-TAN~\cite{zhang2020learning} &\emph{AAAI'20}  &GloVe&C3D & 59.45&44.51 	&26.54 	 & 85.53&77.13 	&61.96 &-   &37.29 	&25.32 & - &57.81 	&45.04&- &-\\

&CSMGAN~\cite{liu2020jointly} &\emph{ACM MM'20} & GloVe&C3D & 68.52&49.11 &29.15 & 87.68&77.43 &59.63&-    &33.90 &27.09 & - &53.98& 41.22&- &-\\

&MS-2D-TAN~\cite{zhang2021multi} &\emph{TPAMI '21}&GloVe&C3D  & 61.04&46.16 	&29.21 	 & 87.30&78.80 	&60.85&- &45.61 	&35.77& 23.44 &69.11 	&57.31&36.09 &-  \\

&MSAT~\cite{zhang2021multi_msat}   &\emph{CVPR'21} &- &C3D & -&48.02 &31.78 & -&78.02 &63.18&-  &48.79 &37.57 & - &67.63 &57.91&- &-\\

&RaNet~\cite{gao2021relation} &\emph{EMNLP'21}&GloVe&C3D & -&45.59 &28.67 & -&75.93 &62.97&-  &43.34& 33.54 & - &67.33 &55.09&- &-\\

&I$^2$N~\cite{ning2021interaction} &\emph{TIP'21}  &GloVe&C3D & - &-& -& -& - &-  & - &31.47&29.25& - &52.65&46.08&- &-\\

&FVMR~\cite{gao2021fast} &\emph{ICCV'21}&GloVe&C3D & 60.63&45.00& 26.85  & 86.11&77.42 &61.04&-  &41.48 &29.12 & - &64.53 &50.00&- &-\\

&SCDM~\cite{yuan2022semantic} &\emph{TPAMI'22}&GloVe&C3D & 55.25&36.90 &20.28 & 78.79&66.84 &42.92 &- &27.64 &23.27 & - &40.06 &33.49&- &-\\

&MGPN~\cite{sun2022you} &\emph{SIGIR'22} &GloVe&C3D & -&47.92 &30.47 & -&78.15 &63.56&-  &{48.81} &36.74 & - &{71.46} &{59.24}&-&-\\

&SPL~\cite{liu2022skimming} &\emph{ACM MM'22}&GloVe&C3D & -&52.89 &{32.04} & -&{82.65} &{67.21}&-   &42.73 &32.58 & - &64.30 &50.17&- &-\\

&DCLN~\cite{zhang2022dual} &\emph{ICMR'22}  &GloVe&C3D  & 65.58&44.41 &24.80 & 84.65&74.04 &56.67&-  & 44.96& 28.72 & - &66.13 &51.91&- &-\\

&CRaNet~\cite{sun2023video} &\emph{TCSVT'23} &GloVe&C3D & -&47.27 &30.34 & -&78.84 &63.51 &-    &47.86 &37.02 & -&70.78 &58.39&- &-\\

&PLN~\cite{zheng2023progressive} &\emph{ACM MM'23} &GloVe&C3D  & 59.65&45.66 &29.28  & 85.66&76.65 &63.06 &44.12  &43.89 &31.12& -  &65.11 &52.89 &- &{29.70}\\

&M$^2$DCapsN~\cite{liu2023m} &\emph{TNNLS'23} &GloVe&C3D & 61.53&47.03&29.99  & -&76.64&62.83&-  &46.41&32.58& - &66.32&52.91&- &-\\

&MMN${\ddagger}$~\cite{wang2022negative} &\emph{AAAI'22} &BERT &C3D & -&48.59 &29.26 & -&79.50 &64.76  &-	 &39.24 &26.17& - & 62.03 &47.39&- &-\\

&PTRM${\ddagger}$~\cite{zheng2023phrase} &\emph{AAAI'23} &BERT &C3D & 66.41&50.44 &31.18  & -&-&-&{47.68} &-&-& - &-&-&- &-\\

&DFM${\ddagger}$~\cite{mm_dfm} &\emph{ACM MM'23} &BERT &C3D & -&45.92 &32.18 & -&-&-&- &40.04 &28.57 & 14.77 &-&-&- &27.35\\

&\multicolumn{2}{l|}{\cellcolor{gray!15}{\bf UniSDNet-S (Ours) }} &\cellcolor{gray!15}GloVe &\cellcolor{gray!15}C3D &\cellcolor{gray!15}68.59  &\cellcolor{gray!15}52.73 &\cellcolor{gray!15}31.08 &\cellcolor{gray!15}\underline{89.57} &\cellcolor{gray!15}84.19 &\cellcolor{gray!15}\underline{72.52} 	&\cellcolor{gray!15}50.13 &\cellcolor{gray!15}51.44 &\cellcolor{gray!15}36.37  &\cellcolor{gray!15}{23.47}	&\cellcolor{gray!15}76.56 &\cellcolor{gray!15}61.06  &\cellcolor{gray!15}{36.22}	 &\cellcolor{gray!15}35.83 \\

&\multicolumn{2}{l|}{\cellcolor{gray!15}{\bf UniSDNet-S (Ours) }} &\cellcolor{gray!15}BERT &\cellcolor{gray!15}C3D  &\cellcolor{gray!15}68.66  &\cellcolor{gray!15}52.35   &\cellcolor{gray!15}32.25  &\cellcolor{gray!15}89.74 &\cellcolor{gray!15}83.35   &\cellcolor{gray!15}70.61  &\cellcolor{gray!15}50.22 &\cellcolor{gray!15}53.46 &\cellcolor{gray!15}36.24	&\cellcolor{gray!15}{23.48}	&\cellcolor{gray!15}76.96		&\cellcolor{gray!15}63.06	&\cellcolor{gray!15}{36.34}	 &\cellcolor{gray!15}36.47
\\

&\multicolumn{2}{l|}{\cellcolor{gray!15}{\bf UniSDNet-M (Ours) }} &\cellcolor{gray!15}GloVe &\cellcolor{gray!15}C3D   &\cellcolor{gray!15}\underline{74.07} &\cellcolor{gray!15}\underline{57.67}    &\cellcolor{gray!15}\underline{35.64}  &\cellcolor{gray!15}\underline{90.49}    &\cellcolor{gray!15}\underline{84.46}   &\cellcolor{gray!15}72.47 &\cellcolor{gray!15}\underline{53.68} &\cellcolor{gray!15}\underline{53.59}  &\cellcolor{gray!15}\underline{38.34} &\cellcolor{gray!15}{23.79}	  &\cellcolor{gray!15}\textbf{79.01}    &\cellcolor{gray!15}\textbf{64.83}   &\cellcolor{gray!15}\underline{36.89}	 &\cellcolor{gray!15}\underline{37.54}\\

&\multicolumn{2}{l|}{\cellcolor{gray!15}{\bf UniSDNet-M (Ours) }} &\cellcolor{gray!15}BERT &\cellcolor{gray!15}C3D  &\cellcolor{gray!15}\textbf{75.85}  &\cellcolor{gray!15}\textbf{60.75}    &\cellcolor{gray!15}\textbf{38.88}   &\cellcolor{gray!15}\textbf{91.17}    &\cellcolor{gray!15}\textbf{85.34}    &\cellcolor{gray!15}\textbf{74.01}   &\cellcolor{gray!15}\textbf{55.47}
&\cellcolor{gray!15}\textbf{55.56}	&\cellcolor{gray!15}\textbf{40.26}	&\cellcolor{gray!15}\underline{24.12}	 &\cellcolor{gray!15}\underline{77.08}	&\cellcolor{gray!15}\underline{64.01}	&\cellcolor{gray!15}\textbf{37.02}	 &\cellcolor{gray!15}\textbf{38.88}
\\

\bottomrule
\end{tabular}
}

\label{tab:nlvg_activity}
\end{table*}

\begin{table*}[t]
\centering

\caption{
Comparison with the state-of-the-arts on the \emph{Charades-STA} dataset for \emph{NLVG} task. ${\ddagger}$ denotes multi-query training mode. Both MMN and our method originate from the exploitation of 2D temporal map. The single-query (UniSDNet-S) and multi-query (UniSDNet-M) training results on this dataset are closest compared to the other two NLVG datasets, due to its distribution of the query number concentrated in 1 (as shown in Fig.~\ref{fig:dataset_nlvg}). 
}
\resizebox{1\textwidth}{!}{
\renewcommand{\arraystretch}{1}
\begin{tabular}{p{0.1cm} p{0.1cm}|l|cc|cc|c|cc|cc|c|cc|cc|c}
\toprule

{\multirow{3}{*}{{\bf }}}&{\multirow{3}{*}{{\bf }}}&{\multirow{3}{*}{{\bf Methods}}} &\multicolumn{5}{c|}{\bf Video Feature: VGG}&\multicolumn{5}{c|}{\bf Video Feature: C3D}&\multicolumn{5}{c}{\bf Video Feature: I3D}\\
\cmidrule(lr){4-18}

&& &\multicolumn{2}{c|}{\bf R@1, IoU@}&\multicolumn{2}{c|}{\bf R@5, IoU@}  &{\multirow{2}{*}{{\bf mIoU}}}&\multicolumn{2}{c|}{\bf R@1, IoU@}&\multicolumn{2}{c|}{\bf R@5, IoU@}  &{\multirow{2}{*}{{\bf mIoU}}}&\multicolumn{2}{c|}{\bf R@1, IoU@}&\multicolumn{2}{c|}{\bf R@5, IoU@}  &{\multirow{2}{*}{{\bf mIoU}}}\\
 
&& & {\bf 0.5} &\multicolumn{1}{c|}{\bf 0.7} & {\bf 0.5} &\multicolumn{1}{c|}{\bf 0.7} &{\multirow{2}{*}{}}& {\bf 0.5} &\multicolumn{1}{c|}{\bf 0.7} & {\bf 0.5} &\multicolumn{1}{c|}{\bf 0.7} &{\multirow{2}{*}{}}& {\bf 0.5} &\multicolumn{1}{c|}{\bf 0.7} & {\bf 0.5} &\multicolumn{1}{c|}{\bf 0.7} &{\multirow{2}{*}{}}\\

\midrule
{\multirow{4}{*}{\rotatebox{90}{proposal-}}}&{\multirow{4}{*}{\rotatebox{90}{free}}}&DRN~\cite{zeng2020dense} &-&-&-&-&-&45.40 &26.40 &\underline{88.01} &55.38&-&53.09 &31.75 &89.06 &60.05&-\\
&&LGI~\cite{mun2020local} &-&-&-&-&-&-&-&-&-&-&59.46 &35.48 &-&-&51.38\\
&&BPNet~\cite{xiao2021boundary}   &-&-&-&-&-&38.25& 20.51&-&-&38.03&50.75 &31.64 &-&-&46.34\\
&&CPNet~\cite{li2021proposal}  &-&-&-&-&-&40.32 &22.47&-&-& 37.36&\underline{60.27}&38.74 &-&-&{52.00}\\
\midrule
\multicolumn{2}{c|}{{\multirow{13}{*}{\rotatebox{90}{proposal-based}}}}&2D-TAN~\cite{zhang2020learning}  &42.80 	&23.25 	&80.54 	&54.14&-&-&-&-&-&-&-&-&-&-&-\\

&&MS-2D-TAN~\cite{zhang2021multi} &45.65 &27.20 &\textbf{86.72} &56.42&-&41.10 &23.25& 81.53& 48.55&-&60.08& 37.39 &89.06 &59.17&-\\

&&FVMR~\cite{gao2021fast}    &-&-&-&-&-&38.16 &18.22 &82.18 &44.96&-& 55.01 &33.74 &89.17 &57.24&-\\ 
&&I$^2$N~\cite{ning2021interaction}  &-&-&-&-&-&-&-&-&-&-&{56.61} &34.14&81.48&55.19&-\\ 
&&CPL~\cite{zheng2022weakly} &-&-&-&-&-&-&-&-&-&-&49.05 &22.61 &84.71 &52.37&-\\
&&PLN~\cite{zheng2023progressive} &45.43 &26.26 & 86.32 &57.02 &41.28&-&-&-&-&-&56.02&{35.16} & 87.63 &62.34 &49.09\\
&&PTRM${\ddagger}$~\cite{zheng2023phrase}  &\underline{47.77}& \underline{28.01}&-&-&{42.77}&-&-&-&-&-&-&-&-&-&-\\
&&CRaNet~\cite{sun2023video} &47.12 &27.39 &83.51 &58.33&-&-&-&-&-&-&\underline{60.94} &\textbf{41.32} &\textbf{89.97} &{65.19}&-\\ 
&&M$^2$DCapsN~\cite{liu2023m} &43.17&25.13&79.35&55.86
&-&40.81 &23.98&77.93&53.52&- &55.03 &31.61&84.33&63.71&-\\

&&MMN${\ddagger}$~\cite{wang2022negative}    &47.31 &27.28  &83.74 &\underline{58.41}&-	&-&-&-&-&-&-&-&-&-&-\\

&&\cellcolor{gray!15}{\bf UniSDNet-S (Ours)}&\cellcolor{gray!15}47.34 &\cellcolor{gray!15}27.45  &\cellcolor{gray!15}84.68   &\cellcolor{gray!15}\underline{58.41}  &\cellcolor{gray!15}\underline{43.32} &\cellcolor{gray!15}\underline{48.71}  &\cellcolor{gray!15}\underline{27.31}   &\cellcolor{gray!15}82.77  &\cellcolor{gray!15}\underline{57.58} &\cellcolor{gray!15}\underline{43.16}&\cellcolor{gray!15}59.41   &\cellcolor{gray!15}38.58   &\cellcolor{gray!15}\underline{89.52}  &\cellcolor{gray!15}\underline{70.65} &\cellcolor{gray!15}\underline{52.07} \\

&&\cellcolor{gray!15}{\bf UniSDNet-M (Ours)}&\cellcolor{gray!15}{\bf48.41}	&\cellcolor{gray!15}\textbf{28.33}&\cellcolor{gray!15}\underline{84.76} &\cellcolor{gray!15}\textbf{59.46}	&\cellcolor{gray!15}\textbf{44.41}
&\cellcolor{gray!15}\textbf{49.57} &\cellcolor{gray!15}\textbf{28.39}&\cellcolor{gray!15}\underline{84.70} &\cellcolor{gray!15}\textbf{58.49}	&\cellcolor{gray!15}\textbf{44.29}
&\cellcolor{gray!15}\textbf{61.02}&\cellcolor{gray!15}\underline{39.70}&\cellcolor{gray!15}\textbf{89.97} &\cellcolor{gray!15}\textbf{73.20} &\cellcolor{gray!15}{\bf52.69}\\

\bottomrule
\end{tabular}
}

\label{tab:sta_main_new}
\end{table*}

\begin{table*}[h]
\centering
\caption{Comparison with state-of-the-art methods on three datasets for \textit{SLVG} task, in which \emph{Charades-STA Speech$^*$} and \emph{TACoS Speech$^*$} are our new collected datasets, described in Section~\ref{sec:dataset}. ${\ddagger}$ denotes multi-query training mode, $^\dag$ denotes our reproduced results using the released code. 
}
\resizebox{1\linewidth}{!}{
\begin{tabular}{cl|c|c|ccc|ccc|c}
\hline
{\multirow{2}{*}{{\bf Dataset}}} &{\multirow{2}{*}{{\bf Method}}}&{\multirow{2}{*}{{\bf Audio Feature}}}&{\multirow{2}{*}{{\bf Video Feature}}}& & {\bf R@1, IoU@} & & & {\bf R@5, IoU@} & &{\multirow{2}{*}{{\bf mIoU}}}\\

&& & & {\bf0.3} & {\bf0.5} & {\bf0.7} & {\bf0.3} & {\bf0.5} & {\bf0.7} &{\multirow{2}{*}{}}\\
 
\hline
{\multirow{6}{*}{ActivityNet Speech~\cite{xia2022video}}}&VGCL~\cite{xia2022video} &CPC~\cite{oord2018representation} &{\multirow{6}{*}{C3D}} & 49.80 & 30.05 & 16.63 & - & - & - &35.36\\

&ISL~\cite{wang2023weakly}&Mel Spectrogram & &49.46 &30.26 &15.22 &82.28 &63.73 &35.48 &34.52\\

&VSLNet~\cite{zhang2020span} & Mel Spectrogram & &46.75 &29.08 &16.24 & - & - & - & 34.01\\

&VSLNet$^\dag$ &Data2vec~\cite{data2vec} &  & 51.02 & 30.38 & 17.45 & - & - & - & 37.04 \\

&MMN${\ddagger}$~\cite{zhang2021multi}$^\dag$ &Data2vec & & {51.98}	& {35.69}	& {20.77} & {85.46}	& {75.29}	& {56.87}	& {37.81}\\

&{\bf UniSDNet-S}&Data2vec & &\underline{64.83}   &\underline{47.82}  &\underline{27.49}  &\underline{90.69}    &\underline{84.16}    &\underline{72.12}  &\underline{47.31}\\

&\cellcolor{gray!15}{\bf UniSDNet-M} &\cellcolor{gray!15}Data2vec &\cellcolor{gray!15} &\cellcolor{gray!15}\textbf{72.27}	&\cellcolor{gray!15}\textbf{56.29}	&\cellcolor{gray!15}\textbf{33.29}	&\cellcolor{gray!15}\textbf{90.41} &\cellcolor{gray!15}\textbf{84.28} &\cellcolor{gray!15}\textbf{72.42}	&\cellcolor{gray!15}\textbf{52.22}\\

\cline{2-11}

&VSLNet$^\dag$ & & &53.06 &32.43 &17.69 &- &- &- &37.22   \\

&MMN${\ddagger}$$^\dag$ &Data2vec &I3D  &53.23   &35.53  &20.09  &83.77 &72.76  &55.88 &38.24 \\

&{\bf UniSDNet-S}& & &\underline{64.16} &\underline{49.28} &\underline{27.94} &\underline{90.05} &\underline{83.38} &\underline{67.09} &\underline{47.47} \\

&\cellcolor{gray!15}{\bf UniSDNet-M}&\cellcolor{gray!15} &\cellcolor{gray!15} &\cellcolor{gray!15}\textbf{69.83} &\cellcolor{gray!15}\textbf{54.93} &\cellcolor{gray!15}\textbf{33.20} &\cellcolor{gray!15}\textbf{90.38} &\cellcolor{gray!15}\textbf{84.21} &\cellcolor{gray!15}\textbf{71.76} &\cellcolor{gray!15}\textbf{51.19} \\

\hline
{\multirow{12}{*}{\emph{Charades-STA Speech$^*$}~\ref{sec:dataset}}}  &VSLNet$^\dag$ &{\multirow{4}{*}{Data2vec}} &{\multirow{4}{*}{VGG}} &50.27 &38.76  &23.25 &- &- &- &35.78  \\
&MMN${\ddagger}$$^\dag$&  & & 56.16   &  42.74  & 24.14   &91.25 & 80.96 &   55.97  &39.15\\
&{\bf UniSDNet-S} & & &\underline{59.19}  &\underline{45.08}  &\underline{25.91} &\underline{92.02} &\underline{82.47} &\underline{57.34}  &\underline{41.26} \\
&\cellcolor{gray!15}{\bf UniSDNet-M} &\cellcolor{gray!15} &\cellcolor{gray!15} &\cellcolor{gray!15}\textbf{60.73} &\cellcolor{gray!15}\textbf{46.37}  &\cellcolor{gray!15}\textbf{26.72}  &\cellcolor{gray!15}\textbf{92.66} &\cellcolor{gray!15}\textbf{82.31}    &\cellcolor{gray!15}\textbf{57.66}  &\cellcolor{gray!15}\textbf{42.28} \\

\cline{2-11}
&VSLNet$^\dag$ & & &52.42 &40.70 &22.36 &- &- &- &36.91   \\

&MMN${\ddagger}$$^\dag$ &Data2vec &C3D &52.28 &39.44 &21.80 &85.24 &74.16 &48.23 &36.09 \\

&{\bf UniSDNet-S}& & &\underline{56.37} &\underline{41.85} &\underline{24.06} &\underline{86.61} &\underline{76.24} &\underline{52.39} &\underline{39.21} \\

&\cellcolor{gray!15}{\bf UniSDNet-M}&\cellcolor{gray!15} &\cellcolor{gray!15} &\cellcolor{gray!15}\textbf{58.20} &\cellcolor{gray!15}\textbf{43.66} &\cellcolor{gray!15}\textbf{25.05} &\cellcolor{gray!15}\textbf{92.23} &\cellcolor{gray!15}\textbf{82.15} &\cellcolor{gray!15}\textbf{55.86} &\cellcolor{gray!15}\textbf{40.56} \\

\cline{2-11}
&VSLNet$^\dag$ &{\multirow{4}{*}{Data2vec}} &{\multirow{4}{*}{I3D}} & 65.46 & 47.55 &28.98  & - & - & - &45.40 \\

&MMN${\ddagger}$$^\dag$&  &  &64.27  &51.75  &31.26   &{93.46}  &{85.90}  &62.69 &45.84 \\

&{\bf UniSDNet-S} & & &\underline{67.37}  &\underline{53.63}  &\underline{33.87} &{94.54} &\underline{87.45}    &\underline{67.77}  &\underline{48.13} \\

&{\bf UniSDNet-M} & & &\textbf{67.45}   & \textbf{53.82}   & \textbf{34.49}   & \textbf{94.81}   & \textbf{87.90}   & \textbf{69.30}   & \textbf{48.27} \\

\hline

{\multirow{12}{*}{\emph{TACoS Speech$^*$}~\ref{sec:dataset}}} &VSLNet$^\dag$ & & 
&29.39 &20.59 &10.92 &- &- &- &21.10   \\

&MMN${\ddagger}$$^\dag$ &Data2vec &VGG &30.12 &20.07 &\underline{11.62} &56.24 &40.64  &22.17 &21.21 \\

&{\bf UniSDNet-S}& & &\underline{38.94} &\underline{23.07} &{11.02} &\underline{68.13} &\underline{50.31} &\underline{24.97} &\underline{27.59} \\

&\cellcolor{gray!15}{\bf UniSDNet-M}&\cellcolor{gray!15} &\cellcolor{gray!15} &\cellcolor{gray!15}\textbf{40.29} &\cellcolor{gray!15}\textbf{26.34} &\cellcolor{gray!15}\textbf{12.85} &\cellcolor{gray!15}\textbf{67.36} &\cellcolor{gray!15}\textbf{51.41} &\cellcolor{gray!15}\textbf{26.24} &\cellcolor{gray!15}\textbf{28.40} \\

\cline{2-11}
&VSLNet$^\dag$ &{\multirow{4}{*}{Data2vec}} &{\multirow{4}{*}{C3D}} & {38.14}	& {27.87}	&16.35& - & - & - & {27.28} \\

 &MMN${\ddagger}$$^\dag$  &  &  &   31.72    &   23.82 &12.55   & {59.16}   & {45.36} & {22.89} &  22.58 \\

&{\bf UniSDNet-S}&&&\underline{47.04}   &\underline{31.77}  &\underline{17.42} &\underline{73.78}  &\underline{60.88} &\underline{32.69} &\underline{33.25}\\

&{\bf UniSDNet-M}&& &\textbf{51.66}	&\textbf{37.77}	&\textbf{20.44}	&\textbf{76.38}	&\textbf{63.48}	&\textbf{33.64}	&\textbf{36.86}
\\ 

\cline{2-11}
&VSLNet$^\dag$ & &  
&30.54 &18.87 &10.67 &- &- &- &19.88 \\

&MMN${\ddagger}$$^\dag$ &Data2vec &I3D &29.39 &20.37 &10.82 &54.46 &42.41 &21.14 &20.86  \\

&{\bf UniSDNet-S}& & &\underline{40.11} &\underline{25.19} &\underline{11.37} &\underline{67.58}  &\underline{50.36} &\underline{24.62} &\underline{27.93} \\

&\cellcolor{gray!15}{\bf UniSDNet-M}&\cellcolor{gray!15} &\cellcolor{gray!15} &\cellcolor{gray!15}\textbf{41.74} &\cellcolor{gray!15}\textbf{26.34} &\cellcolor{gray!15}\textbf{12.25} &\cellcolor{gray!15}\textbf{69.26} &\cellcolor{gray!15}\textbf{51.26} &\cellcolor{gray!15}\textbf{24.94} &\cellcolor{gray!15}\textbf{29.27} \\

\hline
\end{tabular}
}

\label{tab:slvg_main}
\end{table*}

\textbf{Implementation Details.}
For a fair comparison, we utilize the same video features provided by 2D-TAN~\cite{zhang2020learning}, which includes 500-dim C3D feature~\cite{tran2015learning} on ActivityNet Captions, 4096-dim VGG feature~\cite{simonyan2014very} on Charades-STA, and 500-dim C3D feature on TACoS from~\cite{liu2020jointly}. 
Besides, there are currently other popular C3D feature and I3D feature~\cite{carreira2017quo} available on Charades-STA, so we also use the 4096-dim C3D feature from~\cite{zeng2020dense} and 1024-dim I3D feature provided by~\cite{mun2020local}. 
Following previous work~\cite{wang2022negative}, we use the GloVe~\cite{pennington2014glove} and BERT~\cite{sanh2019distilbert} to extract textual feature. 
For the audio feature, we use the HuggingFace~\cite{wolf2019huggingface} implementation of Data2vec~\cite{baevski2022data2vec} with pre-trained model ``facebook/data2vec-audio-base-960h'' for SLVG. 
Specifically, we set the audio sampling rate to 16,000 Hz, and use the python audio standard library ``librosa'' to read the original audio and input it into the Data2vec model to obtain the audio sequence embedding. 
Additionally, we use LayerNorm and AvgPool operations to aggregate the entire audio representation. The feature dimensions of both text and audio are 768.

\textbf{Training and Inference Settings.} 
In this work, we delve into both single-query and multi-query training. For the $M$-query annotations $\mathcal{Q}=\{q_i\}^M_{i=1}$ associated with video $\mathcal{V}$, we specify the number of queries fed into model training at a time to be $m$. When $m=1$, this corresponds to single query training, designated as \textbf{UniSDNet-S}. 
Conversely, for multi-query training, where $m>1$, specifically when $m = M$, all queries relating to video $\mathcal{V}$ are simultaneously fed into the model, referred to as \textbf{UniSDNet-M}. 
It is important to underscore that during the inference phase, regardless of UniSDNet-S or UniSDNet-M, \emph{the evaluation process is a fair single query input} that determines the prediction of a uniquely corresponding moment, consistent with the conventional settings of the NLVG \& SLVG tasks~\cite{gao2017tall,mun2020local,li2021proposal}.

We use the AdamW~\cite{loshchilov2017decoupled} to optimize the proposed model. For ActivityNet Captions and TACoS datasets, the learning rate and batch size are set to $8\times 10^{-4}$ and 12, respectively. For Charades-STA dataset, we set the learning rate and batch size to $1\times 10^{-4}$, and 48, respectively. We train the model (whether UniSDNet-S or UniSDNet-M) with the upper-limit of 15 epochs on ActivityNet Captions and Charades-STA datasets and 200 epochs on TACoS. All experiments are conducted with a GeForce RTX 2080Ti GPU.

\subsection{Comparison with state-of-the-arts for NLVG Task}
\label{sec: nlvg}
We compare our UniSDNet with the state-of-the-art methods for \textbf{\emph{NLVG}} and divide them into two groups. 
\textbf{1) Proposal-free methods}: VSLNet~\cite{zhang2020span}, LGI~\cite{mun2020local}, DRN~\cite{zeng2020dense}, CPNet~\cite{li2021proposal}, VSLNet-L~\cite{zhang2021natural}, BPNet~\cite{xiao2021boundary}, VGCL~\cite{xia2022video}, 
METML~\cite{rodriguez2023memory}, 
MA3SRN~\cite{liu2023exploring}. 
\textbf{2) Proposal-based methods}: 2D-TAN~\cite{zhang2020learning}, CSMGAN~\cite{liu2020jointly},  MS-2D-TAN~\cite{zhang2021multi}, MSAT~\cite{zhang2021multi_msat}, RaNet~\cite{gao2021relation}, I$^2$N~\cite{ning2021interaction}, FVMR~\cite{gao2021fast}, SCDM~\cite{yuan2022semantic}, MMN~\cite{wang2022negative}, MGPN~\cite{sun2022you}, SPL~\cite{liu2022skimming}, DCLN~\cite{zhang2022dual}, CPL~\cite{zheng2022weakly}, PTRM~\cite{zheng2023phrase}, CRaNet~\cite{sun2023video}, PLN~\cite{zheng2023progressive},  M$^2$DCapsN~\cite{liu2023m}, DFM~\cite{mm_dfm}. 
The best and second-best results are marked in \textbf{bold} and \underline{underlined} in experimental tables. 
{
The detailed test results of $R@1, IoU@$\{0.3, 0.5, 0.7\} on three NLVG datasets are reported in Table~\ref{tab:nlvg_activity} and Table~\ref{tab:sta_main_new}. 
{Since most works do not report $R@1, IoU@0.1$ performance, we have removed it from the table. Notably, our method performs well on all metrics on the three NLVG datasets.}
For more prediction 
distributions of our model and other existing methods on NLVG task, see Appendix~A.
}

\subsubsection{Results on the ActivityNet Captions dataset} 
The ActivityNet Captions is the largest open domain dataset for NLVG. 
As shown in Table~\ref{tab:nlvg_activity}, our UniSDNet-S has achieved satisfactory performance to current SOTA methods, but at a very low cost of 0.53M for static modules and 0.68M for dynamic modules (Table~\ref{tab:config}). If the -M (multi-query) mode is utilized in training, there will be a significant increase in performance ({UniSDNet-M} achieves the best performance with scores of 38.88 and 55.47 in terms of $R@1, IoU@0.7$, and $mIoU$, respectively), note that regardless of UniSDNet-S and -M, they are tested in the same fair way, \ie, single-query reasoning at a time. 
And a lot of work has also released M-query training modes such as MMN~\cite{wang2022negative}, PTRM~\cite{zheng2023phrase} and DFM~\cite{wang2023ms}, but their performances are significantly worse than these of our UniSDNet-M due to our efficient modelling of multimodal information. 
Since we adopt a proposal-based backend to favor modal alignment between the video moment and the query, we prefer to compare our method with recently proposed proposal-based methods, especially MMN~\cite{wang2022negative}, PTRM~\cite{zheng2023phrase}, \etc. 
And our research on recent NLVG work has found that proposal-based methods predominate, as shown in Table~\ref{tab:nlvg_activity}. 
Compared to other proposal-based methods, our {UniSDNet-M} performs the best and has substantial improvements in all metrics due to the unique static and dynamic modes.

\subsubsection{Results on the TACoS dataset}
TACoS (Cooking dataset) has the longest video length (approx. 5 min) and the highest number of events ($>$100) per video (more details in Table~\ref{tab:dataset_nlvg}). 
As shown in Table~\ref{tab:nlvg_activity}, 
the proposed UniSDNet-S (BERT) performs well with $R@1, IoU@0.3$ being 53.46, and {UniSDNet-M} achieves the best results across all metrics (\eg, 38.88 on $mIoU$), indicating that our model is better able to construct multi-query multimodal environmental semantics for video understanding. For proposal-based method MSAT~\cite{zhang2021multi_msat} with good performance of 37.57 on $R@1, IoU@0.5$. It focuses only on static feature interactions with a transformer encoder. 
In contrast, our UniSDNet-M uses the lightweight MLP- and dynamic GCN-based network to construct deeper cross-modal associations, and performs better than MSAT, achieving improvements of 6.77 and 2.69  in $R@1, IoU@0.3$ and $R@1, IoU@0.5$ metrics, respectively.

\subsubsection{Results on the Charades-STA dataset}
For the Charades-STA dataset, we report the fair comparison results of our method under VGG, C3D, and I3D  features in Table~\ref{tab:sta_main_new}. 
{ Notably, the different characteristics of Charades-STA compared to the other two NLVG datasets are analysed in Fig.~\ref{fig:dataset_nlvg}, Fig.~\ref{fig:dataset_slvg} and Section~\ref{sec:dataset_analysis}, including smallest query number size, shortest query length and shortest video duration with an average of 30.60s, so that more subtle human movements need to be identified, resulting in that the models are sensitive to different visual features.}  
Despite under this limitation, for the VGG and C3D visual features, our method achieves the best performance on the stringent metric $R@1$, \eg, 28.33 and 28.39 $R@1, IoU@0.7$ on VGG and C3D feature, respectively. For the I3D video features, our UniSDNet-M achieves an outstanding record in $R@1, IoU@0.5$ and $R@5, IoU@0.7$, that are 61.02 and 73.20, demonstrating the robustness and generalization of our model. 
Moreover, we specifically make a fair comparison of ours with MMN~\cite{wang2022negative} based on the same 2D temporal proposal map. Compared with MMN, our UniSDNet has improvements of 1.05 $\uparrow$ in \textbf{$R@1, IoU@0.7$} with VGG feature.

\subsection{Comparison with state-of-the-arts for SLVG Task}
We compare our UniSDNet with the state-of-the-art methods for SLVG, including VGCL~\cite{xia2022video}, SIL~\cite{wang2023weakly}, VSLNet~\cite{zhang2020span} and MMN~\cite{wang2022negative} methods, where VGCL and SIL both have been assessed on the \emph{ActivityNet Speech} dataset. 
In order to make fair comparison and add richer results, we reconstruct VSLNet~\cite{zhang2020span} and MMN~\cite{wang2022negative} models for the SLVG task, where VSLNet is a classic proposal-free method, and MMN is a classic proposal-based method. 

{
In addition, existing NLVG methods evaluated different video features in the experiments, including VGG, C3D, and I3D fatures. 
To validate our UniSDNet on the SLVG task dataset effect, we evaluate it using all existing available video features. Since there is no existing work reporting VGG video feature results on the ActivityNet Captions and ActivityNet Speech datasets, we followed them and do not report this result. 
And in Table~\ref{tab:slvg_main}, except for the video features presented in the implementation details, the other video features on different datasets are taken from the MS-2D-TAN~\cite{zhang2021multi}. 
The detailed test results on ActivityNet Speech dataset and our newly collected two datasets Charades-STA Speech and TACoS Speech are listed in Table~\ref{tab:slvg_main}. Our method perform the best stably under different features. See Appendix A for visualizations of the results. 
}

\subsubsection{Results on the ActivityNet Speech dataset}
The results on the ActivityNet Speech dataset are delineated in Table~\ref{tab:slvg_main}, where we evaluate a broader array of audio features, including Contrastive Predictive Coding (CPC)~\cite{oord2018representation}, Mel Spectrogram, and Data2vec~\cite{data2vec} audio features. 
This analysis aims to elucidate the variations in performance attributable to different pre-extracted audio features. It is observed that our UniSDNet-S and UniSDNet-M achieves state-of-the-art performances across all evaluated metrics (\eg, 33.29 on $R@1, IoU@0.7$). Compared to VGCL~\cite{xia2022video} and ISL~\cite{wang2023weakly}, our UniSDNet-M exhibits a remarkable enhancement, improving by approximately 20 points in $mIoU$. 
This significant gain underscores the superior efficacy of our integrated static and dynamic framework in addressing the SLVG task. 
The reconstructed VSLNet method, which utilizes Data2vec audio features, demonstrates an improvement of approximately 1 point in $R@1, IoU@0.7$ compared to the VSLNet method that uses audio Mel Spectrogram as input. When we account for the differences in input audio features and utilize the common Data2vec audio features, our UniSDNet-M outperforms VSLNet and MMN with scores of 15.84 and 12.52 on $R@1 IoU@0.7$, respectively. This highlights the effectiveness of our method in associating cross-modal information between audio and video.

\subsubsection{Results on Two New Speech datasets}
To advance research in SLVG, we conduct experiments on newly collected datasets, \emph{Charades-STA Speech} and \emph{TACos Speech}, as detailed in Section~\ref{sec:dataset} (Table~\ref{tab:dataset_slvg}) and depicted in Table~\ref{tab:slvg_main}. 
Our UniSDNet-M achieves SOTA performance across all evaluated SLVG datasets, (\eg, $R@1, IoU@0.7$ of 34.49 and 20.44 on the Charades-STA Speech and TACoS Speech, respectively). This underscores its exceptional versatility across a variety of dataset environments. 
When compared to VSLNet, our UniSDNet-M exhibits superior performance, enhancing the $mIoU$ by margins of 2.87 and 9.58 on the Charades-STA Speech and TACoS Speech datasets, respectively. 
Furthermore, in a direct comparison with the baseline model MMN, our UniSDNet-M demonstrates significantly better performance, with $mIoU$ improvements of 3.13 and 14.28 on the Charades-STA Speech and TACoS Speech datasets, respectively. These improvements further highlight the efficacy of our static and dynamic framework in bridging cross-modal information between audio and video, showcasing not only its accuracy but also its ability to effectively associate diverse modalities.

\subsection{Model Efficiency} 
\label{sec:model_eff}
\begin{figure}[t!]
\centering
\includegraphics[width=0.9\linewidth]{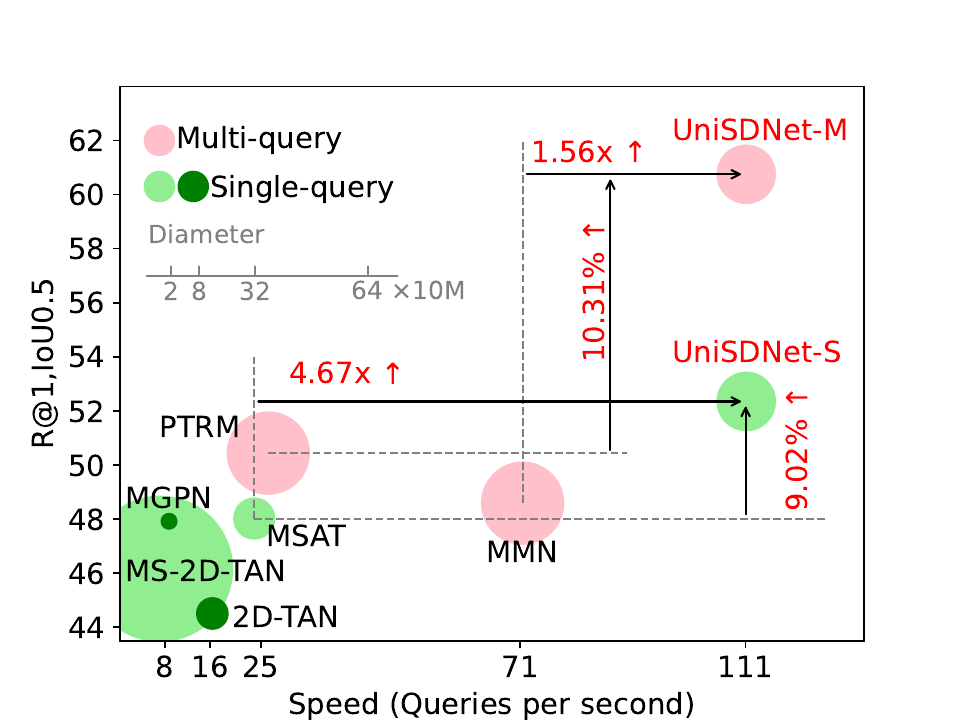}
\caption{
{Model Size \vs R@1, IoU@0.5 Accuracy Comparison of 2D Proposal-based Methods. Our UniSDNet-S has 9.02\% of performance gain on the R@1, IoU@0.5 metric while being 4.67× faster than single-query training SOTA method. Also, UniSDNet-M significantly outperforms other recent 2D proposal-based NLVG methods~\cite{zhang2020learning, zhang2021multi_msat, zhang2021multi, sun2022you, wang2022negative, zheng2023phrase} on ActivityNet Captions dataset.
UniSDNet-M achieves 10.31\% of performance gain on the $\displaystyle R@1, IoU@0.5$ metric while being 1.56$\times$ faster than multi-query training SOTA methods. 
The diameter of the circle indicates the model size (M).} 
}
\label{fig:eff}
\end{figure}

To better distinguish our model from other proposal-based models, we conduct an efficiency comparison on the ActivityNet Captions dataset in both single-query and multi-query training modes. 
The results are presented in Table~\ref{tab:efficiency}. Additionally, the specific parameters of the various modules within our UniSDNet are elaborated in Fig.~\ref{fig:eff} and Table~\ref{tab:config}. From the analysis in Table~\ref{tab:efficiency}, it is evident that our UniSDNet offers moderate parameters and exhibits the fastest inference speed 0.009 s/query, regardless of the training mode (single-query or multi-query). It is worth noting that our UniSDNet-M has only half the number of model parameters compared to the proposal-based multi-query MMN and PTRM models. Nonetheless, our UniSDNet-M achieves a remarkable 35.71\% improvement in efficiency over MMN. 
Compared to the PTRM approach that employs multi-query training, our UniSDNet exhibits notable accuracy enhancement, with an increase of 10.31\% in $R@1, IoU@0.5$. {Meanwhile, under single-query training, UniSDNet-S also has 9.02\% of performance gain on R@1, IoU@0.5 while being 4.67× faster than single-query training SOTA method.}

\begin{table*}[t] 
\centering
\caption{Ablation studies of the static (Section~\ref{sec:3SNet}) and dynamic (Section~\ref{sec:DTFNet}) modules on the \emph{ActivityNet Captions} and \emph{ActivityNet Speech} datasets.
}
\resizebox{0.85\linewidth}{!}{
\renewcommand{\arraystretch}{1}
\begin{tabular}{ccc|ccc|ccc|c}
\toprule 
{\bf Task} &{\bf Static} &{\bf Dynamic} &{\bf R@1, IoU@0.3} &{\bf R@1, IoU@0.5}&{\bf R@1, IoU@0.7}&{\bf R@5, IoU@0.3}&{\bf R@5, IoU@0.5}&{\bf R@5, IoU@0.7}&{\bf mIoU}\\ 

\midrule
{\multirow{4}{*}{{\bf NLVG}}} & \xmark& \xmark   &61.22    &44.46    &26.76  &87.19   &78.63  &  63.60&43.98\\
&\dmark & \xmark &72.32	&55.18	&31.67	&\underline{90.99}	&\underline{84.65}	&\underline{71.46}	&51.46
\\  
& \xmark & \dmark &\underline{72.74}	&\underline{55.99}	&\underline{34.17}	&90.51	&83.95	&71.42	&\underline{52.30}
\\

&\cellcolor{gray!15}{\dmark} &\cellcolor{gray!15}{\dmark} &\cellcolor{gray!15}\textbf{75.85}	&\cellcolor{gray!15}\textbf{60.75}	&\cellcolor{gray!15}\textbf{38.88}	&\cellcolor{gray!15}\textbf{91.16}	&\cellcolor{gray!15}\textbf{85.34}	&\cellcolor{gray!15}\textbf{74.01}	&\cellcolor{gray!15}\textbf{55.47}\\

\midrule

{\multirow{4}{*}{{\bf SLVG}}} & \xmark& \xmark & 53.63& 35.91  &20.51  &84.71  &   74.21   &55.95 &38.23 \\

& \dmark & \xmark &\underline{64.83} &47.82	&27.49	&\underline{90.19} &\underline{84.16}	&\underline{72.12}	&47.31\\

& \xmark & \dmark  &63.77	&\underline{49.68}	&\underline{29.32}	&89.84	&83.33	&70.30	&\underline{47.55}\\

&\cellcolor{gray!15}\dmark &\cellcolor{gray!15}\dmark &\cellcolor{gray!15}\textbf{72.27}	&\cellcolor{gray!15}\textbf{56.29}	&\cellcolor{gray!15}\textbf{33.29}	&\cellcolor{gray!15}\textbf{90.41} &\cellcolor{gray!15}\textbf{84.28} &\cellcolor{gray!15}\textbf{72.42}	&\cellcolor{gray!15}\textbf{52.22}\\

\bottomrule
\end{tabular}
}
\label{tab:ab_main_new}
\end{table*}

\begin{table}[t]
\centering
\caption{Model efficiency comparison on the \emph{ActivityNet Captions} dataset. 
``Infer. Speed'' denotes the average inference time per query.
}
\resizebox{1\linewidth}{!}{
\renewcommand{\arraystretch}{1}
\begin{tabular}{c|lrcc}
\toprule
\textbf{Query} &\textbf{Method} &\textbf{Model Size} &\textbf{Infer. Speed (s/query)}&\textbf{R@1, IoU@0.5}\\
\midrule 
{\multirow{4}{*}{{\bf Single}}} &2D-TAN~\cite{zhang2020learning}  &21.62M  
&0.061&44.51\\
&MS-2D-TAN~\cite{zhang2021multi} &479.46M &0.141&46.16\\
&MSAT~\cite{zhang2021multi_msat} &37.19M  &0.042&48.02	\\
&MGPN~\cite{sun2022you} &5.12M &0.115&47.92\\
&\cellcolor{gray!15}\textbf{UniSDNet-S (Ours)} &\cellcolor{gray!15}76.52M &\cellcolor{gray!15}\textbf{0.009} &\cellcolor{gray!15}\textbf{52.35}\\
\midrule
{\multirow{3}{*}{{\bf Multi}}}&MMN~\cite{wang2022negative}&152.22M&0.014&48.59\\
&PTRM~\cite{zheng2023phrase}&152.25M&0.038&50.44\\
&\cellcolor{gray!15}\textbf{UniSDNet-M (Ours)} &\cellcolor{gray!15}\textbf{76.52M} &\cellcolor{gray!15}\textbf{0.009}&\cellcolor{gray!15}\textbf{60.75}\\
\bottomrule
\end{tabular}}

\label{tab:efficiency}
\end{table}

\subsection{Ablation Studies} 
In this section, we conduct in-depth ablation to analyze each component and specified parameter of UniSDNet. The experiments are conducted in multi-query training mode.

\subsubsection{Ablation Study on Static and Dynamic Modules} 

We remove the static (Section~\ref{sec:3SNet}) and dynamic modules (Section~\ref{sec:DTFNet}) separately to investigate their contribution to cross-modal associativity modeling in our model. The results of NLVG and SLVG are reported in Table~\ref{tab:ab_main_new}. 
In NLVG, the single static module outperforms the baseline (without static and dynamic modules) with improvements of 4.91 and 7.48  in $R@1, IoU@0.7$ and $mIoU$, respectively. 
In addition, the single dynamic module exhibits improvements of 7.41 and 8.32 than the baseline on $R@1, IoU@0.7$ and $mIoU$, which demonstrates its effectiveness of dynamic temporal modeling in the video. 
When combining the static and dynamic modules, all the performance metrics are further improved, such as setting new SOTA records 38.88 in $R@1, IoU@0.7$ and 55.47 in $mIoU$ for NLVG. 
In SLVG, we can observe similar conclusions. These results demonstrate that both static and dynamic modules indeed have a mutual promoting effect on improving accuracy.

\subsubsection{Ablation Study on Static Network Variants}
In the static network, transformer architecture~\cite{vaswani2017attention} or the recent S4 architecture~\cite{gu2021efficiently} can also be used as long-range filter. We have tested the effect of Transformer or S4 as a static network as shown in Table~\ref{tab:ab_static}. 
From the results, in terms of performance and efficiency, Transformer is close to our method, but our results are better. We speculate that the reason is that our network also includes the second stage of graph filtering. The static network uses a lightweight and stable network (more detailed configuration in Table~\ref{tab:config}), which is more conducive to model training. Using Transformer as a static network increases the weight and instability factors~\cite{touvron2022resmlp} of the network.

\subsubsection{Dynamic Network Variants and Hyperparameters}\label{sec:ana_dtfnet}  
\textbf{Different Graph Networks.}
Our dynamic network implementation is based on the graph structure. 
We compare it with the currently popular graph structures, GCN~\cite{kipf2016semi} and GAT~\cite{velivckovic2017graph}, and test other variants of our graph filter, namely {\bf D} and {\bf MLP}. 
Additionally, our proposed temporal filtering graph contains more parametric details, which are analyzed in Section~\ref{sec:ana_dtfnet}. 
Specifically, the message aggregation definitions of these graphs are listed below: 
\begin{itemize}
\setlength{\itemsep}{0.5em} 
\item {\bf GCN:} $v_i^{(l+1)} = \sigma \left(  \sum_{j \in (v_i)} \frac{1}{\sqrt{c_i c_j}} \cdot v_j^{(l)} \right);$

\item {\bf GAT:} $v_i^{(l+1)} = \sigma \left( \sum_{j \in \mathcal{N}(v_i)} a_{ij}^{(l)} \cdot v_j^{(l)}\right);$

\item {\bf D:} $v_i^{(l+1)} = \sigma \left( \sum_{j \in \mathcal{N}(v_i)} \frac{1}{d_{ij}^{(l)}+1} \cdot v_j^{(l)}\right);$

\item {\bf MLP:} $v_i^{(l+1)} = \sigma \left( \sum_{j \in \mathcal{N}(v_i)} {\rm MLP}(d_{ij}^{(l)}) \odot v_j^{(l)}\right);$

\item {\bf Our DTFNet:} $v_i^{(l+1)} = \sigma \left( \sum_{j \in \mathcal{N}(v_i)} \Phi(d_{ij}^{(l)}) \odot v_j^{(l)}\right),$
\end{itemize}
where in these definitions, $\sigma$ is the activation function, $c_i$ is the degree of node $v_i$ for GCN, and $a_{ij}^{(l)}\in \mathbb{R}$ 
is the attention weight for GAT. 
The variant $d_{ij}^{(l)}=(1-a_{ij}^{(l)})\cdot||j-i||\in \mathbb{R}$ has been defined in Section~\ref{sec:DTFNet}, which denotes the joint clue of temporal distance and relevance between two nodes. In particular, ${\rm MLP}(d_{ij}^{(l)}) \in \mathbb{R}^h$ and $\Phi(d_{ij}^{(l)}) \in \mathbb{R}^h$ are two different ways of expanding the $d_{ij}^{(l)}$ dimension.

\begin{table}[t!]
\centering
\caption{Different static networks Comparison on \emph{ActivityNet Captions} dataset. 
}
\resizebox{0.85\linewidth}{!}{
\renewcommand{\arraystretch}{1}
\begin{tabular}{lc|ccc|c}
\toprule
{\multirow{2}{*}{{\bf Method}}}&{\multirow{1}{*}{{\bf Infer. Speed}}} &\multicolumn{3}{c|}{\bf R@1, IoU@} &{\multirow{2}{*}{{\bf mIoU}}}\\

 & (s/query)&\multicolumn{1}{c}{\bf 0.3}&\multicolumn{1}{c}{\bf 0.5} &\multicolumn{1}{c|}{\bf 0.7} &{\multirow{2}{*}{}}\\
\midrule

Transformer~\cite{vaswani2017attention} &0.024 &75.17 	&59.98 	&38.38 &54.97\\

S4~\cite{gu2021efficiently} 	&0.030 &70.41 	&55.11 	&34.93 &51.40 \\

\cellcolor{gray!15}\textbf{Our S$^3$Net} &\cellcolor{gray!15}0.009  &\cellcolor{gray!15}\textbf{75.85}&\cellcolor{gray!15}\textbf{60.75}	&\cellcolor{gray!15}\textbf{38.88} &\cellcolor{gray!15}\textbf{55.47}\\

\bottomrule
\end{tabular}
}
\label{tab:ab_static}
\end{table}

\begin{figure}[t!]
\centering
\includegraphics[width=0.85\linewidth]{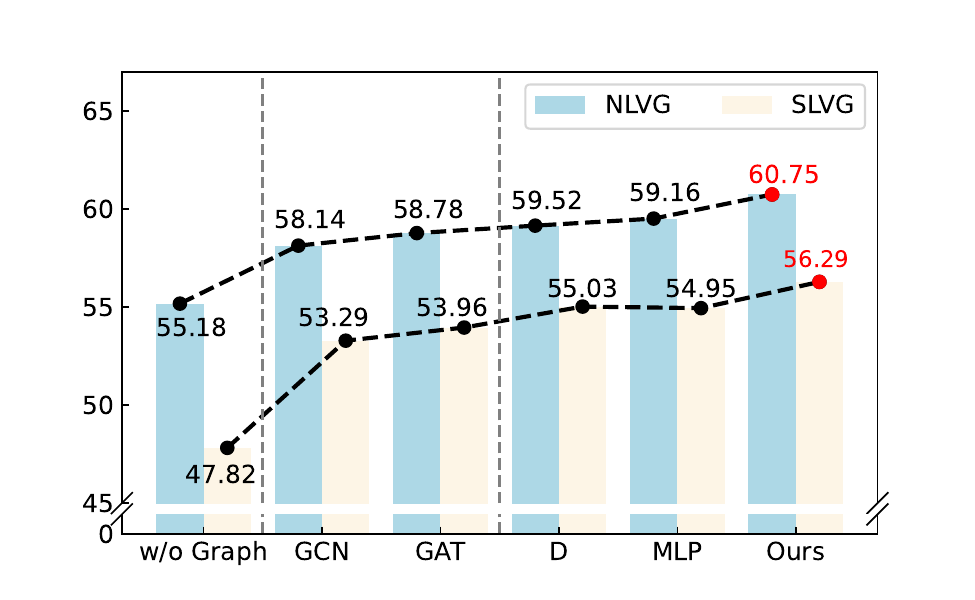}
\caption{
$R@1, IoU@0.5$ results of different message passing strategies in our Graph on \emph{ActivityNet Captions} and \emph{ActivityNet Speech} datasets.}
\label{fig:abl_gcn}
\end{figure}

\begin{table}[t!]
\centering
\caption{Different Gaussian kernel number $h$ and step $z$ on the ActivityNet Captions dataset.}
\resizebox{1\linewidth}{!}{
\renewcommand{\arraystretch}{1}
\begin{tabular}{cc|ccc|ccc|c}
\toprule
{\multirow{2}{*}{{\bf \#Kernels}}}&{\multirow{2}{*}{{\bf Step}}} &\multicolumn{3}{c|}{\bf R@1, IoU@} &\multicolumn{3}{c|}{\bf R@5, IoU@} &{\multirow{2}{*}{{\bf mIoU}}}\\

&&\multicolumn{1}{c}{\bf 0.3} &\multicolumn{1}{c}{\bf 0.5} &\multicolumn{1}{c|}{\bf 0.7} &\multicolumn{1}{c}{\bf 0.3} &\multicolumn{1}{c}{\bf 0.5} &\multicolumn{1}{c|}{\bf 0.7} &{\multirow{2}{*}{}}\\
\midrule

25&0.1 &75.12 &60.20 &38.02 &91.20 &85.82 &74.68 &54.91\\

\cellcolor{gray!15}50&\cellcolor{gray!15}0.1 &\cellcolor{gray!15} \textbf{75.62} &\cellcolor{gray!15}\textbf{60.75} &\cellcolor{gray!15}\textbf{38.88} &\cellcolor{gray!15}\textbf{90.94} &\cellcolor{gray!15}\textbf{85.34} &\cellcolor{gray!15}\textbf{74.01} &\cellcolor{gray!15}\textbf{55.47} \\
100&0.1&74.88 &59.54 &38.53 &91.29 &85.96 &74.75 &54.99\\
200&0.1&74.28 &59.60&38.62 &91.33 &85.91 &75.05 &54.96\\
\midrule
25&0.2&75.11 &60.01 &38.13 &91.25 &85.48 &74.31 &54.99\\
50&0.2&75.12 &60.31 &38.66 &90.95 &85.11 &73.86 &55.25\\
100&0.2&75.30 &59.73 &38.47 &91.65 &86.07 &75.16 &55.13\\
200&0.2&74.69 &59.99 &39.03 &91.30 &85.63 &74.86 &55.18\\

\bottomrule
\end{tabular}
}

\label{tab:ablation-gauss1} 
\end{table}

\begin{table}[t!]
\centering
\caption{Different Gaussian coefficient $\gamma$ on the ActivityNet Captions dataset.}
\resizebox{1\linewidth}{!}{
\renewcommand{\arraystretch}{1}
\begin{tabular}{c|ccc|ccc|c}
\toprule
{\multirow{2}{*}{{\bf Gaussian Coefficient}}} &\multicolumn{3}{c|}{\bf R@1, IoU@} &\multicolumn{3}{c|}{\bf R@5, IoU@} &{\multirow{2}{*}{{\bf mIoU}}}\\

&\multicolumn{1}{c}{\bf 0.3} &\multicolumn{1}{c}{\bf 0.5} &\multicolumn{1}{c|}{\bf 0.7} &\multicolumn{1}{c}{\bf 0.3} &\multicolumn{1}{c}{\bf 0.5} &\multicolumn{1}{c|}{\bf 0.7} &{\multirow{2}{*}{}}\\
\midrule

5.0 	&75.76 	&60.80 	&39.23 	&91.14 	&85.43 	&74.33 	&55.51 \\
\rowcolor{gray!15}
{\bf 10.0} 	&{\bf 75.85} 	&{\bf 60.75} 	&{\bf 38.88} 	&{\bf 91.16}	&{\bf 85.34} 	&{\bf 74.01} 	&{\bf 55.47}\\
25.0 	&75.87 	&60.77 	&39.30 	&91.16 	&85.23 	&74.06 	&55.52\\
50.0 	&75.84 	&60.98 	&38.83 	&91.04 	&85.27 	&73.98 	&55.51\\
75.0 	&75.74 	&60.57 	&38.63 	&90.98 	&85.26 	&73.86 	&55.29\\
\midrule
\rowcolor{gray!15}
average  	&75.81 	&60.77 	&38.97 	&91.10 	&85.31 	&74.05 	&55.46\\
standard deviation &0.06 	&0.15 	&0.28 	&0.08 	&0.08 	&0.17 	&0.10\\

\bottomrule
\end{tabular}
}

\label{tab:ab-lambda} 
\end{table}
\begin{table*}[t]
\centering
\caption{Comparison of different proposal generation strategies on the ActivityNet Captions and ActivityNet Speech datasets. }
\resizebox{0.8\linewidth}{!}{
\renewcommand{\arraystretch}{1}
\begin{tabular}{cccc|ccc|ccc|c}
\toprule
{\multirow{2}{*}{{\bf Task}}} &{\multirow{2}{*}{{\bf Generation}}} & {\multirow{2}{*}{{\bf Features}}} & {\multirow{2}{*}{{\bf Fusion}}} &\multicolumn{3}{c|}{\bf R@1, IoU@} &\multicolumn{3}{c|}{\bf R@5, IoU@} &{\multirow{2}{*}{{\bf mIoU}}}\\

& & & & {\bf 0.3} &\multicolumn{1}{c}{\bf 0.5} &\multicolumn{1}{c|}{\bf 0.7} &\multicolumn{1}{c}{\bf 0.3} &\multicolumn{1}{c}{\bf 0.5} &\multicolumn{1}{c|}{\bf 0.7} &{\multirow{2}{*}{}}\\

\midrule

{\multirow{6}{*}{{NLVG}}} & Conv & Content & - & 75.30  &60.27  &38.20  &90.86  &85.16  &73.17  &55.13
\\

& Conv & Content, Boundary &  Addition & 75.85 & 60.70 & 38.75 & 90.85 &85.05 &73.25 &55.41
\\
& Conv & Content, Boundary &  Concatenation & 74.76  &60.30 & 38.80  & 90.70  & 84.96 & 73.00  &55.15
\\

&MaxPool & Content & - & 75.62 & 60.40 & 38.99 &90.94 &85.22 &73.97 &55.39\\ 
&MaxPool & Content, Boundary &  Addition &\cellcolor{gray!15}\textbf{75.85} &\cellcolor{gray!15}\textbf{60.75} &\cellcolor{gray!15}\textbf{38.88} &\cellcolor{gray!15}\textbf{91.16} &\cellcolor{gray!15}\textbf{85.34} &\cellcolor{gray!15}\textbf{74.01} &\cellcolor{gray!15}\textbf{55.47}
\\
&MaxPool & Content, Boundary &  Concatenation &75.13 &59.96  &38.25  & 91.26  &85.59 &73.91 &54.98 \\

\midrule
{\multirow{6}{*}{{ SLVG}}} & Conv & Content & - & 71.02 & 55.24  & 32.88 & 90.38    &84.25  &71.38&51.66\\
& Conv & Content, Boundary &  Addition &\cellcolor{gray!15} 72.27 & \cellcolor{gray!15}56.29  &\cellcolor{gray!15} 33.29 &\cellcolor{gray!15}90.41 &\cellcolor{gray!15}84.28  &\cellcolor{gray!15}72.42 &\cellcolor{gray!15}\textbf{52.22} \\

&Conv & Content, Boundary &  Concatenation &71.45   &55.79 &33.20 &90.55  &84.16&  71.48 &51.76 \\

&MaxPool & Content & - &71.26 &55.25 &\textbf{33.74} &90.49 &84.29 &72.46 & 51.80\\

&MaxPool& Content, Boundary &  Addition&\textbf{72.60} &\textbf{56.64}  &32.61  &\textbf{90.82}  &\textbf{84.89} &\textbf{72.48} &52.04 \\

&MaxPool& Content, Boundary &  Concatenation & 69.85  &53.96  &32.05&90.36  & 84.12 &72.24 &50.68 \\

\bottomrule
\end{tabular}
}

\label{tab:slvg_ab_2d}
\end{table*}

Observing Fig.~\ref{fig:abl_gcn}, \textbf{w/o Graph} denotes the Dynamic Network is removed from the whole framework, and its performance is the worst.  
The vanilla \textbf{GCN} tracts all the neighbor nodes equally with a convolution operation to aggregate neighbor information. 
\textbf{GAT} is a weighted attention aggregation method~\cite{velivckovic2017graph}. 
Our method outperforms \textbf{GCN} and \textbf{GAT} by 2.61 and 1.97 on $R@1, IoU@0.5$ for NLVG, and by 3.00 and 2.33 on $R@1, IoU@0.5$ for SLVG, respectively. 
For \textbf{D} and \textbf{MLP}, we discuss the Gaussian filter setup in our method. 
In the setting of \textbf{D}, we directly use the message aggregation wight $f_{ij}^{(l)}=1/(d_{ij}^{(l)}+1)$ to replace $f_{ij}^{(l)}=\mathcal{F}_{filter}(d_{ij}^{(l)})$ in Eq.~\ref{eq:f}, which indicates that we still consider the same joint clue of temporal distance and relevance between two nodes $d_{ij}^{(l)}$ but remove the Gaussian filtering calculation from our method. 
This replacement results in a decrease of 1.23 and 1.26 on $R@1, IoU@0.5$ for NLVG and SLVG, respectively. 
\textbf{MLP} uses the operation ${\rm MLP}(d_{ij}^{(l)})$ to replace the Gaussian basis function $\phi (d_{ij}^{(l)})$ in Eq.~\ref{eq:f}. 
In this way, we realize the convolution kernel rather than Gaussian kernel in the dynamic filter. 
Compared to \textbf{Ours}, \textbf{MLP} has a decreased performance of 1.59 and 1.34 on $R@1, IoU@0.5$ for NLVG and SLVG, respectively. Overall, our proposed dynamic filtering network offers irreplaceable benefits.

\begin{figure}[t]
\centering
\includegraphics[width=1\linewidth]{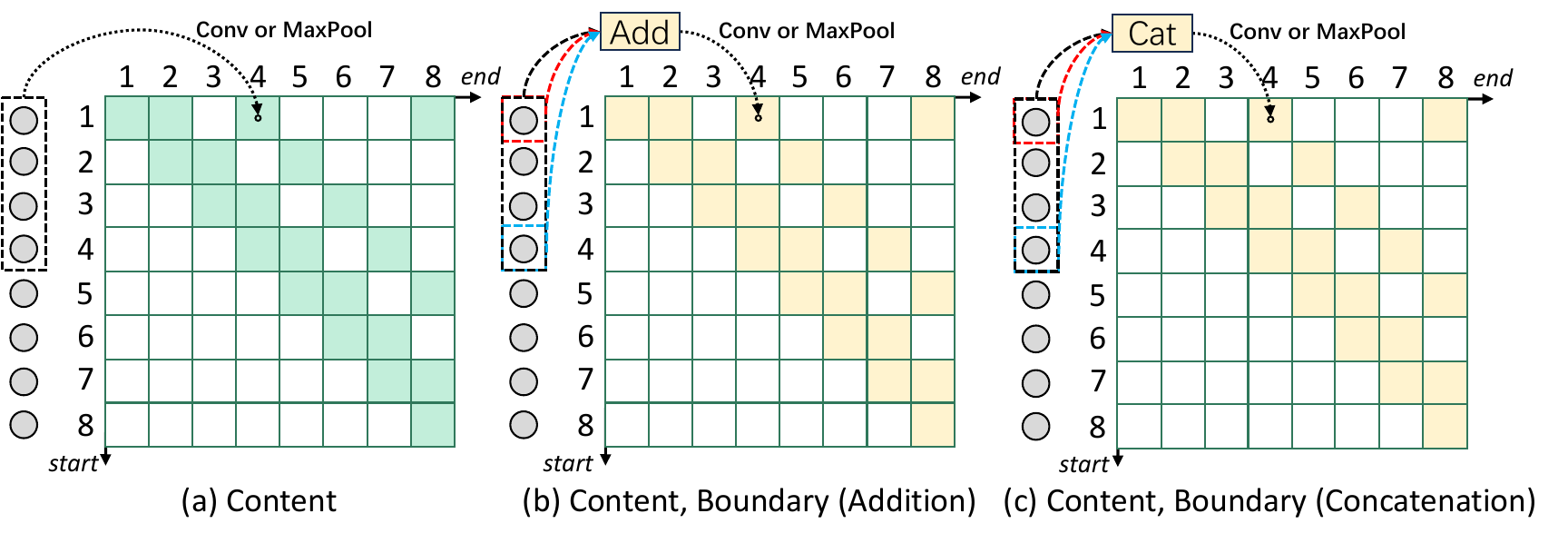}
\caption{Different feature sampling strategies for 2D proposal generation. (a) Only the content feature. (b) The content and boundary features are fused by addition operation. (c) The content and boundary features are fused with concatenation operation. }
\label{fig:boundary}
\end{figure}

\begin{figure}[t!]
\centering
\includegraphics[width=0.8\linewidth]{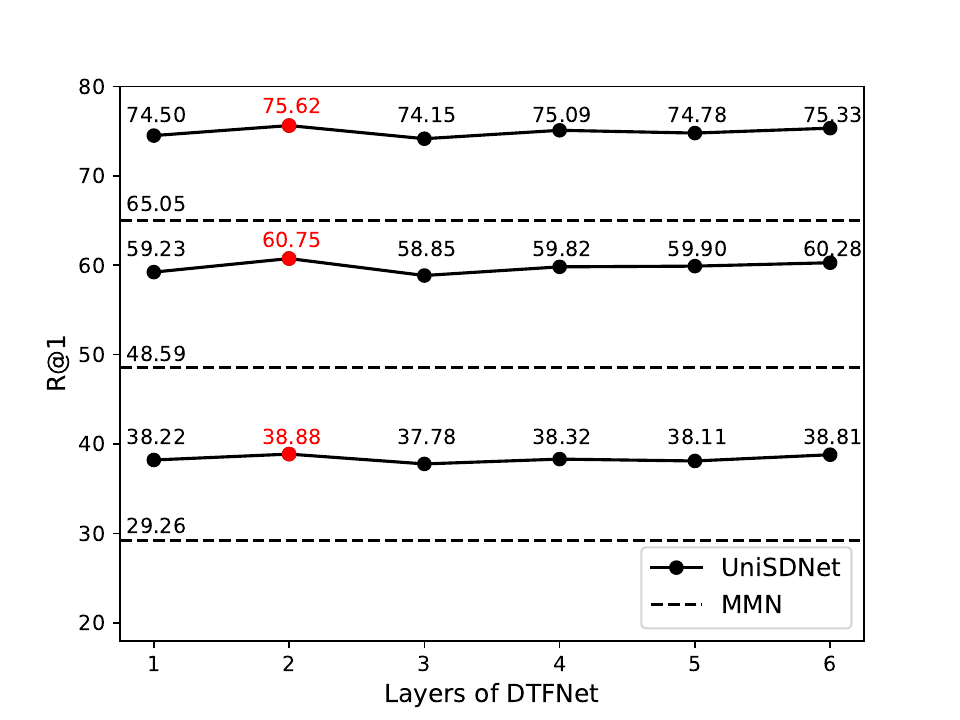}
\caption{The results across different graph layer on the AcvitivtyNet Captions dataset for NLVG. From top to bottom, the metrics are $R@1, IoU@0.3$, $IoU@0.5$, and $IoU@0.7$, respectively. 
}
\label{fig:ablation-layer}
\end{figure}

\noindent\textbf{Hyperparameters in the Dynamic Temporal Filter $\mathcal{F}_{filter}$.} In this work, we employ the multi-kernel Gaussian $\Phi(x)={\rm exp}(-\gamma(x-z_{k})^2),\,k\in[1,h]$ (Section~\ref{sec:DTFNet}), and there are three variables ($z_k,\,h,\,\gamma$): different bias $z_k$ for total $h$ Gaussian kernels and a Gaussian coefficient $\gamma$. To meet the constraint of nonlinear correlated Gaussian kernels, we randomly set biases at equal intervals (\eg, 0.1 or 0.2) starting from 0.0, sweep the value of from 25 to 200 and set the global range of $z_k$ values to $[0,\,5]$ in our experiments, as shown in Table~\ref{tab:ablation-gauss1}. And we can find that the best setting is $h=50$, we speculate that our method achieves the best results when number of Gaussian kernels $h$ is close to the number of graph nodes. Gaussian coefficient $\gamma$ reflects the amplitude of the Gaussian kernel function that controls the gradient descent speed of the function value. 
It can be found that from Table~\ref{tab:ab-lambda}, when $\gamma=25.0$, our model achieves the best performance with $mIoU$ at 55.52. 
We also list the average and standard deviation of the five experimental results and select $\gamma=10.0$ as the empirical setting as its results are closest to the average. To summarize, in our experiments, the final settings of variables ($h,\gamma$) are set to 50 and 10.0, and $z_k$ is set at an equal interval of 0.1. 

\noindent
\textbf{Dynamic Graph Layer.} 
We investigate the influence of the graph layer of our dynamic module. 
As shown in Fig.~\ref{fig:ablation-layer}, we observe that our model achieves the best result (\eg, $R@1, IoU@0.7$ is 38.88) when the total number of graph layer is set to 2. It is speculated that on the basis of informative context modelling by the static module, two-layers dynamic graph module is enough for relational learning of the video. Additionally, graph convolutional networks generally experience over-smoothing problem as the number of layers increases, leading to a performance decline~\cite{li2018deeper}. Our model exhibits good stability on the 1$\sim$6-th graph layers.

\begin{figure}[t]
\centering
\subfloat[NLVG (on ActivityNet Captions)]{\includegraphics[width=0.485\linewidth]{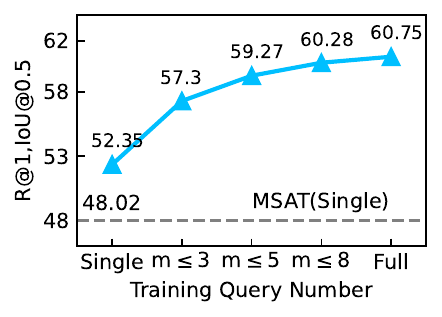}}
\hfill
\subfloat[SLVG (on ActivityNet Speech)]{\includegraphics[width=0.49\linewidth]{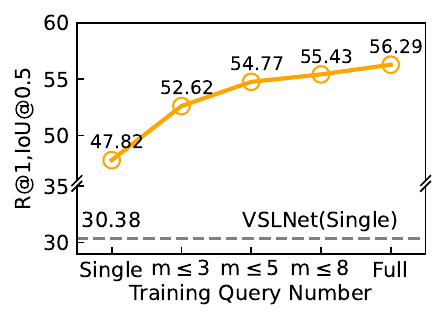}}
\caption{Results of different training query number $m$ of a video for NLVG and SLVG. Because the query number distribution of ActivityNet Captions is concentrated in the [3,8] (Fig.~\ref{fig:dataset_nlvg}), we test $m=1,3,5,8$. When $m=1$, the training mode is single-query, and when $m>1$, the training mode is multi-query. ``Full'' represents all query inputs for a video simultaneously during training. A more detailed comparison of the single-query and multi-query mode methods is given in Table~\ref{tab:efficiency}. }
\label{fig:query-number}
\end{figure}

\begin{figure*}[t]
\centering  
\includegraphics[width=1\textwidth]{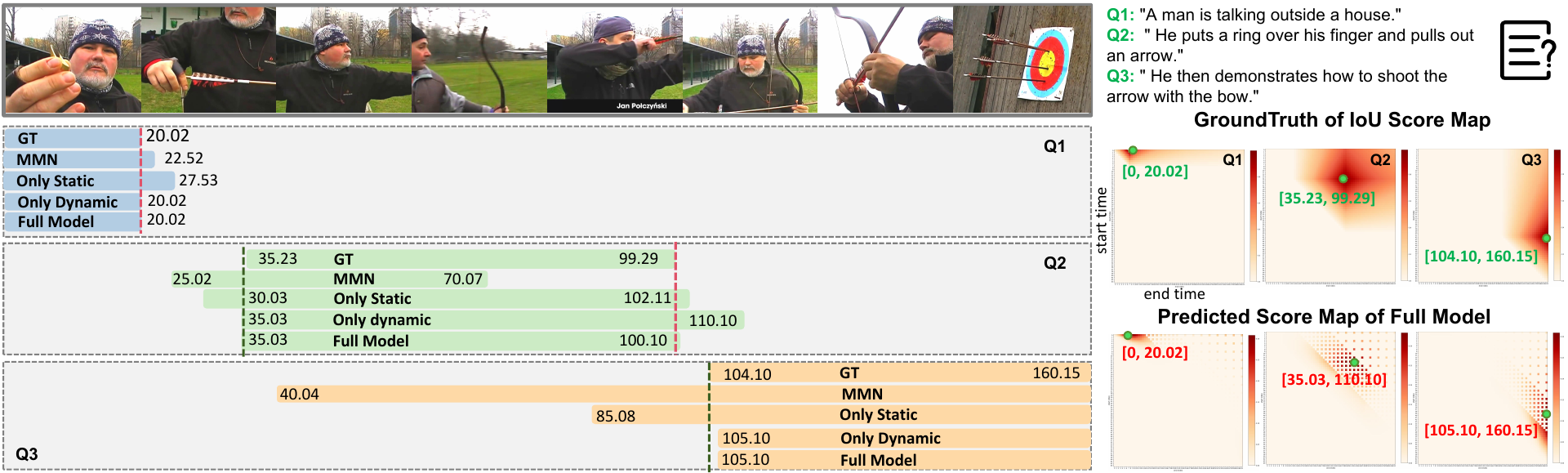}
\caption{Qualitative examples of our UniSDNet. 
The right figures display the groundtruth ${\rm IoU}$ maps and the predicted score maps by our UniSDNet. 
}
\label{fig:visual}
\end{figure*}

\subsubsection{Ablation Study on Proposals Generation}\label{sec:ana_proposal} 
To analyze the sensitivity of the feature sampling strategy for 2D proposals generation, we evaluate the effects of moment content and boundary features. 
As shown in Fig.~\ref{fig:boundary}, we conduct experiments with different proposal generation strategies:  
(a) only the content feature; (b) the addition of content and two boundary features; (c) the concatenation of content and boundary features. 
Here, the content feature refers to ${\rm Gen}(v^L_i,v^L_{i+1},\cdots,v^L_j)$ with ${\rm Gen}$ being 1D Conv or MaxPool~\cite{zhang2020learning}, where $v^L_i$ and $v^L_j$ are the start $i$-th and ending $j$-th video clip features, respectively. 
The experimental results for both NLVG and SLVG tasks are summarized in Table~\ref{tab:slvg_ab_2d}. 
For NLVG, 
the MaxPool strategy outperforms convolution, 
\eg, 38.88 \vs 38.20 in terms of $R@1, IoU@0.7$ when the model using content feature. 
Additionally, addition performs better than concatenation, 
\eg, 55.47 \vs 54.98 when the model uses the content and boundary features. 
SLVG shows similar results. Therefore, we use content and boundary features to generate proposals through MaxPool and Conv for both NLVG and SLVG.

\subsubsection{Ablation Study on Training Mode}\label{sec:ana_query}

In this work, we adopt two training mode: single-query and multi-query training, as described in experimental setup part (Section~\ref{sec:exp_set}, Training and Inference Mode). 
The number of queries is an important variable, in order to explore the effect of our UniSDNet in single-query and multi-query modes, we conduct experiments with different number of queries on the ActivityNet Captions and ActivityNet Speech datasets.
The results are shown in Fig.~\ref{fig:query-number}. 
It can be observed that for single-query training, our model is comparable with state-of-art MSAT~\cite{zhang2021multi_msat} and VSLNet~\cite{zhang2020span}, achieving scores of 52,35 and 47.82 on $R@1, IoU@0.7$ in the NLVG and SLVG tasks, respectively. As the query number upper limit increases, the performance of our model significantly improves, which demonstrates the effectiveness of our model in utilizing multimodal information.

\begin{table}[t]
	\centering
 
 \caption{Performance comparison on QVHighlights \textit{test} split. $*$: introduce audio modality. }
	\resizebox{1\linewidth}{!}
 {\renewcommand{\arraystretch}{1}
        \begin{tabular}{l|ccccccc}
        \toprule
        {\multirow{4}{*}{Method}} & \multicolumn{5}{c}{MR} & \multicolumn{2}{c}{HD} \\ \cmidrule{2-8} 
        & \multicolumn{2}{c}{R1} & \multicolumn{3}{c}{mAP} & \multicolumn{2}{c}{\textgreater{}= Very Good} \\ 
        \cmidrule{2-8} 
         &@0.5 & @0.7 & @0.5 & @0.75 & Avg. & mAP & HIT@1  \\ \midrule
        BeautyThumb~\cite{song2016click}   & - & - & -  & -  & - & 14.36 & 20.88 \\
        DVSE~\cite{liu2015multitask}  & - & - & -  & -  & - & 18.75 & 21.79 \\
        MCN~\cite{anne2017localizing}  & 11.41 & 2.72 & 24.94 & 8.22 & 10.67 & -  & - \\
        CAL~\cite{escorcia2019temporal}  & 25.49 & 11.54 & 23.40 & 7.65 & 9.89 & -  & - \\ 
        XML+~\cite{lei2020tvr} & 46.69 & 33.46 & 47.89 & 34.67 & 34.90 & 35.38 & 55.06  \\ 
        CLIP~\cite{radford2021learning}  & 16.88 &5.19 &18.11 &7.00 &7.67 &31.30 &61.04\\
        Moment-DETR~\cite{lei2021detecting} & 52.89 & 33.02 & 54.82 & 29.40 & 30.73 & 35.69 & 55.60 \\
        UMT$*$~\cite{liu2022umt}  &56.23 &41.18 &53.83 &37.01 &36.12 &38.18 &59.99\\
        MH-DETR~\cite{xu2023mhdetr}& 60.05 &42.48 &60.75 &38.13 &38.38 &38.22 &60.51\\
        QD-DETR~\cite{moon2023query}  &\underline{62.40} & \underline{44.98} &\underline{62.52} &\underline{39.88} &\underline{39.86}&\underline{38.94}       &\underline{62.40}  \\ 
        UniVTG~\cite{lin2023univtg} & 58.86 & 40.86 & 57.60 & 35.59 & 35.47 & 38.20 & 60.96 \\
        \rowcolor{gray!15}
        \textbf{UniSDNet (Ours)} &\textbf{63.49} &\textbf{46.63} &\textbf{62.86} &\textbf{42.51} &\textbf{41.33} &\textbf{39.80} &\textbf{64.66}\\
        \bottomrule
        \end{tabular}
        }
        \label{tab:qv}
\end{table}

\subsection{Extended Evaluation on QVHighlights Dataset} 
We also validate our model on the most recently publicized NLVG dataset QVHighlights~\cite{lei2021detecting} for multi-tasks:
both moment retrieval (MR, also called temporal video grounding) and highlight detection (HD) tasks. 
Following the practice~\cite{lei2021detecting,liu2022umt}, the commonly used metric for moment retrieval is Recall@K, IoU=[0.5, 0.7], and mean average precision (mAP). 
HIT@1 is also used to evaluate the highlight detection by computing the hit ratio of the highest-scored clip. The other settings such as pre-extracted Slowfast 
video and CLIP 
text features, the number of transformer decoder layers and loss weights are the same with Moment-DETR~\cite{lei2021detecting}. The comparison with exiting works are listed in Table~\ref{tab:qv}. 
From the results, our model achieves superior performance to state-of-art models, achieving $R@1, IoU@0.7$ of 63.03 for MR, and HIT@1 of 62.56 for HD, demonstrating its strong universality for both tasks.

\subsection{Qualitative Results}\label{sec:qual}

We provide the qualitative results of our UniSDNet on the ActivityNet Captions dataset with a video named ``v\_q81H-V1\_gGo'' for NLVG, as shown in Fig.~\ref{fig:visual}. MMN~\cite{wang2022negative} exhibits significant semantic bias, making it impossible to distinguish between $Q2$ and $Q3$. 
Our \textbf{Only Static} accurately predicts the moments, which is thanks to the effective static learning of the semantic association between queries and video moments. 
Our \textbf{Only Dynamic} performs well in the three queries too, thanks to the fine dynamic learning of the video sequence context. 
The results of the full model \textbf{Ours} for all queries are the closest to \textbf{Groundtruth (GT)}. 
It shows that the full model can integrate the advantageous aspects of static (differentiating different query semantics and supplementing video semantics) and dynamic (differentiating and associating the related contexts in the video) modules to achieve more accurate target moment prediction.  
The quantitative results confirm the effectiveness of our unified static and static methods in solving both NLVG and SLVG tasks. {More examples  are unfolded in Appendix~C.}

\section{Future Directions}
\label{sec:future}

As a fundamental cross-modal task, TVG research remains focusing on effectively integrating multimodal data for accurate temporal localization. Language-queried video grounding dominates current research due to advanced language models. In the future, several promising directions can advance TVG: 
First, expanding to \emph{more flexible query modes} -- incorporating audio, images, and video clips -- can enhance the model’s ability to handle diverse inputs and improve generalization. 
Second, addressing \emph{fine-grained video grounding} is essential for real-world applications, requiring detailed temporal-spatial interactions and complex scene dynamics capture, by developing larger fine-grained datasets and more sophisticated models. 
Third, \emph{long-form video understanding}, remains challenging, as current methods are typically designed for short videos struggle with extended duration content. Additionally, leveraging advances in large vision-language models (VLMs) like GPT-4V can better align visual and textual features, and explore more complementary modality information. 
Finally, improving model efficiency in computation and memory is crucial for scaling TVG systems to larger datasets and more complex scenarios.

\section{Conclusion}
\label{sec:con}
In this paper, we propose a novel Unified Static and Dynamic Network (UniSDNet) for efficient video grounding. We can adopt  
either single-query or multi-query mode and 
achieve model performance/complexity trade-offs; it benefits from both ``static'' and ``dynamic'' associations between queries and video semantics in a cross-modal environment. 
We adopt a ResMLP architecture that comprehensively considers mutual semantic supplement through video-queries interaction (static mode).
Afterwards, we utilize a dynamic Temporal Gaussian filter convolution to model nonlinear high-dimensional visual semantic perception (dynamic mode).
The static and dynamic manners complement each other, ensuring effective 2D temporary proposal generation. 
We also contribute two new Charades-STA Speech and TACoS Speech datasets for SLVG task. 
UniSDNet is evaluated on both NLVG and SLVG. For both of them we achieve new state-of-the-art results. 
We believe that our work is a new attempt and inspire similar video tasks in the design of neural networks guided by visual
perception biology.

\bibliographystyle{IEEEtran}
\bibliography{cite_paper}

\begin{IEEEbiography}[{\includegraphics[width=1in,height=1.25in,clip,keepaspectratio]{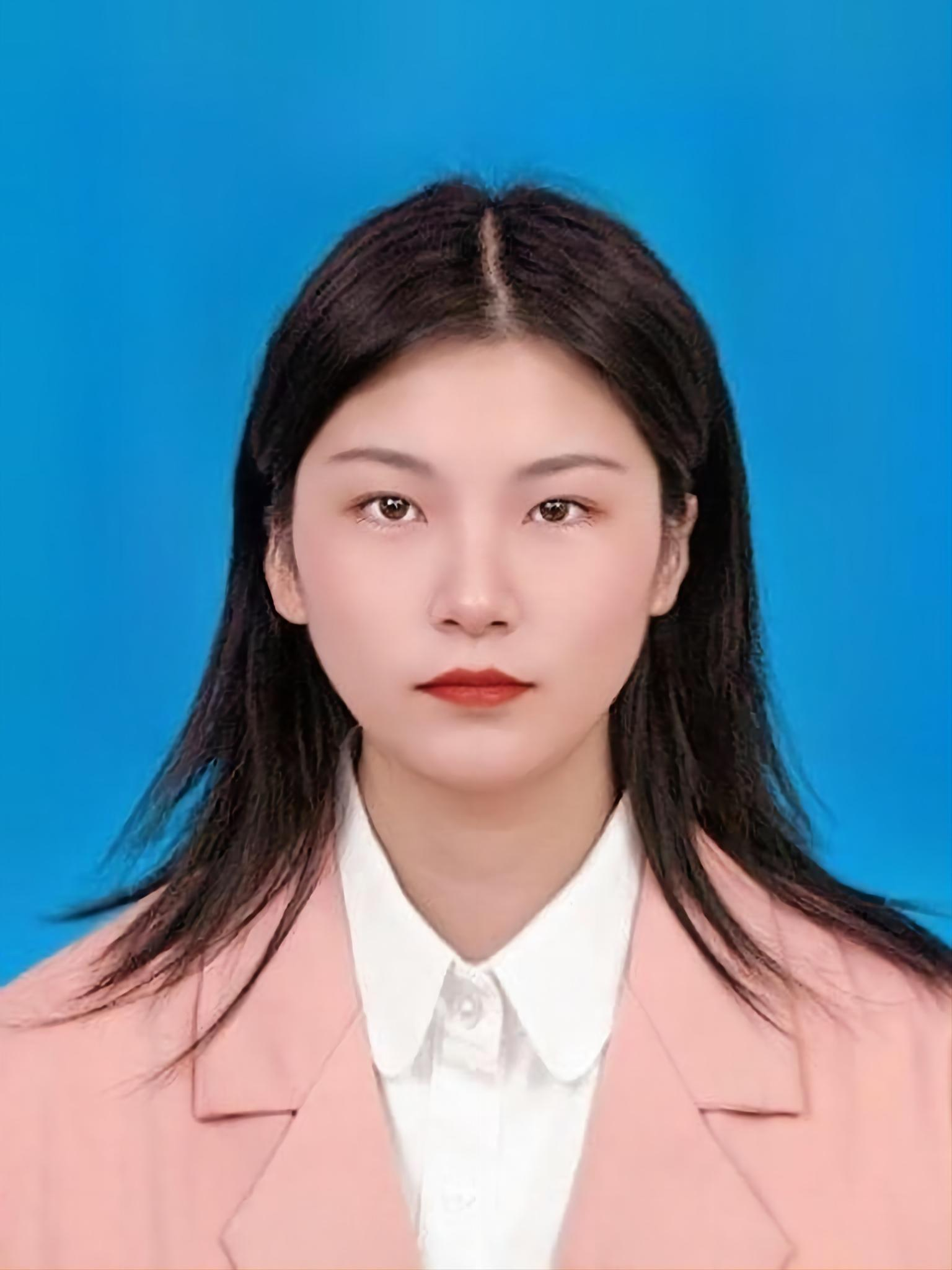}}]{Jingjing Hu} received the B.E. degree in the Internet of Things from Northeastern University, China, in 2022. She
is currently pursuing the master's degree in the School of Computer Science and Information Engineering, Hefei University of Technology, China. Her research interests include multimedia content analysis, computer vision. She currently serves as a reviewer of ACM Multimedia conference.  
\end{IEEEbiography}

\begin{IEEEbiography}[{\includegraphics[width=1in,height=1.25in,clip,keepaspectratio]{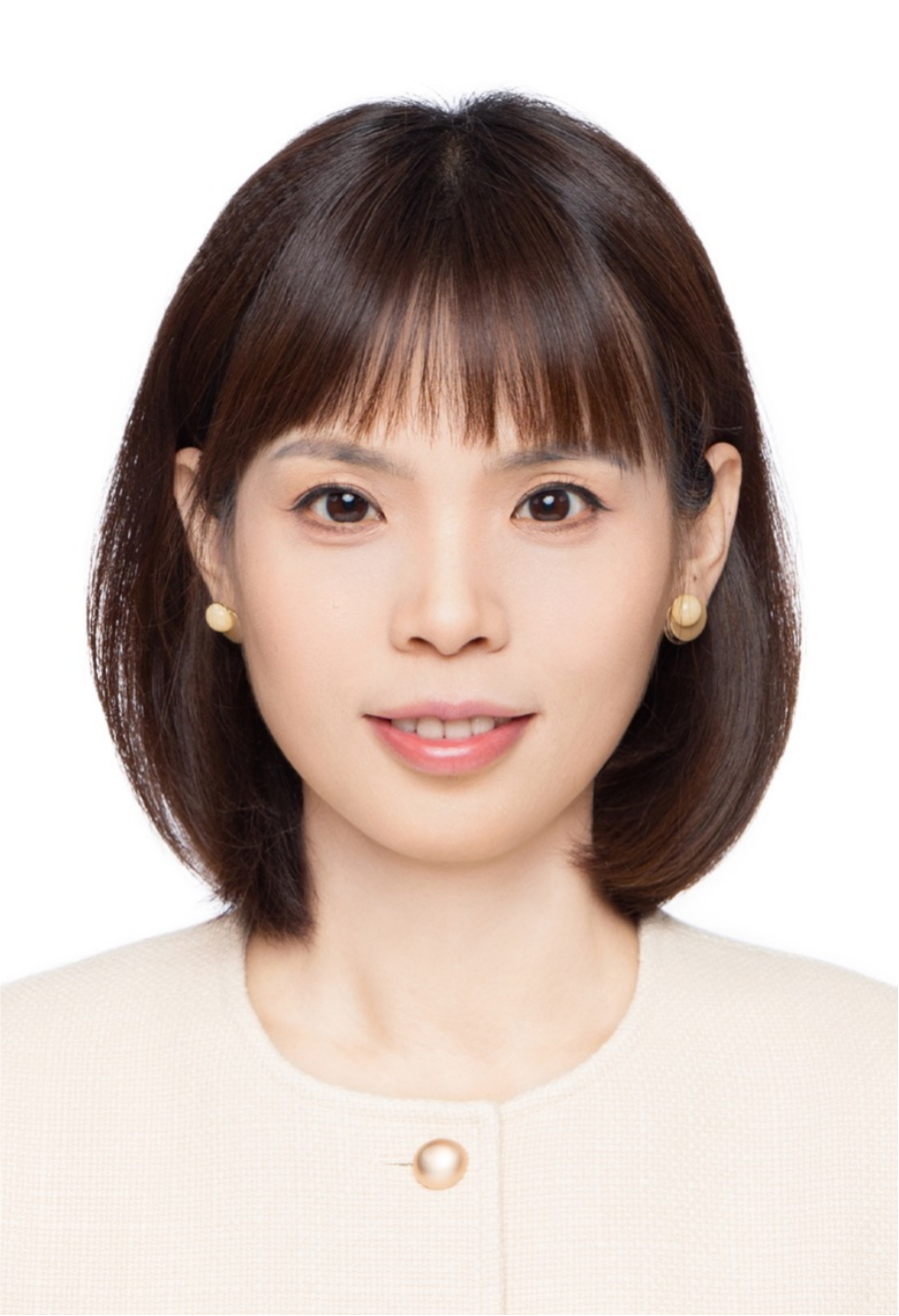}}]{Dan Guo} (Senior Member, IEEE) 
is currently a Professor with the School of Computer Science and Information Engineering, Hefei University of Technology, China.
Her research interests include computer vision, machine learning, and intelligent multimedia content analysis. She serves as a PC Member and for top-tier conferences and prestigious journals in multimedia and artificial intelligence, like ACM Multimedia, IJCAI, AAAI, CVPR and ECCV. She also serves as a SPC Member for IJCAI 2021.
\end{IEEEbiography}

\begin{IEEEbiography}[{\includegraphics[width=1in,height=1.25in,clip,keepaspectratio]{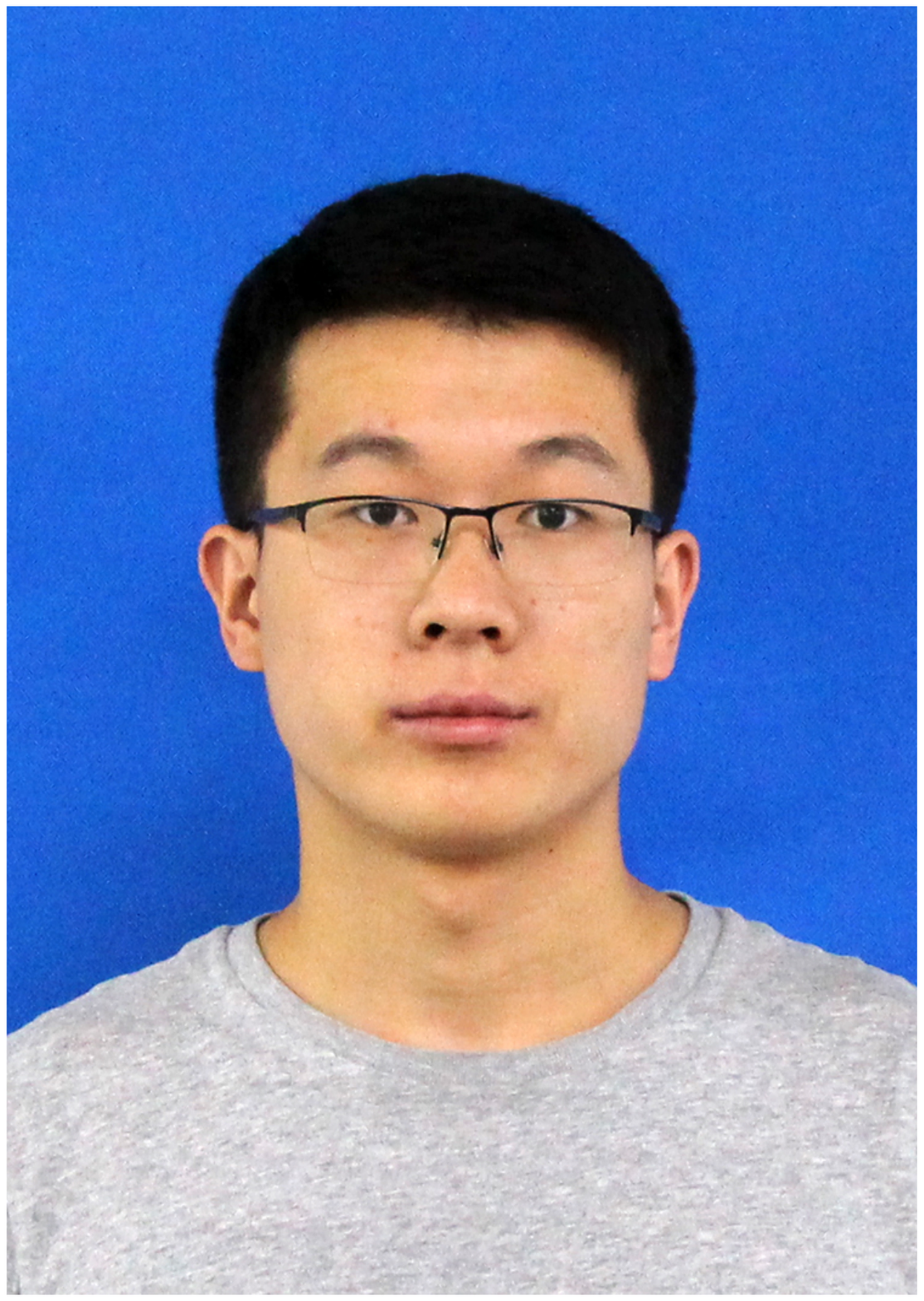}}]{Kun Li} is currently pursuing the Ph.D. degree in the School of Computer Science and Information Engineering, Hefei University of Technology, China. His research interests include multimedia content analysis, computer vision, and video understanding. He regularly serves as a PC Member for top-tier conferences in multimedia and artificial intelligence, like ACM Multimedia, IJCAI, AAAI, CVPR, ICCV, and ECCV.
\end{IEEEbiography}

\begin{IEEEbiography}[{\includegraphics[width=1in,height=1.25in,clip,keepaspectratio]{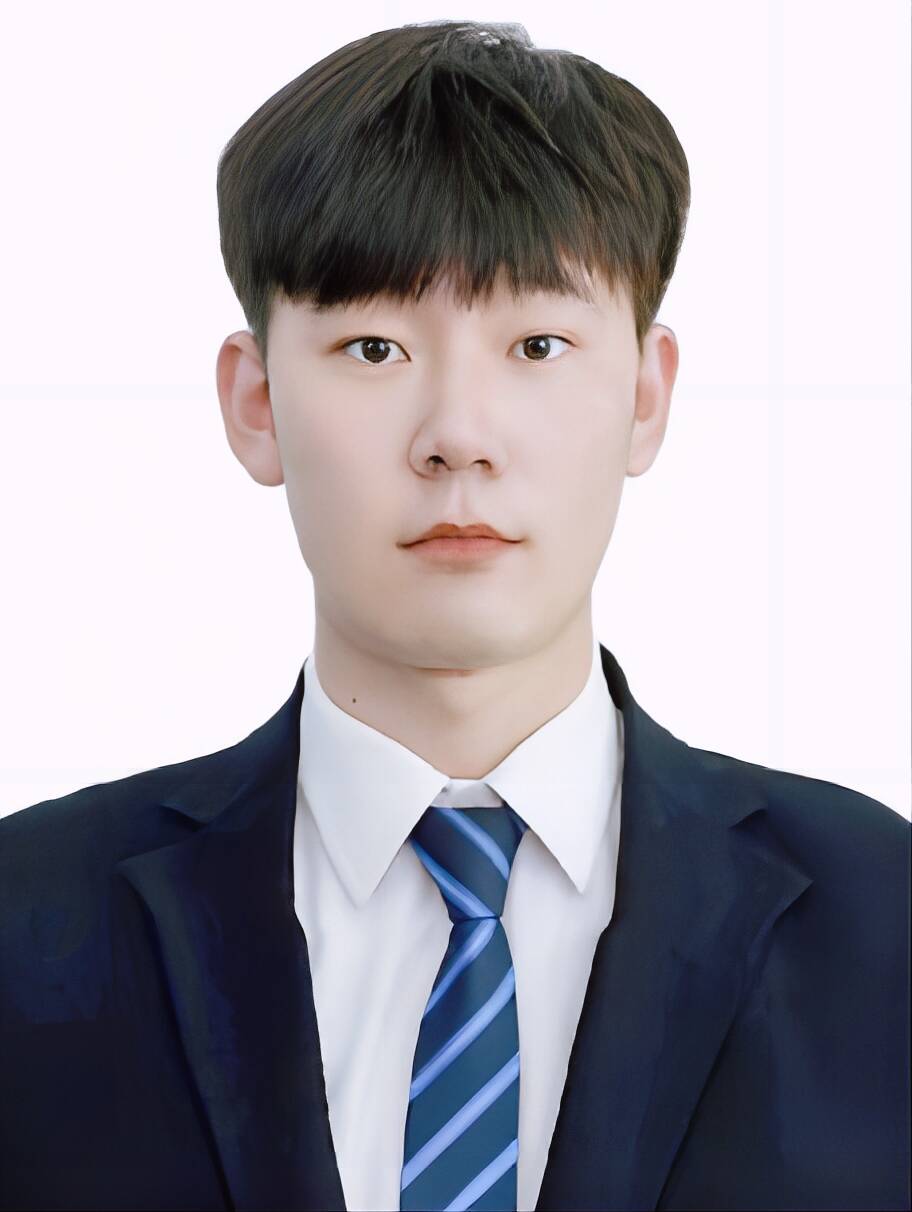}}]{Zhan Si} received the B.E. degree from Shenyang University of Chemical Technology, China, in 2022. He is currently pursuing the master's degree in the School of Chemistry and Chemical Engineering, Anhui University, China. His research interests include AI for science and computational chemistry.
\end{IEEEbiography}

\begin{IEEEbiography}[{\includegraphics[width=1in,height=1.25in,clip,keepaspectratio]{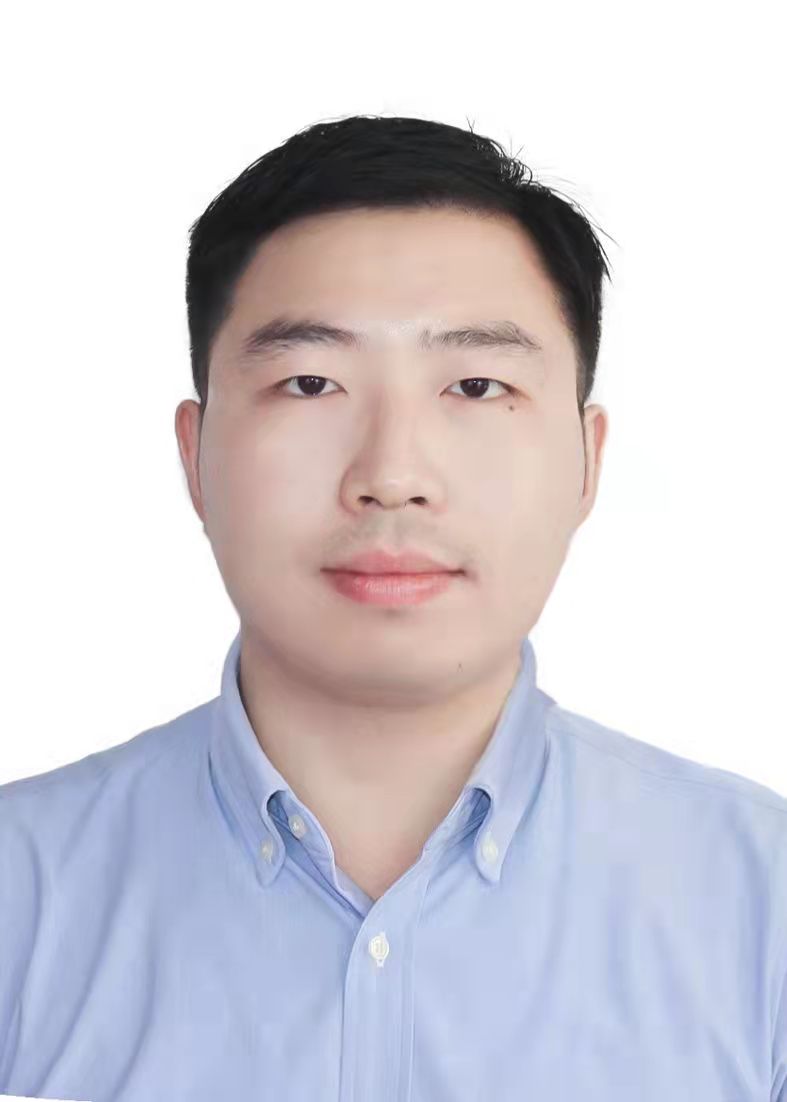}}]{Xun Yang} 
is currently a Professor with the Department of Electronic Engineering and Information Science, University of Science and Technology of China (USTC). 
His research interests include information retrieval, cross-media analysis and reasoning, and computer vision. He served as the Area Chair for the ACM Multimedia 2022. He also serves as the Associate Editor for the IEEE TRANSACTIONS ON BIG DATA journal.
\end{IEEEbiography}

\begin{IEEEbiography}[{\includegraphics[width=1in,height=1.25in,clip,keepaspectratio]{figs/bio/XiaojunChang.pdf}}]{Xiaojun Chang} (Senior Member, IEEE) is currently a Professor at the School of Information Science and Technology (USTC). 
He has spent most of his time working on exploring multiple signals (visual, acoustic, and textual) for automatic content analysis in unconstrained or surveillance videos. He has achieved top performances in various international competitions, such as TRECVID MED, TRECVID SIN, and TRECVID AVS. \end{IEEEbiography}

\begin{IEEEbiography}[{\includegraphics[width=1in,height=1.25in,clip,keepaspectratio]{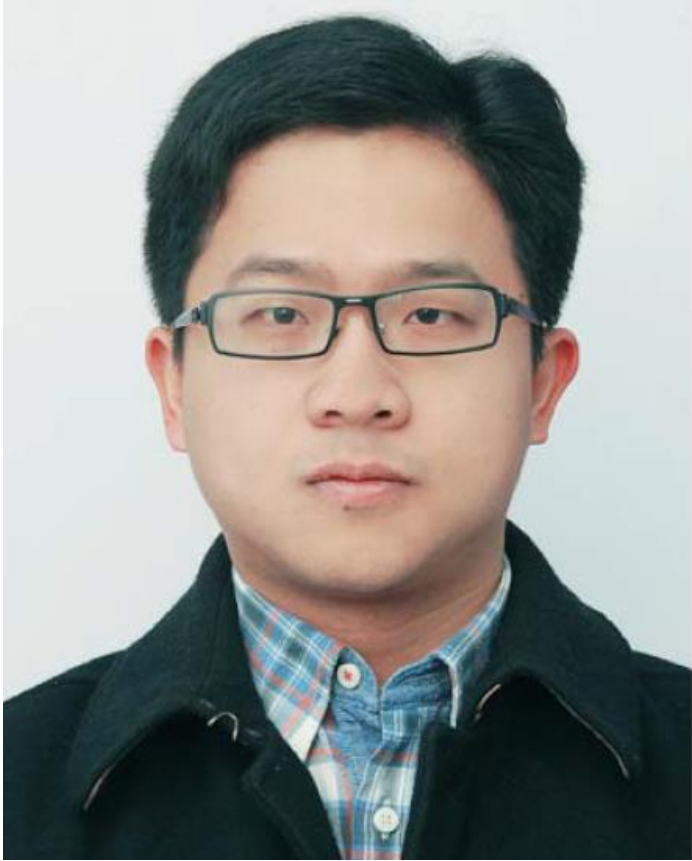}}]{Meng Wang} (Fellow, IEEE) 
is currently a Professor with the Hefei University of Technology, China. 
His current research interests include multimedia content analysis, computer vision, and pattern recognition. He was a recipient of the ACM SIGMM Rising Star Award 2014. He is an Associate Editor of the IEEE TRANSACTIONS ON KNOWLEDGE AND DATA ENGINEERING, the IEEE TRANSACTIONS ON CIRCUITS AND SYSTEMS FOR VIDEO TECHNOLOGY, and the IEEE TRANSACTIONS ON NEURAL NETWORKS AND LEARNING SYSTEMS.
\end{IEEEbiography}

\clearpage 

\onecolumn
\appendices

\section{Overall Prediction Analysis for both NLVG and SLVG Tasks} 
\label{app:overall}
\begin{figure*}[h]
  \centering

  \subfloat[Groundtruth of the target moment distribution for NLVG.]{\includegraphics[width=0.95\linewidth]{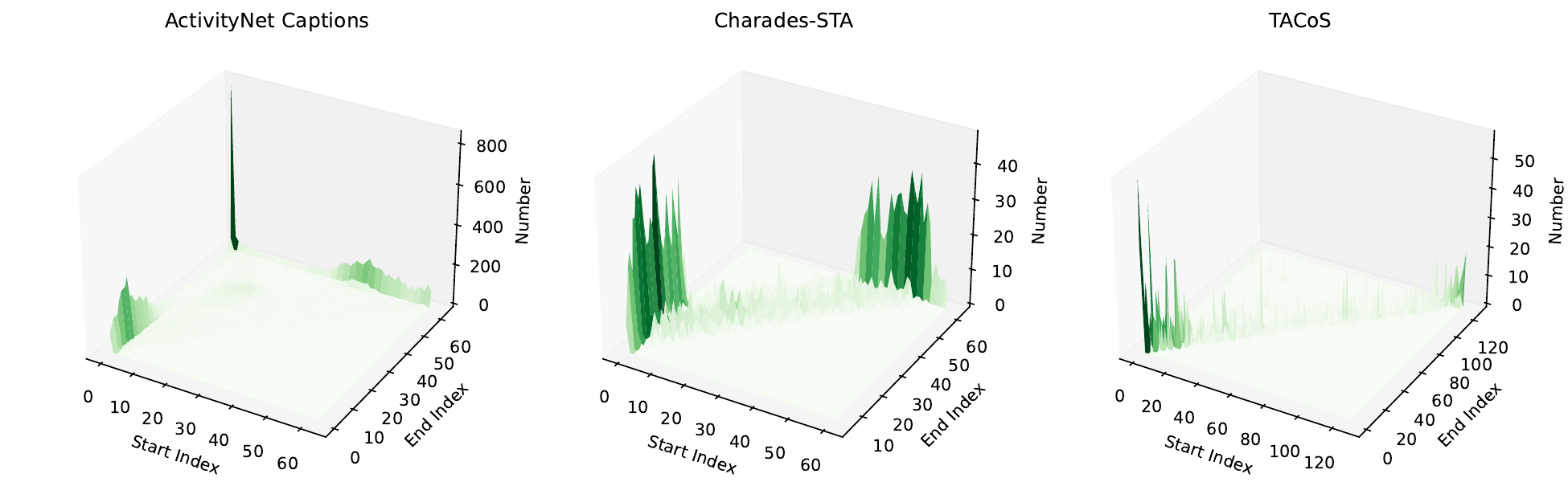}}
  \hfill
  \subfloat[Predicted moment distribution by MMN~\cite{wang2022negative} for NLVG.]{\includegraphics[width=0.95\linewidth]{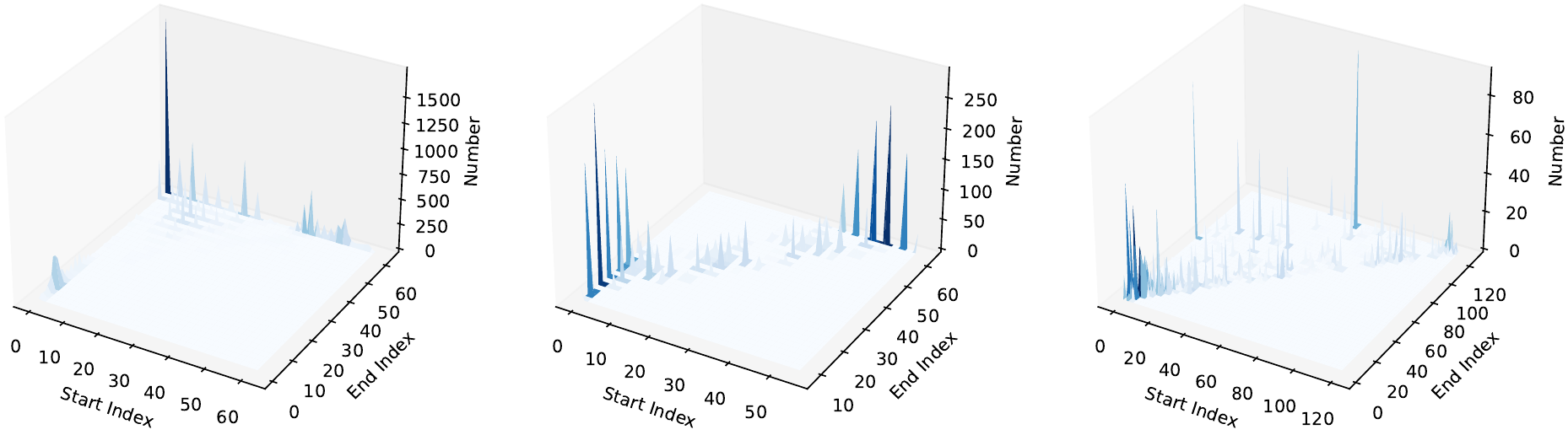}}
  \hfill 
  \subfloat[The distribution of predicted moment by our UniSDNet-M for NLVG.]{\includegraphics[width=0.95\linewidth]{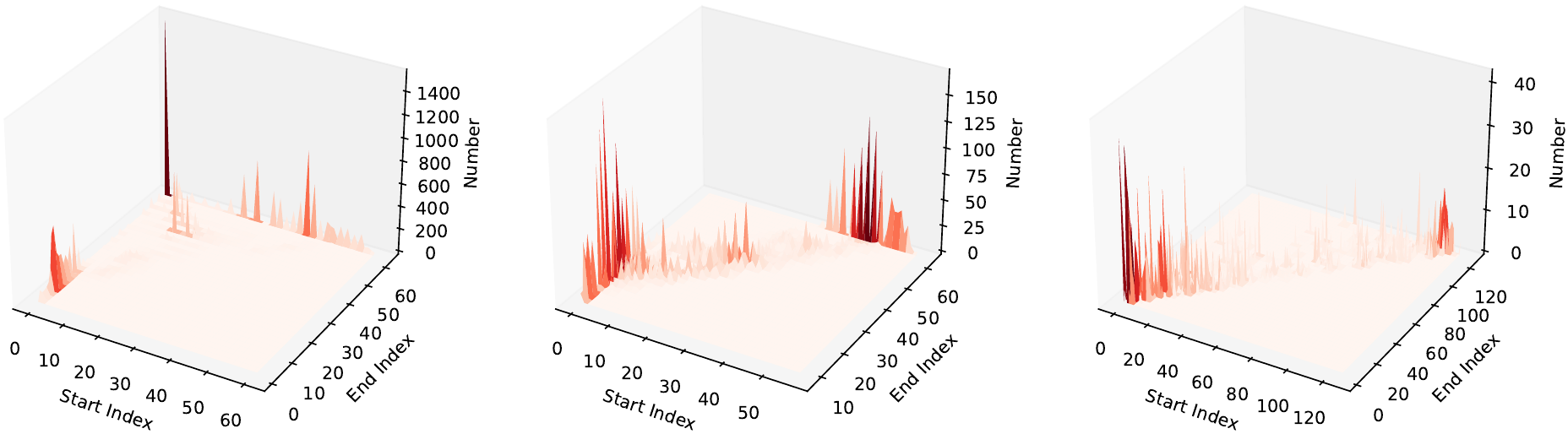}} 
  \caption{The distribution of predicted moments by our UniSDNet-M and MMN~\cite{wang2022negative} on ActivtyNet Captions, Charades-STA, and TACoS datasets for NLVG task. While MMN's predictions are more centrally biased towards regions of high density, our model fits the true distribution of the target moments to a greater extent.
  }
  \label{fig:dataset_dis_nlvg}
\end{figure*}

\begin{figure*}[t]
  \centering

  \subfloat[Groundtruth of the target moment distribution for SLVG.]{\includegraphics[width=0.95\linewidth]{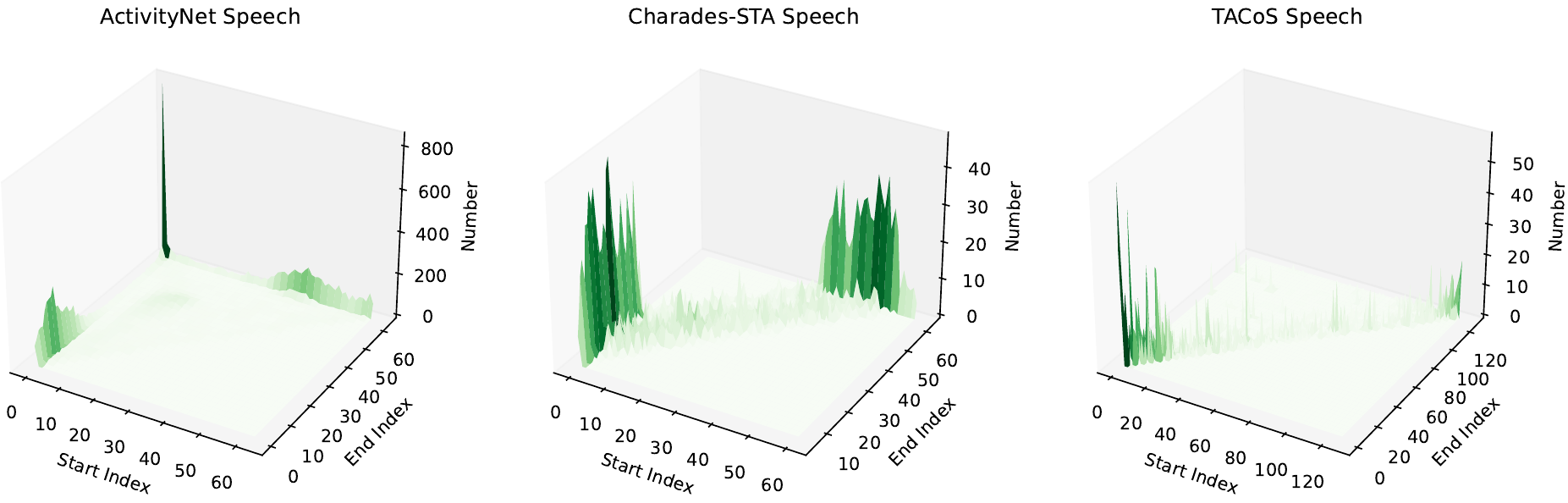}}
  \hfill
  \subfloat[Predicted moment distribution by MMN~\cite{wang2022negative} for SLVG.]{\includegraphics[width=0.95\linewidth]{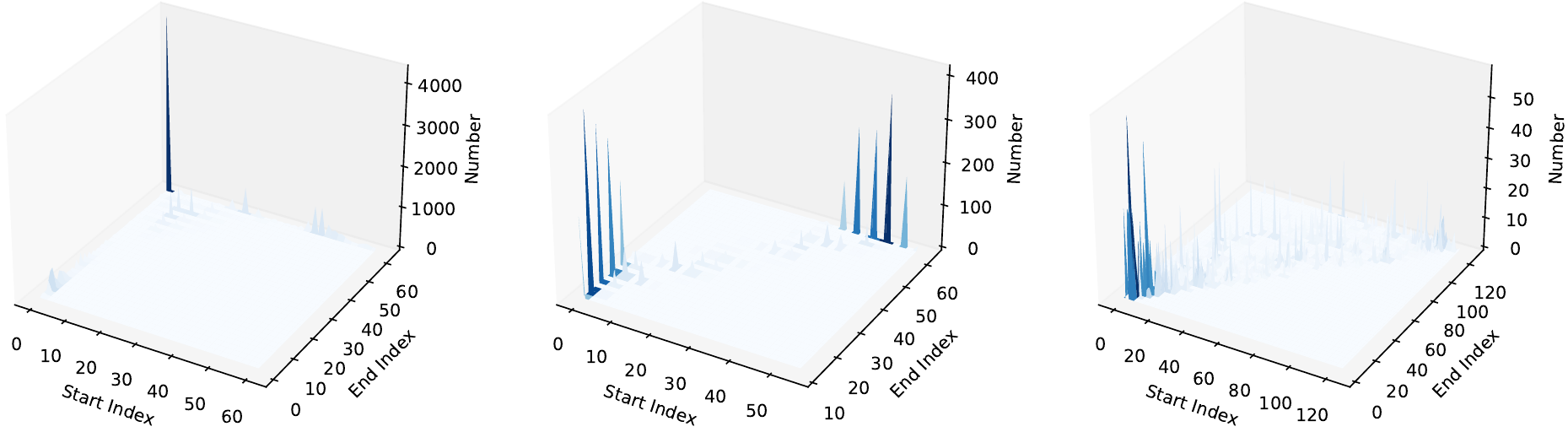}}
  \hfill 
  \subfloat[The distribution of predicted moment by our UniSDNet-M for SLVG.]{\includegraphics[width=0.95\linewidth]{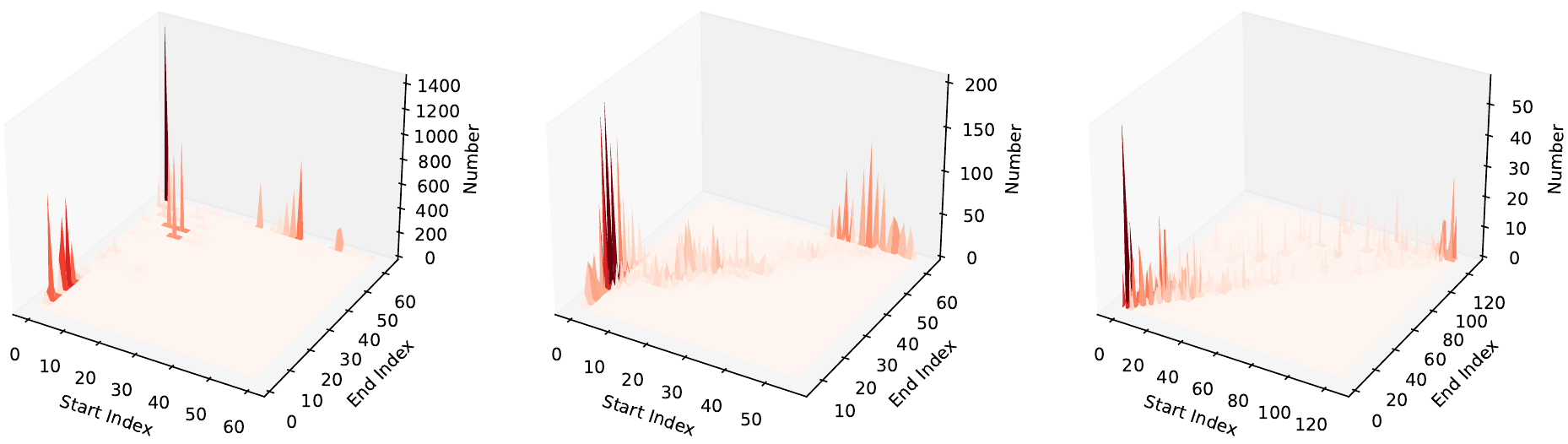}} 
  \caption{The distribution of predicted moments by our UniSDNet-M and MMN~\cite{wang2022negative} on ActivtyNet Speech, Charades-STA Speech, and TACoS Speech datasets for SLVG task. While the prediction of MMN clearly ignores some important regions, \eg, on the ActivtyNet Speech dataset, very few results correspond to the left centre and right centre of the distributions for [start index = 0, end index = 15] and [start index = 40, end index = 60], 
  our model also fits the true distribution of the target moments to a greater extent when using the spoken language as the query.
  }
  \label{fig:dataset_dis_slvg}
\end{figure*}

Fig.~\ref{fig:dataset_dis_nlvg} shows the temporal distribution of target moments on the ActivtyNet Captions, Charades-STA, and TACoS datasets for NLVG task, the distribution of target moments varies among the these datasets, and our method has good predictive performance than MMN~\cite{wang2022negative} on all these datasets, indicating that the model has good robustness. Fig.~\ref{fig:dataset_dis_slvg} shows the temporal distribution of target moments on the ActivtyNet Speech, Charades-STA Speech, and TACoS Speech datasets for SLVG task, it is clear to see that when audio is used as the query, the MMN approach is clearly missing some important moment regions in the predicted response on the ActivityNet Speech and TACoS Speech datasets, whereas our approach responds more comprehensively to the moment peak regions shown by groundtruth. 
It is worth noting that the distribution of video grounding results using text and audio as queries differ for both MMN and our UniSDNet-M. It is reasonable to assume that the predictions using text are closest to groundtruth, text-based TVG outperforms audio-based TVG due to its superior representation capability from pre-training technique.

{
\section{Additional Experimental Results}\label{app:exp}

In this section, we conduct a series of ablation studies to evaluate the hyperparameter $k$ in graph construction, as well as various model settings including the method of adding positional encodings in the feature encoding stage and the semantic matching function in the model's decoding stage.

\subsection{Ablation Study on Hyperparameters $k$ in Graph Construction}\label{app:ab_k}

The hyperparameter $k$ in Eq.~\ref{eq:ckk} of the main paper, the dividing value between short and long distances in the video graph, is a empirical parameter, which is tuned on validation set with the final model are tested on test set. Table~\ref{tab:ab_k} shows the ablation experiments on hyperparameter $k$, and Fig.~\ref{fig:adj} visualizes the video graph connectivity matrix with different $k$ value. 
From the experimental results, the optimal value of $k$ on all three datasets is 16. Either too small or too large a $k$ value can impair performance, a small \( k \) value overly focuses on short-distance information, neglecting long-distance dependencies in videos, while a large \( k \) value adds more redundant edges, increasing the difficulty for the model to recognize video events.

\begin{figure}[h]
\centering
\includegraphics[width=1\linewidth]{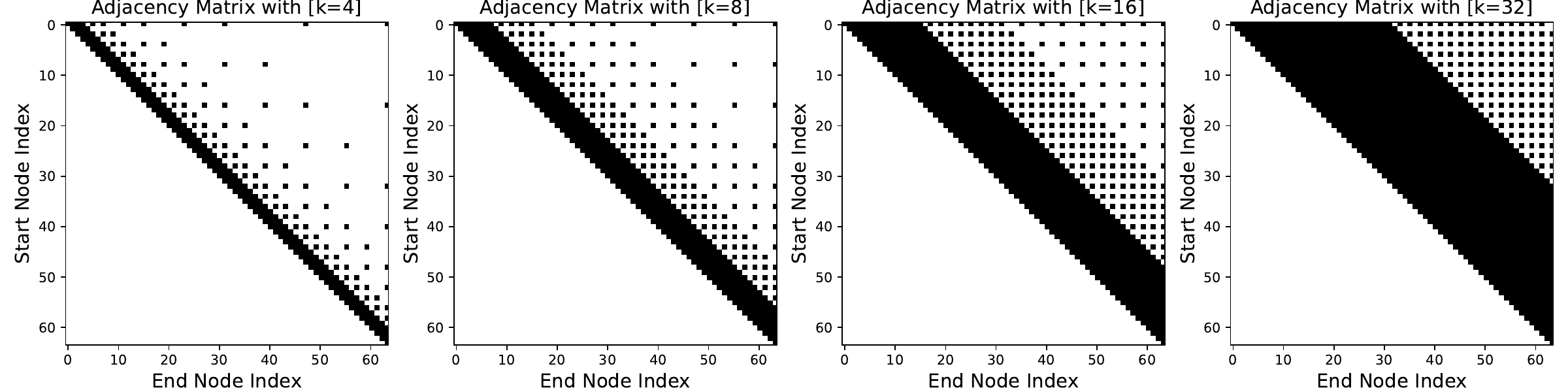}
\caption{{The adjacency matrices for different $k$ values (4, 8, 16, and 32), with the number of video clips $T$ fixed at 64 (row $i$ corresponds to node $v_i$, column $i$ corresponds to node $v_j$). Each sub-figure is a binary value $\{0,1\}$ that shows valid connections between start and end indexes in the video graph.}} 

\label{fig:adj}
\end{figure}

\begin{table*}[h]
  \caption{{Ablation study on hyperparameter $k$ for NLVG task. $L_V$ denotes the average duration of videos in a dataset. $T$ is the number of sampled clips, which is consistent with the settings in~\cite{mun2020local, zhang2020learning, liu2020jointly} for experiment fairness. Here, we use currently popular video features (ActivityNet Captions, C3D)~\cite{zhang2020learning}, (Charades-STA, I3D)~\cite{mun2020local}, (TACoS, C3D)~\cite{liu2020jointly} respectively. 
  }}
  \label{benchmark}
  \centering
\begin{adjustbox}{width=1\linewidth}
  \begin{tabular}{llcr|cccc|cccc|c}
    \hline
    \multirow{2}{*}{{Dataset}} & \multirow{2}{*}{{$L_V$}(s)} & \multirow{2}{*}{{$T$}} & \multirow{2}{*}{{$k$}} &\multicolumn{4}{c|}{\bf R@1} &\multicolumn{4}{c|}{\bf R@5} & \\
    
    & & & &{IoU@0.1} &{IoU@0.3} &{IoU@0.5} &{IoU@0.7} &{IoU@0.1} &{IoU@0.3} &{IoU@0.5} &{IoU@0.7} &{mIoU}  
    \\
\hline
 &\multirow{5}{*}{117.60} &\multirow{5}{*}{64} &4 &89.79  &73.59 &57.92   &35.50  &96.26  &90.59  &84.53 &73.20 &53.45 \\
ActivityNet & & &8  &90.12 &74.97 &60.03  &37.20 &96.26 &90.75  &84.66  &73.21 &54.53 \\
Captions  & & &\cellcolor{gray!15}{16} &\cellcolor{gray!15}{\textbf{90.28}} &\cellcolor{gray!15}{\textbf{75.85}} &\cellcolor{gray!15}{\textbf{60.75}} &\cellcolor{gray!15}{\textbf{38.88}} &\cellcolor{gray!15}{\textbf{96.32}} &\cellcolor{gray!15}{\textbf{91.17}} &\cellcolor{gray!15}{\textbf{85.34}} &\cellcolor{gray!15}{\textbf{74.01}} &\cellcolor{gray!15}{\textbf{55.47}} \\
 & & &32  &90.04   &74.87  &60.05   &36.92  &96.09  &90.72  &84.45  &72.77 &54.48 \\
 & & &64 &89.99   &74.79  &59.67  &37.17  &96.08  &90.42  &84.48 &72.17  &54.44 \\

\hline
&\multirow{5}{*}{30.60} &\multirow{5}{*}{64} &4 &78.04   &70.94 &58.15  &37.42 &98.06 &95.91 &89.60  &70.99 &50.99 \\
Charades- & & &8 &78.44  &71.29  &58.44 &38.98 &97.82 &95.81    &89.68  &72.72 &51.60 \\
STA & & &\cellcolor{gray!15}{16} &\cellcolor{gray!15}{\textbf{79.44}}   &\cellcolor{gray!15}{\textbf{72.18}}  &\cellcolor{gray!15}{\textbf{61.02}}  &\cellcolor{gray!15}{\textbf{39.70}}  &\cellcolor{gray!15}{\textbf{97.55}} &\cellcolor{gray!15}{\textbf{95.35}}  &\cellcolor{gray!15}{\textbf{89.97}}  &\cellcolor{gray!15}{\textbf{73.20}} &\cellcolor{gray!15}{\textbf{52.69}} \\
 & & &32 &78.17  &70.83   &58.23  &39.33  & 97.72   &95.97  &90.30   & 73.23  &51.67 \\
 & & &64  &78.68    &71.53    &58.76    &38.33   &98.06   &96.42     & 89.73    &71.96   &51.66 \\
 
\hline
\multirow{5}{*}{TACoS}  &\multirow{5}{*}{286.59} &\multirow{5}{*}{128} &4 &68.03  &53.34 &37.69 &21.94 &88.63 &76.63 &63.03  &35.67 &37.17 \\
& & &8 &69.78   &53.19 &38.64 &22.14 &87.83  &75.43  &63.16  &35.14   &37.84 \\
& & &\cellcolor{gray!15}16 &\cellcolor{gray!15}{\textbf{70.78}}   &\cellcolor{gray!15}{\textbf{55.56}}  &\cellcolor{gray!15}{\textbf{40.26}}  &\cellcolor{gray!15}{\textbf{24.12}} &\cellcolor{gray!15}{\textbf{89.85}}  &\cellcolor{gray!15}{\textbf{77.08}} &\cellcolor{gray!15}{\textbf{64.01}}  &\cellcolor{gray!15}{\textbf{37.02}}  &\cellcolor{gray!15}{\textbf{38.88}} 
\\
 & & &32 & 66.73 &53.06   &37.72 &21.87   &88.75  &75.88  &63.31  &35.34  &36.90 \\
 
 & & &64 & 68.93   &51.59   &34.82  &18.22  &88.58  &76.66 &63.01  &32.94   &35.50 \\ 
    \hline
  \end{tabular}
  \end{adjustbox}
\label{tab:ab_k}
\end{table*}

{
\subsection{Ablation Study on Adding Position Embeddings for Video and Query}\label{app:pos}
The function of position embedding (PE) is to help the model understand the relative position and order of different elements in the sequence, and thus better capture the semantic information in the sequence. We add sine position embedding~\cite{vaswani2017attention} to the input video clip and query sequence, in order to enhance the temporal relationships between video sequences and the logical relationships between queries. Considering that most existing multimodal Transformers add independent PEs to different modalities, in order to distinguish modality-specific information, and arcitecturally, the static module with ResMLP structure of our model is similar to Transformer in processing multi-modal sequences in parallel~\cite{touvron2022resmlp}. We simply follow existing work, adding independent PEs for video and queries, as shown in Fig.~\ref{fig:pos2}. Specifically, we denote the PE for each video clip $v_i$ or query $q_i$ as:
\begin{equation}\label{eq:pos}
PE(o_i)=\left\{
\begin{aligned}
& sin(i/10000^{j/d}),\quad \text{if $j$ is even}\\
& cos(i/10000^{j/d}),\quad \text{if $j$ is odd}
\end{aligned}
\quad ,
\right.
\end{equation}
where $PE(o_i)\in\mathbb{R}^{1\times d}$, $o_i$ denotes $v_i$ or $q_i$, 
and $j$ varies from 1 to $d$ dimension. We set up two different ways to add PE: adding Independent PE and adding Joint PE. 
These two ways of adding PE correspond to ``w/. Independent PE'' and ``w/. Joint PE'' in Table~\ref{tab:ab_pos}, respectively. 
The results demonstrate that the inclusion of PEs significantly improves the model's performance. 
On the ActivityNet Captions dataset, the $R@1, IoU@0.7$ score improved from 30.16\% to 36.96\%. Similarly, on the TACoS dataset, the $R@1, IoU@0.5$ score increased from 30.94\% to 36.84\%. 
The setting of ``w/. Independent PE'' gives better results than that of ``w/. Joint PE'', which demonstrates the superiority of adding independent PE.
}

\begin{figure}[h]
\centering
\includegraphics[width=0.9\linewidth]{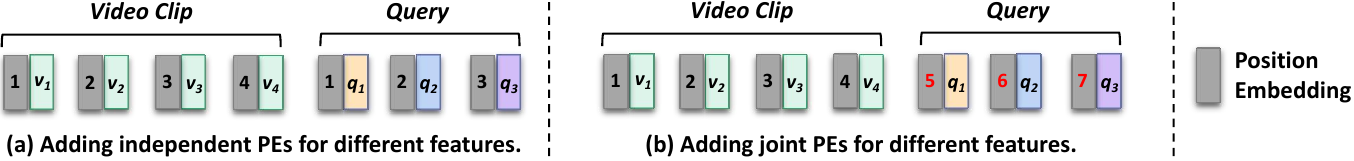}
\caption{{Illustration of two different ways to add position embedding (PE) for different modality features (query and video). }}
\label{fig:pos2}
\end{figure}

\begin{table*}[h]
\caption{{\bf Ablation Study on adding position embedding for different modality features.} The setting of ``w/o. PE'' refers to the model without any position embedding; the setting of ``w/. Joint PE'' refers to the model with the joint position encoding added to all modalities; the setting of ``w/. Independent PE'' refers to the model with independent positional embedding for different modalities (queries and video clips). 
}
\label{benchmark}
\centering
\begin{adjustbox}{width=1\linewidth}
\begin{tabular}{ll|cccc|cccc|c}
\hline
\multirow{2}{*}{\textbf{Dataset}}  & \multirow{2}{*}{\textbf{Model Setting}} &\multicolumn{4}{c|}{\bf R@1} &\multicolumn{4}{c|}{\bf R@5} & \\

&  &\textbf{IoU@0.1} &\textbf{IoU@0.3} &\textbf{IoU@0.5} &\textbf{IoU@0.7} &\textbf{IoU@0.1} &\textbf{IoU@0.3} &\textbf{IoU@0.5} &\textbf{IoU@0.7} &\textbf{mIoU}  \\
\hline
\multirow{2}{*}{ActivityNet}&w/o. PE &88.42  &70.11  &54.12  &30.16  &  95.93 &89.89  &81.43 &67.82 &49.91 \\
\multirow{2}{*}{Captions}&w/. Joint PE &89.88  &74.18  &58.29 &36.96    &96.14  &90.12  &83.67 &72.08 &53.96 \\

&\cellcolor{gray!15}w/. Independent PE &\cellcolor{gray!15} \textbf{90.28} &\cellcolor{gray!15}\textbf{75.85} &\cellcolor{gray!15}\textbf{60.75} &\cellcolor{gray!15}\textbf{38.88} &\cellcolor{gray!15}\textbf{96.32} &\cellcolor{gray!15}\textbf{91.17} &\cellcolor{gray!15}\textbf{85.34} &\cellcolor{gray!15}\textbf{74.01} &\cellcolor{gray!15}\textbf{55.47}\\  

\hline
\multirow{2}{*}{Charades-}&w/o. PE &71.99   &63.60  &50.56 &31.18  &97.61  &94.68  &86.53 &64.78 &45.09 \\
\multirow{2}{*}{STA}&w/. Joint PE &77.96   &70.81  &57.80 &36.53 &98.09  &96.08  &90.13   &71.18 &50.81 \\

&\cellcolor{gray!15}w/. Independent PE &\cellcolor{gray!15}\textbf{79.44} &\cellcolor{gray!15}\textbf{72.18}  &\cellcolor{gray!15}\textbf{61.02} &\cellcolor{gray!15}\textbf{39.70}   &\cellcolor{gray!15}\textbf{97.55} &\cellcolor{gray!15}\textbf{95.35}  &\cellcolor{gray!15}\textbf{89.97}  &\cellcolor{gray!15}\textbf{73.20}  &\cellcolor{gray!15}\textbf{52.69} \\ 

\hline
\multirow{3}{*}{TACoS}  &w/o. PE &62.48  &45.21  &30.94 &16.97 &88.30    &75.01  &58.69  &31.22  &31.99 \\
&w/. Joint PE &65.96  &50.81   &36.84 &19.70 &90.85 &79.18  &65.28 &35.02  &35.59 \\

&\cellcolor{gray!15}w/. Independent PE &\cellcolor{gray!15}\textbf{70.78} &\cellcolor{gray!15}\textbf{55.56} &\cellcolor{gray!15}\textbf{40.26} &\cellcolor{gray!15}\textbf{24.12} &\cellcolor{gray!15}\textbf{89.85} &\cellcolor{gray!15}\textbf{77.08} &\cellcolor{gray!15}\textbf{64.01} &\cellcolor{gray!15}\textbf{37.02} &\cellcolor{gray!15}\textbf{38.88}\\  

\hline
\end{tabular}
\end{adjustbox}
\label{tab:ab_pos}
\end{table*}

{
\subsection{Ablation Study on Modality Alignment Measurement Method}\label{app:ab_sim}
In this part, we investigate different cross-modal semantic similarity matching methods. The ablation study in Table~\ref{tab:ab_match} compares cosine similarity used in ``Section 3.4 2D Proposal Generation'', Eq.~\ref{eq:sim} of the main paper, with other similarity measures, including \textbf{(1) Mean Hadamard product}:
$
\text{Hada}_{\text{Mean}}({S}^{\mathcal{M}}, {S}^{\mathcal{Q}}) = \frac{1}{d} \sum_{i=1}^{d} ({S}^{\mathcal{M}}_i \odot {S}^{\mathcal{Q}}_i)
$. 
\textbf{(2) Euclidean distance} measures the straight-line distance between two vectors and is defined as:
$
\text{E-Dis}({S}^{\mathcal{M}}, {S}^{\mathcal{Q}}) = \sqrt{\sum_{i=1}^{d} ({S}^{\mathcal{M}}_i - {S}^{\mathcal{Q}}_i)^2}
$. 
\textbf{(3) Manhattan distance}, also known as L1 distance, is calculated as:
$
\text{M-Dis}({S}^{\mathcal{M}}, {S}^{\mathcal{Q}})= \sum_{i=1}^{d} |{S}^{\mathcal{M}}_i - {S}^{\mathcal{Q}}_i|
$. 
From Table~\ref{tab:ab_match}, it is evident that cosine similarity performs best, and the Hadamard product provides competitive results. 
Based on these findings, we confirm that cosine similarity is an effective measure for our semantic matching module. Nevertheless, the alternative similarity measures provide valuable insights and potential areas for further exploration.
}

\begin{table*}[h]
  \caption{{\bf Ablation Study on different similarity measure functions.} 
  We report the experimental results of similarity measures: ``cosine" $\text{CoSine}(\cdot, \cdot)$, ``Mean Hadamard product'' $\text{Hada}_{\text{Mean}}(\cdot, \cdot)$, ``Euclidean distance'' $\text{E-Dis}(\cdot, \cdot)$, and ``Manhattan distance'' $\text{M-Dis}(\cdot, \cdot)$.} 
  \label{benchmark}
  \centering
  \begin{adjustbox}{width=1\linewidth}
  \renewcommand{\arraystretch}{1.25}
  \begin{tabular}{ll|cccc|cccc|c}
    \hline
    \multirow{2}{*}{\textbf{Dataset}}  &  {\textbf{Semantic Matching}} &\multicolumn{4}{c|}{\bf R@1} &\multicolumn{4}{c|}{\bf R@5} & \\
    
    &\textbf{Measure Method} &\textbf{IoU@0.1} &\textbf{IoU@0.3} &\textbf{IoU@0.5} &\textbf{IoU@0.7} &\textbf{IoU@0.1} &\textbf{IoU@0.3} &\textbf{IoU@0.5} &\textbf{IoU@0.7} &\textbf{mIoU}  \\
\hline
\rowcolor{gray!15}
\multirow{4}{*}{ActivityNet}&$\text{CoSine}(\cdot, \cdot)$ &\textbf{90.28} &\textbf{75.85} &\textbf{60.75} &\textbf{38.88} &\textbf{96.32} &\textbf{91.17} &\textbf{85.34} &\textbf{74.01} &\textbf{55.47}\\ 
\multirow{4}{*}{Captions}&$\text{Hada}_{\rm Mean}(\cdot, \cdot)$ &89.08  &74.23 &58.89  &36.85 &95.94 &90.59  &84.73   &73.16 &54.04 \\
&$\text{E-Dis}(\cdot, \cdot)$ &89.37  &74.68 &58.40 &35.83 &95.96 &90.32  &84.16 &71.26 &53.57 \\
&$\text{M-Dis}(\cdot, \cdot)$ &89.06 &73.21   &56.68 &34.01  &96.29  &90.66  &84.45 &72.93 &52.69 \\

\hline
\rowcolor{gray!15}
\multirow{4}{*}{Charades-}&$\text{CoSine}(\cdot, \cdot)$ &\textbf{79.44} &\textbf{72.18} &\textbf{61.02} &\textbf{39.70} &\textbf{97.55} &\textbf{95.35} &\textbf{89.97} &\textbf{73.20} &\textbf{52.69}\\ 
\multirow{4}{*}{STA}&$\text{Hada}_{\rm Mean}(\cdot, \cdot)$  &78.68  &71.53  &58.76  &38.33  &98.06 &96.42  &89.73 &71.96 &51.66 \\
&$\text{E-Dis}(\cdot, \cdot)$ &78.01  &70.59 &57.28  &37.37 &98.31  &96.29 &90.27  &71.64 &50.82 \\
&$\text{M-Dis}(\cdot, \cdot)$ &77.72   &69.95  &57.47  &37.26  &97.77    &95.43  &89.25  &70.97  &50.48 \\

\hline
\rowcolor{gray!15}
\multirow{4}{*}{TACoS}&$\text{CoSine}(\cdot, \cdot)$ &\textbf{70.78} &\textbf{55.56} &\textbf{40.26} &\textbf{24.12} &\textbf{89.85} &\textbf{77.08} &\textbf{64.01} &\textbf{37.02} &\textbf{38.88} \\ 
&$\text{Hada}_{\rm Mean}(\cdot, \cdot)$ &69.51   &54.19   &38.59 &23.03 &89.65  &78.78   &64.48  &35.87 &37.60 \\
&$\text{E-Dis}(\cdot, \cdot)$  &68.03   &53.34  &37.69   &22.94   &89.63  &77.63   &64.03  &35.67  &37.17 \\
&$\text{M-Dis}(\cdot, \cdot)$ &66.58  &53.11  &37.44  &22.87  &88.78    &75.43   &62.76  &35.09  &36.90 \\

    \hline
  \end{tabular}
  \end{adjustbox}
\label{tab:ab_match}
\end{table*}

}

\section{More Visualization of Prediction Results}\label{app:visual}

In order to clearly demonstrate the specific role of our proposed unified static and dynamic networks in cross-modal video grounding, we provide more challenging visualization cases in this section as a supplement to Sec.~\ref{sec:qual}.

\subsection{Visualization on ActivityNet Captions for NLVG}

\begin{figure*}[t!]
\vspace{-0.5em}
\centering
\includegraphics[width=0.9\linewidth]{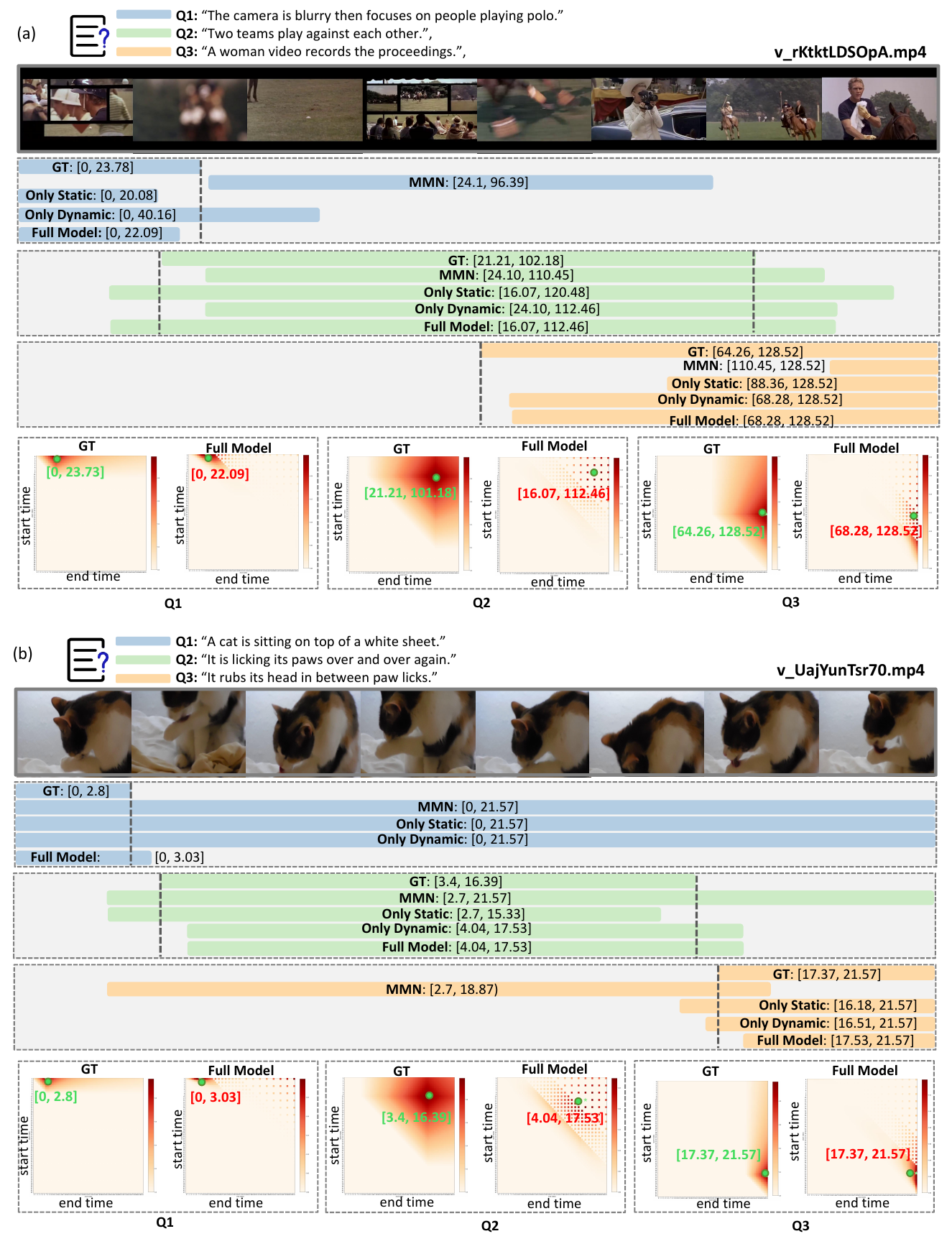}
\vspace{-1.5em}
\caption{Qualitative examples on ActivityNet Captions for NLVG.
\textbf{(a)} The video contains complex scene transitions and overlap.
\textbf{(b)} The video scenes that are difficult to distinguish.
\textbf{MMN} makes significant errors in predicting the location range of the queried events, \ie, $Q1$ and $Q3$ in cases (a) and (b), respectively.
Our \textbf{Only Static} has an advantage in predicting transitions ($Q1$ in case (a)), our \textbf{Only Dynamic} performs better in predicting overlapping. It is difficult to distinguish scenarios ($Q2$ and $Q3$ in both cases (a) and (b)).
Our \textbf{Full Model} performs best in both challenging scenarios, as it combines the advantages of static (query semantic differentiation) and dynamic (video sequence context association) modules.
}
\label{fig:app_vis_anet1}
\end{figure*}

\textbf{Video Sample with Complex Scene Transitions.} The ActivityNet Captions dataset contains a large amount of open-world videos with more shot transitions. We choose typical samples of this type for visualisation and analysis. 
As shown in Fig~\ref{fig:app_vis_anet1} (a), there are multiple scene transitions in video sample ``ID: v\_rKtktLDSOpA'' from the ActivityNet Captions dataset and different events have serious intersection in the temporal sequence of video. 
For example, there is an intersection between the end of the moment corresponding to $Q1$ and the beginning of the moment corresponding to $Q2$ and another big intersection exists between the moments corresponding to $Q2$ and $Q3$. 
From Fig.~\ref{fig:app_vis_anet1}, \textbf{MMN}~\cite{wang2022negative} makes a serious prediction for $Q1$, locating the moment corresponding to $Q2$. 
Meanwhile, when predicting $Q3$, \textbf{MMN} omits the temporal region intersected with $Q2$ but correct temporal region also belonged to the moment of $Q3$ for the final prediction. 
Compared to \textbf{MMN}, our \textbf{Only Static} and \textbf{Only Dynamic} predict more accurate moments for each query, and they can accurately comprehend the intersection of $Q2$ and $Q3$.
\textbf{Only Static} performs better at identifying transitions, while \textbf{Only Dynamic} performs better at recognizing overlapping events.
Our \textbf{Full Model} performs best in these challenging scenarios because it combines the advantages of \textbf{Only Static} and \textbf{Only Dynamic}.

\textbf{Video Sample with Similar Scenes.} For the NLVG task that employs textual queries, it is also challenging to use the semantic guidance of the text to distinguish some video clips that are similar in the front and back frames (without transitions). As shown in Fig.~\ref{fig:app_vis_anet1} (b), the frames in video sample ``ID: v\_UajYunTsr70'' from the ActivityNet Captions dataset also have high similarity, you can find it to locate the corresponding moment corresponding to $Q1$: ``\emph{A cat is sitting on top of a white sheet.}''
\textbf{MMN} is basically unable to distinguish the video content for the three different queries. It almost predicts the entire video for each query.
Even through our \textbf{Only Static} performs poorly in this situation too, our \textbf{Only Dynamic} performs much better than MMN. 
Finally, our \textbf{Full model} locates the most accurate target moment. 
This is thanks to our model that combines the advantages of static and dynamic modules, especially for that the latter learns a tighter contextual correlation of video in this case.

\subsection{Visualization on ActivityNet Speech for SLVG.}
We also provide quantitative results of our UniSDNet on SLVG to demonstrate the effectiveness of our model in the video grounding task based on spoken language.

\begin{figure*}[t!] 
\centering
\includegraphics[width=0.9\linewidth]{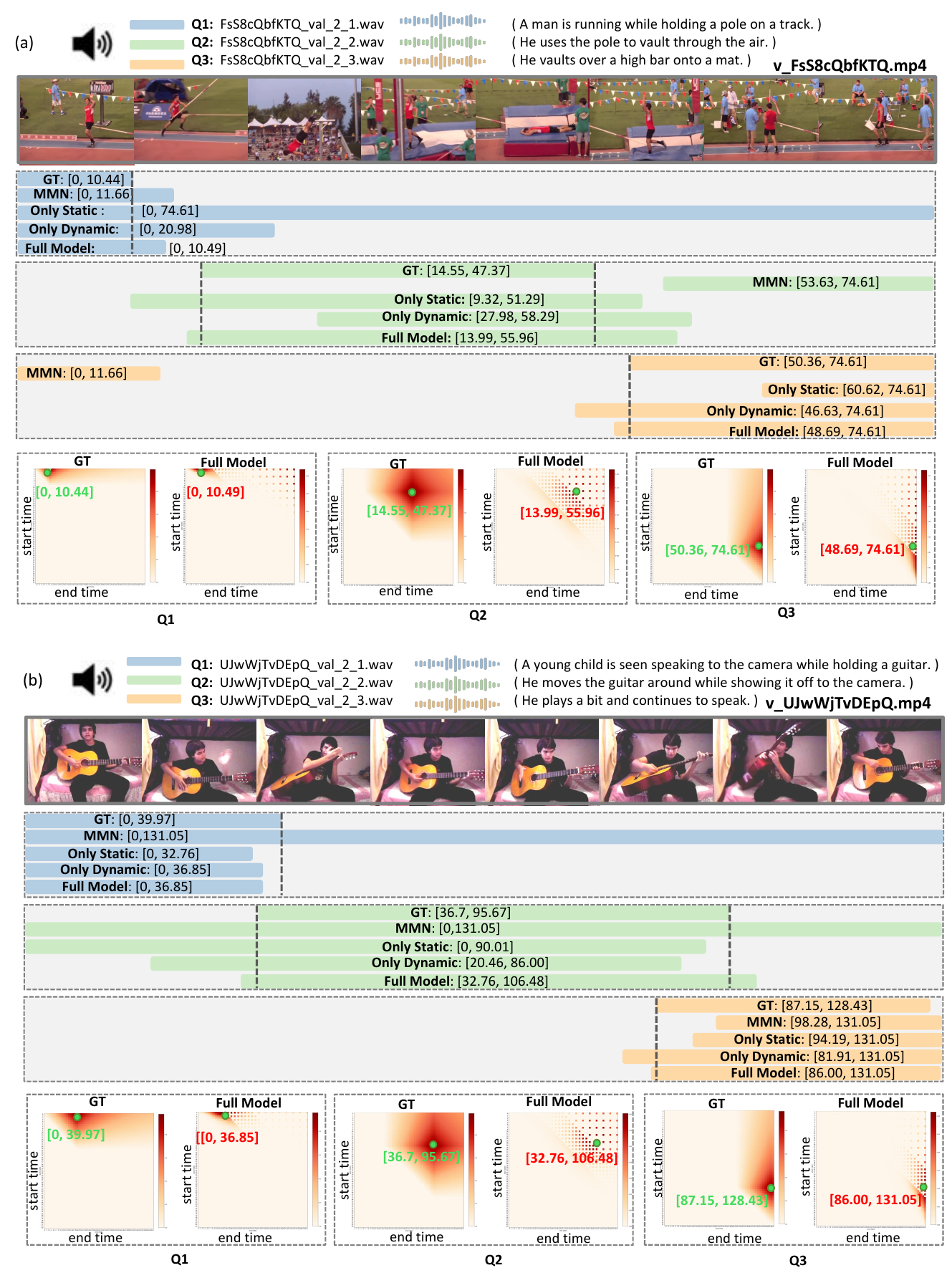}
\caption{Qualitative examples on ActivityNet Speech for SLVG.
\textbf{(a)} The scenes that contains a noisy background.
\textbf{(b)} The Videos with Continuous and Varied Actions. 
\textbf{MMN} makes significant errors in predicting the location ($Q2$ and $Q3$ in case (a)) and location coverage areas of events ($Q1$ and $Q2$ in case (b)). These two cases are challenging. Encouragingly, our \textbf{Full Model} achieves the best performance in these video grounding cases based on audio queries, which confirms the effectiveness and generalization of our unified static and dynamic methods in this task.}
\label{fig:app_vis_anet_slvg2}
\end{figure*}

\textbf{Video Sample with Noisy Background.} When using audio as a query, we prefer to analyze how well the model understands the interaction between audio and video by performing visualizations of video cases that contain more background noise. 
We instantiate the video sample ``ID: v\_FsS8cQbfKTQ'' from the ActivityNet Speech dataset in Fig.~\ref{fig:app_vis_anet_slvg2} (a) using audio queries under noisy background interference. 
We can see that \textbf{MMN} predicts the video clips corresponding to $Q2$ and $Q3$ with significant deviations, and the predicted moments totally do not intersect with \textbf{GT} at all. This video is a challenging case. Compared to \textbf{MMN}, \textbf{Only Static} and \textbf{Only Dynamic} coverage the queried moment but have somewhat boundary shifts, exhibits a strong advantage, as it correctly predicts the relative positions of all events has a large intersection ratio with \textbf{GT} video clips. 
Compared to \textbf{MMN}, our \textbf{Full model} exhibits the best prediction results for all queries, as it correctly predicts the queried moment and has a large intersection ratio with \textbf{GT} video clips. From the 2D map in the figure, it can be seen that our model still performs well in video grounding task based on audio queries, fully demonstrating the generalization of our model. 

\textbf{The Videos with Continuous and Varied Actions.} 
Similarly to the NLVG task, we analyse the video case without transitions but with continuous action changes for the SLVG task to quantify the model's ability of identifying event boundaries. 
Taking video sample ``ID: v\_UJwWjTvDEpQ'' from the ActivityNet Speech dataset in Fig.~\ref{fig:app_vis_anet_slvg2} (b) as an example,
the video shows a scene with a clean background, but in which a boy's actions are continuously changing. 
In this case, for different event divisions, it is necessary to finely distinguish the contentual semantics of the boy's actions and the differences between them.
\textbf{MMN} fails to recognize such densly varied actions and incorrectly assigns the entire video as the answer (\eg, $Q1$ and $Q2$). 
\textbf{Our Static} predicts the approximate location of each event.
\textbf{Our Dynamic} exhibits excellent performance in distinguishing the semantics of continuous actions, it not only correctly distinguishes the semantic centers of three events, but also more accurately predicts the boundaries of each event, compared to \textbf{MMN} and \textbf{Our Static}. 
Inspiring, \textbf{Full Model} achieves the most accurate prediction of the location and semantic boundaries of events, this is thanks to the combination of static and dynamic modes, which deepens the understanding of video context and enables the model to distinguish different action semantics. 

\subsection{More Visualization of Plethoric Multi-query Cases}

\begin{figure*}[t!]
\centering
\includegraphics[width=0.9\linewidth]{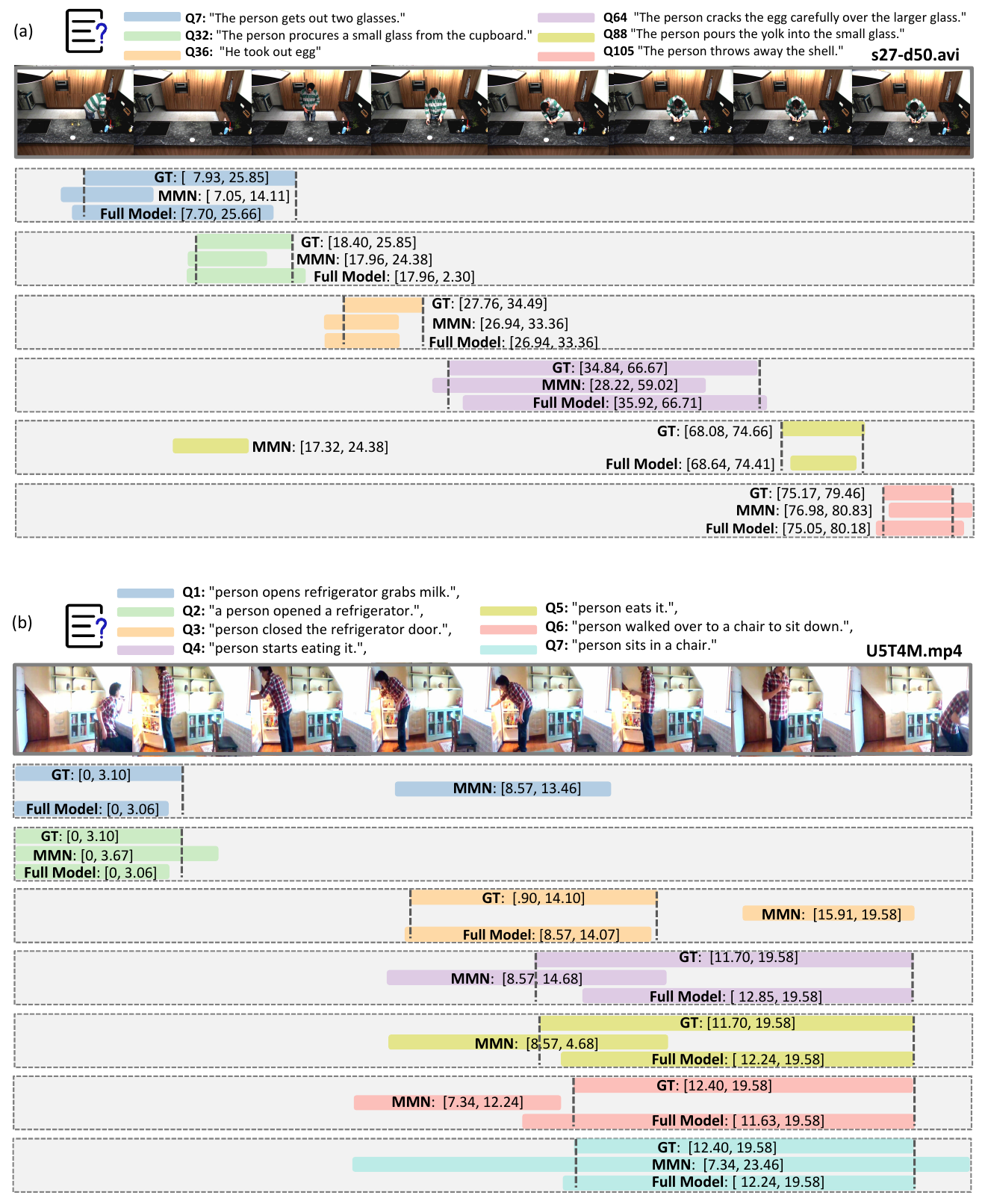}
\caption{Quantitative examples of plethoric multi-query cases.
\textbf{(a)} Examples on the TACoS dataset for NLVG.
\textbf{(b)} Examples on the Charades-STA dataste for NLVG.
\textbf{MMN} has a significant semantic bias when predicting $Q7$ in case (a), and $Q4,Q5,Q7$ in case (b), there is also a large positional deviation in predicting $Q88$ in case (a), and $Q1, Q3$ in case (b).
Our \textbf{Full Model} correctly predicts the location of all the queried events, and the predicted moment interval is closest to that of \textbf{GT}, this is thanks to model capacity of mutual learning of video and multiple queries and effectively capturing the video context associated with multiple queries.
}
\label{fig:sta-app}
\vspace{-0.5em}
\end{figure*}

\textbf{Visualization Examples on the TACoS Dataset.}
Taking the video sample ``ID: s27-d50'' in Fig.~\ref{fig:sta-app} (a) as an example, we provide the grounding results of our model and MMN. Note that the total duration of the video is 82.11 s, which includes 119 query descriptions. Limited by page size and layout, we select and show 6 very challenging queries here. The video depicts a person cooking in a kitchen. 
\textbf{MMN} experiences a significant prediction error in the moment corresponding to the query $Q88$. On the contrary, our \textbf{Full model} accurately determines the relative positions of the video segments corresponding to all queries. The qualitative results highlight the effectiveness of learning semantic associations between multi-queries (\ie, multi-queries contextualization) for cross-modal video grounding.

\textbf{Visualization Examples on the Charades-STA Dataset.}
The video sample ``ID: U5T4M'' in Fig.~\ref{fig:sta-app} (b) has a duration of 19.58 s, which describes the indoor activities of a person, and contains 7 queries. Our \textbf{Full model} infers the localization results of all queries corresponding to the video at once. 
In all queries, $Q1$ and $Q2$ are similar descriptions of an event, respectively. The same situation also includes queries of $Q4$ and $Q5$, $Q6$ and $\displaystyle Q7$. 
Our \textbf{Full model} accurately predicts the boundaries of each query, and effectively distinguishing the semantics among similar but with slightly different events. 

\end{document}